\RequirePackage{fix-cm}

\documentclass[smallextended]{svjour3}       

\smartqed  
\usepackage{graphicx}
\graphicspath{{figures/}}
\usepackage{amsmath}
\usepackage{amssymb}
\usepackage{textcomp} 
\usepackage{gensymb}
\usepackage{enumerate}
\usepackage[colorlinks=true, allcolors=blue]{hyperref}
\usepackage{bbold}
\usepackage{array}
\usepackage{doi}
\usepackage{booktabs}
\usepackage{soul}
\usepackage{tikz}
\usetikzlibrary{math}
\usetikzlibrary{calc}
\usetikzlibrary{intersections}
\usepackage{array}

\newcommand{\symmdiff}{\ominus}
\newcommand{\M}{{\mathcal M}}
\newcommand{\x}{{\bf x}}
\newcommand{\y}{{\bf y}}

\newcommand{\p}{\textup{\bf p}}
\newcommand{\s}{\textup{\bf s}}
\newcommand{\B}{{\mathcal{B}}}
\newcommand{\exact}{\operatorname{Exact}(\sigma)}
\newcommand{\approxone}{\operatorname{Approx}^{(1)}(\sigma)}
\newcommand{\approxtwo}{\operatorname{Approx}^{(2)}(\sigma)}

\newcommand{\tilt}{{\operatorname{tilt}_{\M, \B, \p}}}

\spnewtheorem{Theorem}{Theorem}{\bfseries}{\itshape}
\spnewtheorem{Lemma}[Theorem]{Lemma}{\bfseries}{\itshape}
\spnewtheorem{Corollary}[Theorem]{Corollary}{\bfseries}{\itshape}
\spnewtheorem{Proposition}[Theorem]{Proposition}{\bfseries}{\itshape}
\spnewtheorem{Propdef}[Theorem]{Proposition/Definition}{\bfseries}{\itshape}

\spnewtheorem{Definition}{Definition}{\bfseries}{\rmfamily}
\spnewtheorem{Lemdef}[Definition]{Lemma/Definition}{\bfseries}{\rmfamily}
\spnewtheorem{Cordef}[Definition]{Corollary/Definition}{\bfseries}{\rmfamily}

\spnewtheorem{Example}{Example}{\itshape}{\rmfamily}
\spnewtheorem{Remark}{Remark}{\itshape}{\rmfamily}

\usepackage{collectbox}



\begin{document}

\title{Manifold learning with arbitrary norms}

\author{Joe~Kileel$^{\star}$ \and Amit~Moscovich \and Nathan~Zelesko \and Amit~Singer}

\authorrunning{Joe Kileel, Amit Moscovich, Nathan Zelesko, Amit Singer}

\institute{
$^{\star}$Author to whom correspondence should be addressed. \\[4pt]
    Joe Kileel \at Department of Mathematics and Oden Institute, University of Texas at Austin\\
    \email{jkileel@math.utexas.edu}
    \and
    Amit Moscovich \at Department of Statistics and Operations Research, Tel-Aviv University\\
    \email{amit@moscovich.org}
    \and
     Nathan Zelesko \at Department of Mathematics, Brown University \\ \email{nathan@zelesko.com}
     \and
    Amit Singer \at  Department of Mathematics and Program in Applied and Computational Mathematics, Princeton University\\     \email{amits@math.princeton.edu}
}

\date{}

\maketitle
\vspace{-5em}
\begin{abstract}
    Manifold learning methods play a prominent role in nonlinear dimensionality reduction and other tasks involving high-dimensional data sets with low intrinsic dimensionality.
    Many of these methods are graph-based: they associate a vertex with each data point and a weighted edge with each pair.   
     Existing theory shows that the Laplacian matrix of the  graph converges to the Laplace-Beltrami operator of the data manifold, under the assumption that the pairwise affinities are based on the Euclidean norm.
    In this paper, we determine the limiting differential operator for graph Laplacians constructed using \emph{any} norm.
    Our proof involves an interplay between the second fundamental form of the manifold and the convex geometry of the given norm's unit ball. 
    To demonstrate the potential benefits of non-Euclidean norms in manifold learning, we consider the task of mapping the motion of large molecules with continuous variability. In a numerical simulation we show that a modified Laplacian eigenmaps algorithm, based on the Earthmover's distance, outperforms the classic  Euclidean Laplacian eigenmaps, both in terms of computational cost and the sample size needed to recover the intrinsic \nolinebreak geometry. 
\end{abstract}

\section{Introduction}

Manifold learning is broadly concerned with analyzing high-dimensional data sets that have a low intrinsic dimensionality.
The standard assumption is that the input data set $\mathcal{X} = \{\x_1, \ldots, \x_n\} \subseteq \mathbb{R}^D$ 
 lies on or near a $d$-dimensional submanifold $\M \subseteq \mathbb{R}^D$ where $ d \ll D$.
The key tasks are dimensionality reduction \cite{TenenbaumDesilvaLangford2000,RoweisSaul2000,DonohoGrimes2003,BelkinNiyogi2003,ZhangZha2004,CoifmanLafon2006,VandermaatenHinton2008,McinnesEtal2018,ZhangMoscovichSinger2021}, function representation and approximation \cite{GavishNadlerCoifman2010,ChengWu2013,LiaoMaggioniVigogna2016,SoberAizenbudLevin2021} and semi-supervised learning \cite{BelkinNiyogi2004,GoldbergEtal2009,MoscovichJaffeNadler2017}.
Most data analysis methods in this setting rely on pairwise  Euclidean  distances between the data points. 

In this paper, we focus on  manifold learning methods that use a graph Laplacian. These include the popular spectral embedding methods Laplacian eigenmaps \cite{BelkinNiyogi2003,BelkinNiyogi2004} and diffusion maps \cite{CoifmanLafon2006}. Both methods map the input points to the eigenvectors of a graph Laplacian operator $\mathcal{L}_n$ (or weighted variant thereof).
By definition, $\mathcal{L}_n$ acts on a function $f: \mathcal{X} \rightarrow \mathbb{R}$ \nolinebreak via
\begin{align} \label{def:Ln}
    \left(\mathcal{L}_n f \right)(\x_i) := \sum_{j=1}^n W_{ij} \left( f(\x_j) - f(\x_i) \right), \quad&& W_{ij} := \exp\left(-\frac{\|\x_j-\x_i\|_2^2}{\sigma_n^2}\right).
\end{align}
Under suitable conditions, as $n \to \infty$ the discrete graph Laplacian operator $\mathcal{L}_n$ converges to the continuous Laplace-Beltrami operator $\Delta_{\mathcal{M}}$ on the manifold~\cite{BelkinNiyogi2008},
and its eigenvectors converge to the Laplacian eigenfunctions \cite{BelkinNiyogi2007}.
While the convergence results can be extended to more general affinity kernels of the form $W_{ij} = K_{\sigma_n}(\|\x_i-\x_j\|_2)$, the role of the Euclidean norm here is essential.
This poses a potential limitation for graph Laplacian methods since Euclidean metrics are not always the best choice for all application domains \cite{BelletHabrardSebban2015}.
Furthermore, some non-Euclidean metrics use compressed representations, which can have practical benefits in terms of runtime and memory requirements. 
Following this line of reasoning leads to the following questions: can the machinery of discrete Laplacian operators be generalized to non-Euclidean metrics?
Does doing so yield any practical benefits? If so, what is the underlying theory?

This paper is an initial step in answering these questions.
The main contribution of the paper is the derivation of the continuum limit of discrete Laplacian operators similar to \eqref{def:Ln} but with an affinity kernel based on an arbitrary norm.
Our key result (Theorem~\ref{thm:limit}) is a proof that using \emph{any norm}, graph Laplacians converge to an explicit second-order differential operator on $\mathcal{M}$.
In contrast to the Euclidean case, in the general case the limiting operator is not intrinsic to the manifold, i.e., it depends
on the embedding of $\mathcal M$ in $\mathbb{R}^D$.
Furthermore, it has non-vanishing and possibly discontinuous first-order terms.
The second-order coefficients of the limiting differential operator at a point $\p \in \M$ is given by the second moments of the intersection of the tangent plane to $\mathcal{M}$ at $\p$ and the given norm's unit ball. 
The first-order terms depend on the second fundamental form of $\mathcal{M}$ at $\p$ and the tangent cones to the norm's unit sphere, through a function we call $\operatorname{tilt}$, defined in Section \ref{subsec:tilt-const}.

In a second contribution, which was the original motivation for this work, we present in Section~\ref{sec:experiments} a variant of Laplacian eigenmaps that is based on a norm that approximates the Earthmover's distance (EMD), also known as the Wasserstein-1 metric, for learning volumetric shape spaces.
This is motivated by an important problem in structural biology: learning the conformation space of flexible proteins and
other macromolecules with continuous variability from cryo-electron microscopy images.
Empirically, we demonstrate that classical (Euclidean) Laplacian eigenmaps are at a disadvantage compared to Laplacian eigenmaps based on this approximate EMD, as it requires far fewer sample points to recover the intrinsic manifold of motion.
Furthermore, as we show in Section \ref{sec:sparsity}, the proposed method can achieve faster runtime and a smaller memory footprint through an intermediate compressed representation.
This demonstrates, at least for certain data sets, the use of non-Euclidean norms in Laplacian-based manifold learning is desirable from a practical view.

\subsection{\textbf{Related work}}

Our convergence proof builds on the well-known proof of Belkin and Niyogi's for the case of the Euclidean norm  \cite{BelkinNiyogi2008}.
However, the argument for the Euclidean case is \textit{not} directly
adaptable to the case of other norms. 
It relies on a special property of the Euclidean norm: that Euclidean distances provide a second-order
approximation to manifold geodesic distances (see [6, Figure 1]).  
This fails for general norms, which do not even give a first-order approximation to geodesic distances. 
This difference introduces a first-order derivative term in the limit in the general case.
Another technical difference is that,
in the standard case, the intersection of an embedded tangent space with the Euclidean unit ball is rotationally symmetric.  This gives rise to the Laplace-Beltrami operator, which is the only second-order rotationally symmetric differential operator (up to scale).
The property fails for general norms, thereby introducing ``cross-terms" in the second-order term of the general limit. 

In \cite{TingHuangJordanICML2010}, a different extension of the convergence proof for graph Laplacian methods appeared.  That work analyzed $k$-nearest neighbor graphs and other constructions, but based on a Euclidean norm.
We do not pursue this direction.

To the best of our knowledge, most Laplacian-based manifold learning works employ the standard Euclidean norm.
Two notable exceptions are the works of Mishne and collaborators \cite{MishneEtal2016,MishneEtal2017} where tree-based metrics \cite{CoifmanLeeb2013} were used as a basis for diffusion maps.
These metrics can be interpreted as hierarchical Earthmover's distances. 
However since the trees are data-dependent, our main theorem does not apply since we do require a data-independent norm.

The application section in this paper is an extension of  \cite{ZeleskoMoscovichKileelSinger2020}, where we first proposed to use a variant of diffusion maps based on an approximate Earthmover's distance.
In \cite{RaoMoscovichSinger2020}, the same approximate Earthmover's distance was used for the clustering of cryo-EM images.
Lieu and Saito's work \cite{LieuSaito2011} is another  that combines diffusion maps and the Earthmover's distance.
However, the order of operations is different: they first use Euclidean diffusion maps and only then apply the Earthmover's distance to the resulting point clouds.

\begin{table}
    \begin{center}
        \renewcommand{\arraystretch}{1.55}
        \normalsize
        \begin{tabular}{ll} 
            \bf Symbol  & \bf Description\\\hline
            $\M \subseteq \mathbb{R}^D$ & Compact   embedded Riemannian submanifold \\
            $d = \dim(\mathcal{M})$ & Dimension of $\mathcal{M}$ \\
            $\p \in \mathcal{M}$ & Point on $\mathcal{M}$\\
            $T_\p \M$ & Tangent space to $\M$ at $\p$ \\
            $\textup{exp}_{\p}: T_{\p} \M \rightarrow \M$ & Exponential map for $\mathcal{M}$ at $\p$ \\
            $\s \in \mathbb{R}^d \cong T_{\p}\mathcal{M}$  & Geodesic normal coordinates for $\mathcal{M}$ around $\p$ \\
                        $f : \mathcal{M} \rightarrow \mathbb{R}$ & Function on $\mathcal{M}$ \\
            $\widetilde f = f \circ \operatorname{exp}_{\p}$ & Function pulled-back to tangent space\\
            $\operatorname{grad}\widetilde f: \mathbb{R}^d \to \mathbb{R}^d$ & Gradient of $\widetilde{f} $\\
            $\operatorname{hess}\widetilde f: \mathbb{R}^d \to \mathbb{R}^{d \times d}$ & Hessian of $\widetilde{f}$\\                       
            $L_\p: T_\p \M \rightarrow \mathbb{R}^D$ & Differential of exponential map at $\p$ (Eq. \eqref{eq:expp-taylor}) \\ 
            $Q_\p:T_\p\M \rightarrow \mathbb{R}^D$ & Second fundamental form of $\mathcal{M}$ at $\textbf{p}$ (Eq. \eqref{eq:expp-taylor})
  \\
             $\mathcal{B} \subseteq \mathbb{R}^D$ & Origin-symmetric convex body \\
            $\| \cdot \|_\B : \mathbb{R}^D \to \mathbb{R}_{\ge 0} $ & Norm with unit ball $\B$ \\
            $\| \cdot \|_2: \mathbb{R}^D \to \mathbb{R}_{\ge 0}$ & Euclidean norm \\
            $\| \cdot \|_{\textbf{w},1}: \mathbb{R}^D \to \mathbb{R}_{\ge 0}$ & Weighted $\ell_1$-norm  \\
            $K_{\sigma} : \mathbb{R}_{\geq 0} \rightarrow \mathbb{R}_{\geq 0}$ & Affinity kernel with width parameter $\sigma > 0$ \\
            $\mathcal{L}_{n}$ & Point-cloud Laplacian based on  $\| \cdot \|_2$ (Eq. \eqref{eq:discrete-Lap})\\
            $\mathcal{L}_{n, \B}$ & Point-cloud Laplacian based on $\| \cdot \|_{\mathcal{B}}$ (Definition \ref{discrete_laplacian_B})\\
            $\Delta_\M$ & Laplace-Beltrami operator on $\mathcal{M}$ \\
            $\Delta_{\mathcal{M}, \mathcal{B}}$ & Laplacian-like differential operator (Definition \ref{def:Laplace-like})\\
            $\tilt$ & Tilt function at $\p$ (Proposition/Definition \ref{def:tilt-const}) \\
                $\overline{S}, S^{\circ}, \partial S$ & Closure, interior, boundary of a set  \\
             $TC_{\textbf{y}}(\mathcal{Y}) \subseteq \mathbb{R}^D$ & Tangent cone to $\mathcal{Y} \subseteq \mathbb{R}^D$
at $\textup{\textbf{y}} \in \mathcal{Y}$ (Eq. \eqref{eq:def-tan-cone}) \\
      $\mathcal W$ & Wavelet transform  \\
                 $\langle \cdot, \cdot \rangle$ & Inner product  \\
                  $\mathbb{R}_{\ge 0}, \mathbb{R}_{>0}$ & Non-negative/strictly positive real numbers
        \end{tabular}
    \end{center}
    \caption{\small List of notation.} \label{table:notation}
\end{table}
\addcontentsline{toc}{subsection}{List of symbols}

\section{\textbf{Background: graph Laplacian methods}} \label{subsec:graphLaplacian}

In this section, we review graph Laplacian methods in more detail than in the introduction.
Given a subset $\mathcal{X} =\{\x_1, \ldots, \x_n\} \subseteq \mathbb{R}^D$, and an affinity function $K_{\sigma_n} : \mathbb{R}_{\geq 0} \rightarrow \mathbb{R}_{\geq 0}$, consider the symmetric matrix of pairwise affinities:
\begin{align} \label{def:W_ij}
    W_{ij} := K_{\sigma_n}(\|\x_i - \x_j\|_2).
\end{align}
The canonical choice for $K_{\sigma_n}$ is the Gaussian kernel, $K_{\sigma_n}(t) = \exp(-t^2/\sigma_n^2)$ (up to normalization conventions) where the width parameters $\sigma_n$ decay to zero at an appropriate rate.
Another possibility is the 0/1 kernel, $K_{\sigma_n}(t) = \mathbb{1}(t \leq \sigma_n)$.
The matrix $W$ defines a weighted graph  $G=(\mathcal{X},E,W)$ where the set of edges $E$ consists of all the pairs $(i,j)$ for which $W_{ij} > 0$.
Define the diagonal degree matrix by $D_{ij} = \delta_{ij}\sum_k W_{ik}$.  The (unnormalized, negative semi-definite) Laplacian matrix of $G$, or the \emph{graph Laplacian}, is defined to be
\begin{align} \label{def:LG}
    \mathcal{L}_G := W - D.
\end{align}

\vspace{-0.5em}

\begin{Remark} \label{remark:negsemidefinite}
    As a warning, several other authors use the \textit{positive} semi-definite graph Laplacian convention, $\mathcal{L}^{\textup{psd}}_G := D - W$. 
    In this paper, we chose the negative semi-definite conventions for both the discrete and continuous Laplacians.
\end{Remark}
The graph Laplacian acts on vectors $f \in \mathbb{R}^n$.  We think of $f$ as a real-valued function on the vertex set $\mathcal{X}$. 
Then the graph Laplacian averages, for each vertex, the differences between the function's value at the vertex and its neighbors:
\begin{align}\label{eq:rightmost}
      \left(\mathcal{L}_G f \right)(\x_i) = \sum_{j=1}^n W_{ij}(f(\x_j) - f(\x_i)) = \sum_{j=1}^n K_{\sigma_n}(\|\x_j - \x_i\|_2) \left( f(\x_j) - f(\x_i)\right).
\end{align}
Note $\mathcal L_G$ is an $n \times n$ symmetric negative semi-definite matrix.  
We list its eigenvalues in descending order
\begin{align*}
    0 = \lambda_0 \geq \lambda_1 \geq \ldots \geq \lambda_{n-1},
\end{align*}
and choose corresponding real orthonormal eigenvectors
\begin{align*}
 \phi_0, \ldots, \phi_{n-1} \in \mathbb{R}^n
\end{align*}
where $\phi_0 = n^{-1/2} \mathbf{1} $.
These eigenvectors give an orthonormal basis of functions on $\mathcal{X}$. 
Two common uses for the Laplacian eigenvectors are:
\begin{enumerate}
    \item As a basis for function representation and approximation  of real-valued functions $g$ defined on  $\mathcal X$ \cite{BelkinNiyogi2004,CoifmanEtal2005,LeeIzbicki2016},
    \begin{align*}
        g(\x_i) = \sum_{j=0}^{n-1} \alpha_j \phi_j(\x_i).
    \end{align*}
    \item As a method for dimensionality reduction of the input set $\mathcal{X}$ \cite{BelkinNiyogi2003,CoifmanLafon2006},
        \begin{align*}
        \x_i \mapsto \left( \phi_1(\x_i), \ldots, \phi_m (\x_i) \right).
    \end{align*}
    Here, each $\textbf{x}_i$ is mapped into $\mathbb{R}^m$ via the $i$-th coordinates of the first $m$ nontrivial Laplacian eigenvectors.
    This usage of eigenvectors is motivated by the fact that any closed connected Riemannian manifold is smoothly embedded into $\mathbb{R}^m$ by its first $m$ Laplacian eigenfunctions for some $m$ \cite{Bates2014}.
\end{enumerate}
The Laplacian matrices $\mathcal{L}_{G}$ are often analyzed as points are added to $\mathcal{X}$ and the graph $G$ grows.  
In this context, the \textit{manifold assumption}
is standard, namely that $\textbf{x}_i$ are drawn i.i.d. from some embedded submanifold $\mathcal{M} 
\subseteq \mathbb{R}^D$ 
Other works, for example in the study of clustering, have analyzed the limit of graph Laplacians without making the manifold assumption (see \cite{VonLuxburg2007,TrillosSlepcev2018}).
It is convenient to work with an extension of the graph Laplacian that acts on any function $f$ whose domain is a superset of $\mathcal{X}$.  Specifically we define the (unnormalized, negative semi-definite) \textit{point-cloud Laplacian} $\mathcal{L}_n$ computed using $\mathcal{X}$ as follows: for each $f : \mathcal{Y} \rightarrow \mathbb{R}$ where $\mathcal{X} \subseteq \mathcal{Y}$, define $\mathcal{L}_n f : \mathcal{Y} \rightarrow \mathbb{R}$ by
\begin{align} \label{eq:discrete-Lap}
    \mathcal{L}_nf(\p) := \frac{1}{n} \sum_{j=1}^n K_{\sigma_n}(\| \textbf{x}_j - \p \|_2)(f(\textbf{x}_j) - f(\p)).
\end{align}
After rescaling by $n$ the point-cloud Laplacian \eqref{eq:discrete-Lap} extends the graph Laplacian \eqref{eq:rightmost}, because for each $f : \mathcal{Y} \rightarrow \mathbb{R}$ where $\mathcal{X} \subseteq \mathcal{Y}$ it holds $\left( n\mathcal{L}_n f  \right)\!|_{\mathcal{X}} = \mathcal{L}_G(f|_{\mathcal{X}})$.

\subsection{\textbf{Existing theory: graph Laplacians using the Euclidean norm}}

It is known that using Euclidean norms to compute affinities, the point-cloud Laplacian converges to the Laplace-Beltrami operator under the manifold assumption
(see \cite{HeinAudibertLuxburg2005,Singer2006,BelkinNiyogi2008,GineKoltchinskii2006}).  Here is a precise statement.

\begin{Theorem}[\normalfont {\cite[Th.~3.1]{BelkinNiyogi2008}}:  Convergence of the point-cloud Laplacian based on the Euclidean norm] \label{thm:classical}
    Let $\M$ be a compact $d$-dimensional embedded Riemannian submanifold of $\mathbb{R}^D$ 
    with Laplace-Beltrami operator $\Delta_\M$.  
    Let $\textup{\textbf{x}}_1, \ldots, \textup{\textbf{x}}_n$ be  i.i.d. draws from the uniform measure on $\mathcal{M}$. 
    Fix any constant $\alpha > 0$, and set $\sigma_n = 2 n^{-1/(2d+4+\alpha)}$ and $c_n = \frac{\pi^{d/2}}{4} \sigma_n^{d+2} $.
    Let $\mathcal{L}_n$ be the point-cloud Laplacian defined in Eq.~\eqref{eq:discrete-Lap} using the Gaussian affinity based on the Euclidean norm, $K_{\sigma}(\|\textup{\textbf{x}}_{j} - \p\|) = \exp(\| \textup{\textbf{x}}_j - \p \|_2^2/\sigma_n^2)$.
    Then given a three-times continuously differentiable function $f : \mathcal{M} \rightarrow \mathbb{R}$ and a point $\p \in \mathcal{M}$,
    we have the following convergence in probability:
\begin{align*}
    \frac{1}{c_n} \mathcal{L}_n f(\p) \xrightarrow{\,\,\,\,\, p \,\,\,\,} \frac{1}{\operatorname{vol}(\mathcal{M})} \Delta_{\mathcal{M}}f(\p).
\end{align*}
\end{Theorem}
\noindent Our Theorem~\ref{thm:limit} extends Theorem~\ref{thm:classical} to the case of non-Euclidean norms.

Meanwhile, for a non-uniform sampling distribution, variants of the point-cloud Laplacian are known to converge pointwise to a weighted Laplacian (or Fokker--Planck)
operator, which has an additional density-dependent drift term  \cite{CoifmanLafon2006,NadlerLafonCoifmanKevrekidis2005,TingHuangJordanICML2010}.
Theorem~\ref{thm:nonuniform} extends this to the case of non-Euclidean norms.

In addition to pointwise consistency, spectral consistency has been proven when the norm is Euclidean \cite{BelkinNiyogi2007,HeinAudibertVonLuxburg2007,VonluxburgBelkinBousquet2008,RosascoBelkinDevito2010,TrillosSlepcev2018,TrillosGerlachHeinSlepcev2020,WormellReich2021}.
This is a stronger mode of convergence than pointwise convergence: the eigenvalues/eigenvectors of the graph Laplacians converge to the eigenvalues/eigenfunctions of the limiting operator.  We leave such considerations for arbitrary norms to future work.

\section{Ingredients, main theorem statement, first properties}

In this section the primary goal is to formulate our main result, Theorem~\ref{thm:limit}, in Section~\ref{subsec:thm-statement}. 
Before this, we collect tools from differential geometry and convex geometry.  
Then in Section~\ref{subsec:tilt-const}, we define a particular function that depends on the second-order geometry of a manifold and the unit ball of a given norm; this turns out to give the correct first-order derivative term in Theorem~\ref{thm:limit}.
After the main statement, we explain how it adapts to non-uniform sampling of the manifold (Theorem~\ref{thm:nonuniform}).
Then we discuss first properties of the limiting differential operator and show that it reduces to the Laplace-Beltrami operator in the Euclidean case.
As a non-Euclidean example, we calculate the limit explicitly for a circle in the plane where the ambient norm is weighted~$\ell_1$.

\subsection[\textbf{Preliminaries from Riemannian geometry}]{\textbf{Preliminaries from Riemannian geometry}} \label{subsec:prelim-riem}

We start by reviewing some basics from Riemannian geometry.  
These notions are later used for our theorem statement and its proof.  
Textbook accounts of differential geometry are abundant; we particularly like Lee's books \cite{Lee-Riem-Book,LeeBook2012}.

Throughout the paper, $\mathcal{M} \subseteq \mathbb{R}^D$ denotes a $d$-dimensional compact embedded \textit{Riemannian submanifold} of $\mathbb{R}^D$.
We let $\textbf{p} \in \mathcal{M}$ denote a point (typically fixed in our considerations). 
We write $T_{\textbf{p}}\mathcal{M}$ for the abstract \textit{tangent space} to $\mathcal{M}$ at $\textbf{p}$ (defined in \cite[Ch.~3]{LeeBook2012}).  
In particular, $T_{\textbf{p}}\mathcal{M}$ is a $d$-dimensional real vector space equipped with an inner product $\langle \cdot, \cdot \rangle = \langle \cdot , \cdot \rangle_{\textbf{p}}$.
Further, $0 \in T_{\textbf{p}}\mathcal{M} $ and we do \textit{not} consider $T_{\textbf{p}} \mathcal{M}$ to be embedded in $\mathbb{R}^D$. 
The canonical mapping from the tangent space into the manifold is the \textit{exponential map} at $\textbf{p}$ \cite[Ch.~5]{Lee-Riem-Book}, denoted   $\textup{exp}_{\textbf{p}} : T_{\textbf{p}} \mathcal{M} \rightarrow \mathcal{M}$.\footnote{Compactness of $\mathcal{M}$ and the Hopf-Rinow theorem imply that $\textup{exp}_{\textbf{p}}$ is defined on the entire tangent space $T_{\textbf{p}} \mathcal{M}$.}  
This is a $C^{\infty}$-map that carries straight lines on $T_{\textbf{p}} \mathcal{M}$ through the origin to \textit{geodesics} on $\mathcal{M}$  through the point $\textbf{p}$.
By the inverse function theorem, there exist open neighborhoods $\mathcal{U} \subseteq T_{\textbf{p}}\mathcal{M}$ of $0$ and $\mathcal{V} \subseteq \mathcal{M}$ of $\textbf{p}$ (which we fix once and for all) such that the exponential map restricts to a \textit{diffeomorphism} between these neighborhoods,
\begin{align*}
    \textup{exp}_{{\bf p}}:  U \xrightarrow{\sim} V. 
\end{align*}
Further, let us fix once and for all an orthonormal basis on $T_{\textbf{p}}\mathcal{M}$ with respect to $\langle \cdot, \cdot \rangle_{\textbf{p}}$, and write $\textbf{s} = (s_1, \ldots, s_d)^{\top}$ for coordinates on $U$ with respect to this basis; these are \textit{geodesic normal coordinates} for $\mathcal{M}$ around $\textbf{p}$ with the chart given by the exponential map.  
Identifying
$\textup{exp}_{\textbf{p}}$ with $\iota \circ \textup{exp}_{\textbf{p}}$, where $\iota : \mathcal{M} \hookrightarrow \mathbb{R}^D$ is inclusion,  $\textup{exp}_{\textbf{p}}$ is a smooth mapping from an open subset of Euclidean space $\mathbb{R}^d$ into Euclidean space $\mathbb{R}^D$, and thus it admits a Taylor expansion around $\textbf{s}=0$:
\begin{align} \label{eq:expp-taylor}
    \textup{exp}_{\textbf{p}}(\textbf{s}) = \textbf{p} + L_{\textbf{p}}(\textbf{s}) + \frac{1}{2} Q_{\textbf{p}}(\textbf{s}) + O(\|\textbf{s}\|_2^3).
\end{align}
Equation \eqref{eq:expp-taylor} links \textit{intrinsic coordinates} to \textit{extrinsic coordinates} for $\mathcal{M}$.  Here:
\begin{itemize}
  \item  $L_{\textbf{p}} : T_{\textbf{p}}\mathcal{M} \rightarrow \mathbb{R}^D$  is a homogeneous linear function,  the \textit{differential} of the exponential map at $\textbf{p}$, namely $L_{\textbf{p}} = D\textup{exp}_{\textbf{p}}(0)$; and 
    \item $Q_{\textbf{p}} : T_{\textbf{p}}\mathcal{M} \rightarrow \mathbb{R}^D$ is a homogeneous quadratic function (equivalently a linear function of $\textbf{s}\textbf{s}^{\top}$) called the \textit{second fundamental form} of $\mathcal{M}$ at $\textbf{p}$~\cite{Monera2014}.
\end{itemize}
Consider the image (respectively, translated image) of $L_{\textbf{p}}$:
\begin{align} 
     L_{\textbf{p}}(T_{\textbf{p}} \mathcal{M}) & = \{ L_{\textbf{p}}(\textbf{s}) : \textbf{s} \in T_{\textbf{p}}\mathcal{M}   \} \subseteq \mathbb{R}^D, \text{ and } \label{eq:emd-linear-tang} \\
    \textbf{p} + L_{\textbf{p}}(T_{\textbf{p}} \mathcal{M}) &= \{ \textbf{p} + L_{\textbf{p}}(\textbf{s}) : \textbf{s} \in T_{\textbf{p}}\mathcal{M}   \} \subseteq \mathbb{R}^D.  \nonumber
\end{align}
We call these the \textit{linear (respectively, affine) embedded tangent space} of $\mathcal{M}$ at $\textbf{p}$.
It is well-known that $L_{\textbf{p}}$ provides an \textit{isometric embedding} of $T_{\textbf{p}}\mathcal{M}$ into $\mathbb{R}^D$, 
\begin{equation} \label{eq:isometric}
    \| L_{\textbf{p}}(\textbf{s}) \|_2 = \| \textbf{s} \|_2 \,\, \text{ for all } \textbf{s} \in T_{\textbf{p}}\mathcal{M}.
\end{equation}
Another important fact is that the second fundamental form $Q_{\textbf{p}}$ takes values in the \textit{normal space} to $\mathcal{M}$ at $\textbf{p}$, that is $Q_{\textbf{p}}(T_{\textbf{p}} \mathcal{M}) \subseteq  L_{\textbf{p}}(T_{\textbf{p}}\mathcal{M})^{\perp} \subseteq \mathbb{R}^D$, i.e.,
\begin{equation} \label{eq:perp}
    \langle L_{\textbf{p}}(\textbf{s}), Q_{\textbf{p}}(\textbf{s}') \rangle_{\mathbb{R}^D} = 0 \,\, \text{for all } \textbf{s}, \textbf{s}' \in T_{\textbf{p}}\mathcal{M}, 
\end{equation}
Finally, let $\mu$ denote the \textit{density} on $\mathcal{M}$ uniquely determined by the Riemannian structure on $\mathcal{M}$ as in \cite[Prop.~16.45]{LeeBook2012}.  The density  determines a measure on $\mathcal{M}$, which we refer to as the \textit{uniform measure}. 
The measure enables integration of measurable functions $f : \mathcal{M} \rightarrow \mathbb{R}$, which we write as $ \int_{\mathcal{M}} f(\textbf{x})  d\mu(\textbf{x})$. 
Then the Riemannian \textit{volume} of $\mathcal{M}$ is  $\operatorname{vol}(\mathcal{M}) = \int_{\mathcal{M}} 1 d\mu(\textbf{x}).$

\subsection[\textbf{Preliminaries from  convex geometry}]{\textbf{Preliminaries from  convex geometry}}\label{subsec:convex-prelim}
We next give a quick reminder on general norms in finite-dimensional vector spaces, and their equivalence with certain convex bodies.  A few facts about tangent cones that we will need are also recorded.  
The only (possibly) novel content here is Proposition~\ref{prop:tan-con-boundary}. 
A nice textbook on convex geometry is \cite{convex-book}.

Let $\| \cdot \|: \mathbb{R}^D \rightarrow \mathbb{R}$ denote an arbitrary vector space \textit{norm} on $\mathbb{R}^D$.  This means:
\begin{itemize} \setlength\itemsep{0.4em}
\item $\| \textbf{v} \| \geq 0$ for all $\textbf{v} \in \mathbb{R}^D$; 
\item $\| \lambda \textbf{v} \| = | \lambda | \| \textbf{v} \| $ for all $\lambda \in \mathbb{R}$ and $\textbf{v} \in \mathbb{R}^D$; 
\item $\| \textbf{u} + \textbf{v} \| \leq \| \textbf{u} \| + \| \textbf{v} \|$ for all $\textbf{u}, \textbf{v} \in \mathbb{R}^D$. 
\end{itemize}
Recall that $\| \cdot \|$ is necessarily a continuous function on $\mathbb{R}^D$.
Also standard is that all norms on $\mathbb{R}^D$ are \textit{equivalent}, that is if $|\!|\!| \cdot |\!|\!|$ is another norm on $\mathbb{R}^D$, then there exist finite positive constants $c, C$ (depending only on $\| \cdot \|$, $|\!|\!| \cdot |\!|\!|$) such \nolinebreak that 
\begin{equation} \label{eq:equiv-norms}
    c |\!|\!| \textbf{v} |\!|\!|  \leq  \| \textbf{v} \|  \leq  C |\!|\!| \textbf{v}|\!|\!| \textup{ for all $\textbf{v} \in \mathbb{R}^D$}. 
\end{equation}
We write $\mathcal{B} \subseteq \mathbb{R}^D$ for the \textit{unit ball} with respect to the norm $\| \cdot \|$, 
\begin{equation} \label{eq:unit-ball}
    \mathcal{B} = \{ \textbf{v} \in \mathbb{R}^D : \| \textbf{v} \| \leq 1 \}.
\end{equation}
Then, $\mathcal{B}$ is a \textit{convex body} in $\mathbb{R}^D$. This means:  a compact convex subset of $\mathbb{R}^D$ with non-empty interior.  Furthermore, the unit ball is \textit{origin-symmetric},  i.e., $\textbf{v} \in \mathcal{B}$ implies $-\textbf{v} \in \mathcal{B}$ for all $\textbf{v} \in \mathbb{R}^D$.  Conversely, it is well-known that any origin-symmetric convex body in  $\mathbb{R}^D$ occurs as the unit ball for some norm on $\mathbb{R}^D$. 
Thus, there is a one-to-one correspondence  (see \cite[Chapter~2]{convex-book}):
\begin{equation*}
\{ \text{norms } \| \cdot \| \text{ on } \mathbb{R}^D \} \quad \longleftrightarrow  \quad \{\text{origin-symmetric convex bodies } \mathcal{B} \subseteq \mathbb{R}^D \}. 
\end{equation*} 
To emphasize this bijection, we shall let $\| \cdot \|_{\mathcal{B}}$ stand for the norm on $\mathbb{R}^D$ with unit ball $\mathcal{B}$ (except in the case of the $\ell_p$-norm where we write $\| \cdot \|_p$), i.e.,
\begin{equation*}
\| \cdot \|_{\mathcal{B}} \, \longleftrightarrow \, \mathcal{B}.
\end{equation*}
A few general topological remarks follow.  Given any subset $\mathcal{Y} \subseteq \mathbb{R}^D$.  
The (relative) \textit{topological boundary} of $\mathcal{Y}$ is the closure of $\mathcal{Y}$ minus the relative interior of $\mathcal{Y}$, written $\partial \mathcal{Y} := \overline{\mathcal{Y}} \, \setminus \, \operatorname{relint}(\mathcal{Y})$.
In the case of the unit ball \eqref{eq:unit-ball}, the boundary is the \textit{unit sphere}:
\begin{align*}
    \partial \mathcal{B} = \{ \textbf{v} \in \mathbb{R}^D : \| \textbf{v} \|_\B = 1\}.
\end{align*}
Given any point $\textbf{y} \in \mathcal{Y}$, the \textit{tangent cone} to $\mathcal{Y}$ at $\textup{\textbf{y}}$ is defined to be
\begin{align} \label{eq:def-tan-cone}
    TC_{\textbf{y}}(\mathcal{Y})
    :=
    \left\{  
         \!\textbf{d} \in \mathbb{R}^D \!:   \exists (\y_{k})_{k=1}^{\infty} \subseteq \mathcal{Y}, (\tau_k)_{k=1}^{\infty} \subseteq \mathbb{R}_{>0} \textup{ s.t. } \tau_k \rightarrow 0, \frac{\y_k - \y}{\tau_k} \rightarrow \textbf{d}\!     \right\}\!.
\end{align}
Note that, unlike abstract tangent spaces to manifolds, tangent cones to sets reside in $\mathbb{R}^D$ by definition.  We now give a few quick examples of tangent cones.
\begin{Example}[Tangent cones in familiar cases]
For a submanifold $\mathcal{Y} \subseteq \mathbb{R}^D$, the tangent cone and embedded tangent space always agree: $TC_{\textbf{y}}(\mathcal{Y}) = L_{\textbf{p}}(T_{\textbf{y}}(\mathcal{Y}))$.  
If $\mathcal{Y} = \{ (x_1, x_2) : x_2^2 = x_1^3 + x_1^2\} \subseteq \mathbb{R}^2$ is the nodal cubic plane curve, and $\textbf{y} = 0$ is the node, the tangent cone is the union of two lines: $TC_{\textbf{y}}(\mathcal{Y}) = \mathbb{R} \begin{pmatrix} 1 \\ 1 \end{pmatrix} \cup \mathbb{R} \begin{pmatrix} 1 \\ -1 \end{pmatrix}$.
If $\mathcal{Y} = \{(x_1, x_2) : x_2^2 = x_1^3\} \subseteq \mathbb{R}^2$ is the cuspidal cubic plane curve, and $\textbf{y} = 0$ is the cusp, the tangent cone is a half-line: $TC_{\textbf{y}}(\mathcal{Y}) = \mathbb{R}_{\geq 0} \begin{pmatrix} 1 \\ 0 \end{pmatrix}$.
Finally if $\mathcal{Y} = \{ \textbf{x} : \sum_{i=1}^D |x_i| \leq 1 \} \subseteq \mathbb{R}^D$ is the $\ell_1$ unit ball,  the tangent cone $TC_{\textbf{y}}(\mathcal{Y})$ is all of $\mathbb{R}^D$, a half-space in $\mathbb{R}^D$ or a polyhedron in $\mathbb{R}^D$ depending on whether $\textbf{y}$ lies in the interior of the unit ball, the relative interior of a facet of the unit sphere or elsewhere on the boundary.
\end{Example}

We will use the following easy and well-known facts about tangent cones.

\begin{Lemma} \label{lem:basic_tangent_cone}
\begin{enumerate}
\item \cite[Lem.~3.12]{nonlinear-optim-book} For all sets $\mathcal{Y} \subseteq \mathbb{R}^D$ and points $\textup{\textbf{y}} \in \mathcal{Y}$, we have 
$TC_{\textup{\textbf{y}}}(\mathcal{Y})$ is a closed cone. 
\item For all sets $\mathcal{Y} \subseteq \mathbb{R}^D$, points $\textup{\textbf{y}} \in \mathcal{Y}$ and linear subspaces $\mathcal{S} \subseteq \mathbb{R}^D$, we have  $TC_{\textup{\textbf{y}}}(\mathcal{Y}) \cap \mathcal{S} = TC_{\textup{\textbf{y}}}(\mathcal{Y} \cap \mathcal{S})$.
\item \cite[Lem.~3.13]{nonlinear-optim-book} For all convex sets $\mathcal{Y} \subseteq \mathbb{R}^D$ and points $\textup{\textbf{y}} \in \mathcal{Y}$, we have the explicit description (overline denotes closure in the Euclidean topology):
\begin{equation} \label{eq:nice-tan-cone}
    TC_{\textup{\textbf{y}}}(\mathcal{Y}) \, = \, \overline{\mathbb{R}_{\geq 0}\left(\mathcal{Y} - \textup{\textbf{y}}\right)} \, := \, \overline{\left\{ \beta (\widetilde{\textup{\textbf{y}}} - \textup{\textbf{y}}) \in \mathbb{R}^D: \, \beta \in \mathbb{R}_{>0}, \widetilde{\textup{\textbf{y}}} \in \mathcal{Y} \right\}}. 
\end{equation}
In particular, if $\mathcal{Y}$ is convex (respectively, convex with non-empty interior), then $TC_{\textup{\textbf{y}}}(\mathcal{Y})$ is convex (respectively, convex with non-empty interior). 
\end{enumerate}
\end{Lemma}

In light of the third item, we know all the possibilities in the plane for the tangent cone to convex sets with non-empty interior.

\begin{Example} \label{rem:coneR2}
The closed convex cones in $\mathbb{R}^2$ with non-empty interior are precisely  $\mathbb{R}^2$, closed half-spaces and the \textit{conical hull} of two linearly independent vectors: 
\begin{align*}
  \operatorname{coni}\{\textbf{d}_1, \textbf{d}_2\} := \{\beta_1 \textbf{d}_1 + \beta_2 \textbf{d}_2 : \beta_1, \beta_2 \in \mathbb{R}_{\geq 0} \}  \subseteq \mathbb{R}^2, \quad \textbf{d}_1, \textbf{d}_2 \in \mathbb{R}^2.
\end{align*}
In the latter case, the pair $\textbf{d}_1, \textbf{d}_2$ is unique up to positive scales, and one says that they generate the cone's \textit{extremal rays}, $\operatorname{coni}\{\textbf{d}_1\}$ and $\operatorname{coni}\{\textbf{d}_2\}$. 
\end{Example}
Finally, for technical purposes of the ``tilt construction" developed in the next section, we
need to observe that the topological boundary and tangent cone operations commute, at least in the case of our interest.
\begin{Lemma}
 \label{prop:tan-con-boundary}
For $\mathcal{B} \subseteq \mathbb{R}^D$ the unit ball of a norm $\| \cdot \|_\B$ and a boundary point $\textup{\textbf{y}} \in \partial \mathcal{B}$, the boundary of the tangent cone is the tangent cone of the boundary:
\begin{equation} \label{eq:annoying}
    \partial \left( TC_{\textup{\textbf{y}}}(\mathcal{B}) \right) \, = \, TC_{\textup{\textbf{y}}}(\partial \mathcal{B}).
\end{equation}
\end{Lemma}
We include a proof of Lemma~\ref{prop:tan-con-boundary} in Appendix~\ref{app:tan-con-boundary}, since we could not readily find this statement in the literature.

\subsection[\textbf{Tilt construction}]{\textbf{Tilt construction}} \label{subsec:tilt-const}

In this section, we present a construction 
that relates the second-order geometry of a submanifold $\mathcal{M} \subseteq \mathbb{R}^D$ around a point $\textbf{p} \in \mathcal{M}$ to  tangent cones to the unit sphere $\partial \mathcal{B} \subseteq \mathbb{R}^D$ of a norm $\| \cdot \|_{\mathcal{B}}$.
We name this construction the \textit{tilt function}, and denote it by $\tilt$.
Though not apparent initially, the relevance is that 
this function is required to define the limiting differential operator for point-cloud Laplacians formed by sampling $\mathcal{M}$ and computing affinities using $\| \cdot \|_{\mathcal{B}}$.
Specifically, it appears in the first-order derivative term \nolinebreak in \nolinebreak Eq.~\eqref{eq:def-DeltaMB}.

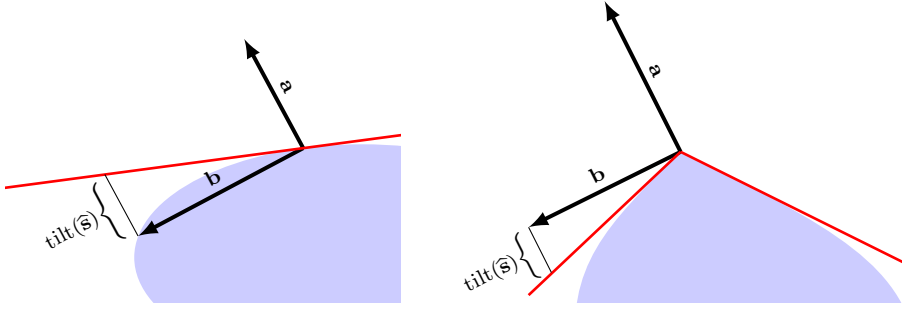
\begin{figure}[ht]
    \def\Cx{0}
    \def\Cy{-2}
    \def\Rx{3}
    \def\Ry{1.5}
    \def\angle{105}
    \def\EllipsePointX{\Cx+\Rx*cos(\angle)}
    \def\EllipsePointY{\Cy+\Ry*sin(\angle)}
    \def\TangentHalfLength{2}
    \begin{tikzpicture}[scale=1]
        \clip(-4.7,-2.6) rectangle (0.5,1);

        \draw [name path=ellipse,fill=blue!20,draw=none] ({\Cx},{\Cy}) ellipse ({\Rx} and {\Ry});

        \coordinate (p) at ({\EllipsePointX}, {\EllipsePointY}); 
        \coordinate (avector) at ({\EllipsePointX + \Rx*cos(\angle)},{\EllipsePointY + \Ry*sin(\angle)});
        \coordinate (bvector) at ({\EllipsePointX - 1.5*\Ry*sin(\angle)}, {\EllipsePointY + 1.5*\Rx*cos(\angle)});
        \draw[-latex,line width=1.5pt] (p) -- node[sloped, anchor=center, above] {${\bf a}$} (avector);
        \draw[-latex,line width=1.5pt] (p) -- node[sloped, anchor=center, above] {${\bf b}$} (bvector);

        \coordinate (tangentstart) at ({\EllipsePointX - \TangentHalfLength*\Rx*sin(\angle)}, {\EllipsePointY + \TangentHalfLength*\Ry*cos(\angle)});
        \coordinate (tangentend) at ({\EllipsePointX + \TangentHalfLength*\Rx*sin(\angle)}, {\EllipsePointY - \TangentHalfLength*\Ry*cos(\angle)});
        \draw[color=red,line width=1pt, name path={tangent}] (tangentstart) -- (tangentend);

        \path[name path={tilt}] (bvector) -- ({\EllipsePointX - 1.5*\Ry*sin(\angle) + 100*\Rx*cos(\angle)}, {\EllipsePointY
+1.5*\Rx*cos(\angle) + 100*\Ry*sin(\angle)});
        \draw [name intersections={of=tilt and tangent, by={intersect}}]
            (bvector) -- node[sloped,left,rotate=90]{{$ \text{tilt}(\widehat{\textbf{s}}) \Bigg\{$}} (intersect);
    \end{tikzpicture} 
    \quad
    \begin{tikzpicture}[scale=1]
        \clip(-5.3,2) rectangle (0.9,6);
        \fill[color=blue!20] (-1,-6) .. controls (-4,2) .. (-2,4) .. controls (2,2) ..   (-1,-6) -- (-3,-4);

        \draw[-latex,line width=1.5pt] (-2,4) -- node[sloped, anchor=center, above] {${\bf a}$} (-3,6);
        \draw[-latex,line width=1.5pt] (-2,4) -- node[sloped, anchor=center, above] {$\textbf{b}$} (-4,3);
    
        \draw[color=red,line width=1pt] (-2,4) -- (-4,2.1);
        \draw[color=red,line width=1pt] (-2,4) -- (1,2.5);

        \draw (-4,3) -- node[sloped,left,rotate=90] {$ \text{tilt}(\widehat{\textbf{s}}) \bigg\{$}(-3.7,2.4);
    \end{tikzpicture}
    \qquad
    \caption{\textit{Tilt construction}.  
   These diagrams take place in the 2D 
    linear subspace
   $\mathcal{S} := \textup{Span}\{\textbf{a}, \textbf{b}\} \subseteq \mathbb{R}^D$, where $\textbf{a}:=L_{\textbf{p}}(\widehat{\textbf{s}})$ and $\textbf{b}:=\tfrac12Q_{\textbf{p}}(\widehat{\textbf{s}})$ are tangent and normal vectors to $\mathcal{M}$ at $\textbf{p}$ respectively.
    Blue indicates $\widetilde{\mathcal{B}} = \mathcal{B} \cap \mathcal{S}$ (2D linear section of the unit ball $\mathcal{B}$).
    Red indicates the tangents to $\partial \widetilde{\mathcal{B}}$ at $\textbf{a}/\|\textbf{a}\|_{\mathcal{B}}$. 
     By definition, $\tilt(\widehat{\s})$ equals the signed $\ell_2$-length of the braced line segment.
    (left) An example where $TC_{\textbf{a}/\|\textbf{a}\|_{\mathcal{B}}}(\partial \widetilde{\mathcal{B}})$ consists of one well-defined tangent line.  Here tilt is positive; (right) An example where $TC_{\textbf{a}/\|\textbf{a}\|_{\mathcal{B}}}(\partial \widetilde{\mathcal{B}})$ consists of two tangent rays due to a singularity of $\partial \widetilde{\mathcal{B}}$ at $\textbf{a}/\|\textbf{a}\|_{\mathcal{B}}$.  Here tilt is negative.
   }\label{fig:understand-tilt}.  
\end{figure}

\begin{Propdef}[\normalfont Tilt function] \label{def:tilt-const}
    Let $\mathcal{M} \subseteq \mathbb{R}^D$ be a compact embedded Riemannian submanifold, let $\textup{\textbf{p}} \in \mathcal{M}$ be a point, and let $\widehat{\textup{\textbf{s}}} \in T_{\textup{\textbf{p}}}\mathcal{M}$ be a tangent vector to $\mathcal{M}$ at $\textup{\textbf{p}}$ with $\| \widehat{\textup{\textbf{s}}} \|_{2}=1$.  
     Following Eq.~\eqref{eq:expp-taylor}, consider the differential of the exponential map at $\textup{\textbf{p}}$ and the second fundamental form at $\textup{\textbf{p}}$ both evaluated at $\widehat{\textup{\textbf{s}}}$, and write
    \begin{align*}
        \textup{\textbf{a}} := L_{\textup{\textbf{p}}}(\widehat{\textup{\textbf{s}}}),
        \qquad
        \textup{\textbf{b}} := \frac{1}{2} Q_{\textup{\textbf{p}}}(\widehat{\textup{\textbf{s}}}).
    \end{align*}
    Further, let $\| \cdot \|_{\mathcal{B}}$ denote a norm on $\mathbb{R}^D$ with unit ball $\mathcal{B} \subseteq \mathbb{R}^D$ and unit sphere $\partial \mathcal{B} \subseteq \mathbb{R}^D$.
    Also write $TC$ to denote tangent cones as defined by Eq.~\eqref{eq:def-tan-cone}.
     
     Then, there exists a unique scalar $\eta \in \mathbb{R}$ such that 
\begin{equation} \label{eq:tilt-def}
    \frac{\textup{\textbf{b}}}{\|\textup{\textbf{a}}\|_{\mathcal{B}}^2} + \eta \textup{\textbf{a}} \, \in \, TC_{\textup{\textbf{a}}/\|\textup{\textbf{a}}\|_\mathcal{B}}(\partial
\mathcal{B}).
\end{equation}
We define the tilt function by
\begin{align}
    \tilt(\widehat{\s}) := \eta.
\end{align}
Hence 
$\tilt$ 
is a well-defined function from Euclidean-normalized tangent vectors to $\mathcal{M}$ at $\textup{\textbf{p}}$ into the real numbers.
\end{Propdef}

\begin{Remark}
In the course of proving Propostion/Definition~\ref{def:tilt-const} below, we shall show that the tilt function $\tilt(\widehat{\textbf{s}})$ only depends on the norm $\| \cdot \|_{\mathcal{B}}$ through the following (typically) two-dimensional central slice of the unit ball:
\begin{equation*}
    \operatorname{Span}\{L_{\textbf{p}}(\widehat{\textbf{s}}), Q_{\textbf{p}}(\widehat{\textbf{s}}) \} \cap \mathcal{B}.
\end{equation*}
This is two-dimensional origin-symmetric convex body (unless $Q_{\textbf{p}}(\widehat{\textbf{s}})=0$, in which case it is an origin-symmetric line segment).  
We make two remarks. 
First, as a consequence, we can visualize the tilt function using two-dimensional figures on the page (see Figure~\ref{fig:understand-tilt}).
Second,  in general, the central planar sections of the unit ball of a norm can vary significantly, and indeed qualitatively, across different slices.  For example, for the $\ell_1$-ball in $\mathbb{R}^3$, there is not a unique combinatorial type for a central planar section: instead either a quadrilateral or a hexagon can occur depending on the specific \nolinebreak slice.
\end{Remark}

\begin{proof}
In the defining equation~\eqref{eq:tilt-def} for $\tilt(\widehat{\textbf{s}})$, note that it is equivalent to require the left-hand side lies in the linear slice of the tangent cone:
\begin{align} \label{eq:simpler-tilt}
   &\mathcal{S} \cap TC_{\textbf{a}/\|\textbf{a}\|_{\mathcal{B}}}(\partial \mathcal{B})  \\[0.2em]  
  &\textup{where } \mathcal{S} :=  \operatorname{Span}\{L_{\textup{\textbf{p}}}(\widehat{\textbf{s}}), Q_{\textup{\textbf{p}}}(\widehat{\textbf{s}})\} \subseteq \mathbb{R}^D, \nonumber
\end{align}
since membership in the linear space $\mathcal{S}$ is guaranteed by definition.  

We shall rewrite the set \eqref{eq:simpler-tilt} using basic properties relating tangent cones, boundaries, and intersection by linear spaces.  
Firstly, we have
\begin{equation} \label{eq:tilt-proof-1}
    \mathcal{S} \cap TC_{\textbf{a}/\|\textbf{a}\|_{\mathcal{B}}}(\partial \mathcal{B}) 
    = TC_{\textbf{a}/\|\textbf{a}\|_{\mathcal{B}}}\left(\mathcal{S} \cap \partial \mathcal{B} \right) 
\end{equation}
by Lemma~\ref{lem:basic_tangent_cone}, item~2.  
Next, let
\begin{equation*}
\widetilde{\mathcal{B}} := \mathcal{S} \cap \mathcal{B} \subseteq \mathcal{S},
\end{equation*}
and note this is the unit ball of the restriction of the norm $\| \cdot \|_{\mathcal{B}}$ to the subspace $\mathcal{S}$, which is a norm in its own right on $\mathcal{S}$ (in our notation, $\| \cdot \|_{\widetilde{\mathcal{B}}}$).  Then,
\begin{equation*} 
    \mathcal{S} \cap \partial \mathcal{B} = \{ \textbf{t} \in \mathcal{S} : \| \textbf{t} \|_{\mathcal{B}} = 1 \} = \{ \textbf{t} \in \mathcal{S} : \| \textbf{t} \|_{\widetilde{\mathcal{B}}} = 1 \} = \partial \widetilde{\mathcal{B}},
\end{equation*}
from which it follows
\begin{equation} \label{eq:tilt-proof-2}
    TC_{\textbf{a}/\|\textbf{a}\|_{\mathcal{B}}}(\mathcal{S} \cap \partial \mathcal{B}) = TC_{\textbf{a}/\|\textbf{a}\|_{\mathcal{B}}}(\partial \widetilde{\mathcal{B}}).
\end{equation}
Now by Lemma~\ref{prop:tan-con-boundary},
\begin{equation} \label{eq:tilt-proof-3}
    TC_{\textbf{a}/\|\textbf{a}\|_{\mathcal{B}}}(\partial \widetilde{\mathcal{B}}) = \partial TC_{\textbf{a}/\|\textbf{a}\|_{\mathcal{B}}}(\widetilde{\mathcal{B}}).
\end{equation}
Combining Eq.~\eqref{eq:tilt-proof-1}, \eqref{eq:tilt-proof-2} and \eqref{eq:tilt-proof-3}, we get that
\begin{equation} \label{eq:tilt-proof-4}
    \mathcal{S} \cap TC_{\textbf{a}/\|\textbf{a}\|_{\mathcal{B}}}(\partial \mathcal{B}) = \partial TC_{\textbf{a} / \| \textbf{a} \|_{\mathcal{B}}}(\widetilde{\mathcal{B}}).
\end{equation}
The upshot is that in the defining equation for $\tilt$ it is equivalent to require membership in the right-hand side of \eqref{eq:tilt-proof-4}.

We shall now obtain a more explicit description of the set \eqref{eq:tilt-proof-4}.
Firstly, note that $\textbf{a} = L_{\textbf{p}}(\widehat{\textbf{s}}) \neq 0$, since $\| L_{\textbf{p}}(\widehat{\textbf{s}}) \|_{2} = \|\textbf{s}\|_2$ (Eq.~\eqref{eq:isometric}).
If $\textbf{b} = \frac{1}{2}Q_{\textbf{p}}(\widehat{\textbf{s}}) =0$, then the subspace $\mathcal{S}$ is one-dimensional.
In this case, the existence and uniqueness of $\eta$ in Eq.~\eqref{eq:tilt-def} is clear: 
$\widetilde{\mathcal{B}}$ is a line segment, 
$TC_{\textbf{a}/\|\textbf{a}\|_{\mathcal{B}}}(\widetilde{\mathcal{B}})$ is a ray (half-line), and its boundary $\partial TC_{\textbf{a}/\|\textbf{a}\|_{\mathcal{B}}}(\widetilde{\mathcal{B}})$ is the origin.  
Thus in the light of Eq.~\eqref{eq:tilt-proof-4}, we must take $\eta = 0$, so that $\tilt(\widehat{\textbf{s}}) = 0$ when the second fundamental form vanishes.
Therefore, assume $\textbf{b} \neq 0$.  
Since $\langle \textbf{a}, \textbf{b} \rangle = 0$ (Eq.~\eqref{eq:perp}), it follows that $\mathcal{S} \cong \mathbb{R}^2$ is two-dimensional and $\widetilde{\mathcal{B}}$ is a convex body in $\mathbb{R}^2$.
By Lemma~\ref{lem:basic_tangent_cone}, items 1 and 3, we know the tangent cone
$TC_{\textbf{a}/\|\textbf{a}\|_{\mathcal{B}}}$ is a closed convex cone in $\mathbb{R}^2$ with non-empty interior.  Then by Example~\ref{rem:coneR2}, the tangent cone is either all of $\mathbb{R}^2$, a half-plane in $\mathbb{R}^2$ or it is conically spanned by two linearly independent vectors.
We claim $TC_{\textbf{a}/\|\textbf{a}\|_{\mathcal{B}}}(\widetilde{\mathcal{B}}) \neq \mathbb{R}^2$.
Indeed since $\widetilde{\mathcal{B}}$ is convex and $\textbf{a}/\|\textbf{a}\|_{\mathcal{B}}$ lies in the boundary of $\widetilde{\mathcal{B}}$, the supporting hyperplane \nolinebreak theorem \nolinebreak implies:
    \begin{align*}
    \exists \, \textbf{v} \in \mathcal{S} \setminus \! \{0\} \,\,  \exists \, \gamma \in \mathbb{R} \, \textup{ s.t. } \langle \textbf{v}, \textbf{a}/\| \textbf{a} \|_\B \rangle = \gamma \,\, \wedge \,\, \forall \,\textbf{u} \in \widetilde{\mathcal{B}}, \,  \langle\textbf{v},  \textbf{u} \rangle \geq \gamma.
\end{align*}
Combining this with Eq.~\eqref{eq:nice-tan-cone}, it follows 
\begin{align*}
    TC_{\textbf{a}/\|\textbf{a}\|_\B}(\widetilde{\mathcal{B}}) 
    = 
    \overline{\mathbb{R}_{\geq 0}(\widetilde{\mathcal{B}} - \textbf{a}/\|\textbf{a}\|_{\mathcal{B}})} \subseteq \{\textbf{u} \in \mathcal{S} : \langle \textbf{v}, \textbf{u} \rangle \geq 0 \}.
    \end{align*}
In particular, $TC_{\textbf{a}/\|\textbf{a}\|_\B}(\widetilde{\mathcal{B}}) \neq \mathbb{R}^2$,
 so by Example~\ref{rem:coneR2} 
 the tangent cone is either a half-plane or conically spanned by two extremal rays.  For now, assume the latter case:
 there exist linearly independent vectors $\textbf{d}_1, \textbf{d}_2 \in \mathcal{S} \cong \mathbb{R}^2$ such that 
\begin{equation} \label{eq:tilt-proof-5}
    TC_{\textbf{a}/\|\textbf{a}\|_{\mathcal{B}}}(\widetilde{\mathcal{B}}) = \operatorname{coni}\{\textbf{d}_1, \textbf{d}_2\} \subseteq \mathbb{R}^2.
\end{equation}
The set \eqref{eq:tilt-proof-3} is thus the union of two rays:
\begin{equation} \label{eq:tilt-proof-membership}
    \partial TC_{\textbf{a}/\|\textbf{a}\|_{\mathcal{B}}}(\widetilde{\mathcal{B}}) = \mathbb{R}_{\geq 0} \textbf{d}_1 + \mathbb{R}_{\geq 2} \textbf{d}_2. 
\end{equation}

We shall now finish by proving the existence and uniqueness of $\eta \in \mathbb{R}$ such that
\begin{equation}
    \frac{\textbf{b}}{\|\textbf{a}\|_{\mathcal{B}}^2} + \eta \textbf{a} \,\, \in \,\, \mathbb{R}_{\geq 0} \textbf{d}_1  \, \cup \, \mathbb{R}_{\geq 0} \textbf{d}_2.
\end{equation}
To this end, first note that
\begin{align*}
-\textbf{a} \, = \, 0 - \|\textbf{a}\|_{\mathcal{B}}\left( \textbf{a} / \| \textbf{a} \|_{\mathcal{B}}\right) \, &\in \, \operatorname{relint}( \overline{\mathbb{R}_{\geq 0} (\widetilde{\mathcal{B}} - \textbf{a}/\|\textbf{a}\|_{\mathcal{B}})}) \\[0.2em] &= \, \operatorname{relint}(TC_{\textbf{a}/\|\textbf{a}\|_{\mathcal{B}}}(\widetilde{\mathcal{B}})) \\[0.2em]
&= \, \mathbb{R}_{>0} \textbf{d}_1 + \mathbb{R}_{>0} \textbf{d}_2.
\end{align*}
where the penultimate equality is again by  Eq.~\eqref{eq:nice-tan-cone} and the last equality is by Eq.~\eqref{eq:tilt-proof-5}.  Thus there exist positive scalars $\beta_1, \beta_2 \in \mathbb{R}_{>0}$ such that $-\textbf{a} = \beta_1 \textbf{d}_1 + \beta_2 \textbf{d}_2$.
Substituting this into  $\langle \textbf{a}, \textbf{b} \rangle = 0$ (Eq.~\eqref{eq:perp}), we get 
\begin{equation} \label{eq:tilt-proof-6}
\beta_1 \langle \textbf{d}_1, \textbf{b} \rangle + \beta_2 \langle \textbf{d}_2, \textbf{b} \rangle = 0.
\end{equation}
Since $\textbf{d}_1, \textbf{d}_2$ form a basis for $\mathcal{S}$ and $\textbf{b} \in \mathcal{S}$ and we are presently assuming $\textbf{b}\neq 0$, it cannot be that $\langle \textbf{d}_1, \textbf{b} \rangle = \langle \textbf{d}_2, \textbf{b} \rangle = 0$.
Instead, Eq.~\eqref{eq:tilt-proof-6} combined with $\beta_1, \beta_2 >0$ imply that exactly one of the inner products $\langle \textbf{d}_1, \textbf{b} \rangle, \langle \textbf{d}_2, \textbf{b} \rangle$ is strictly positive while the other is strictly negative.  
Relabeling if necessary, we can assume that $\langle \textbf{d}_1, \textbf{b} \rangle > 0 > \langle \textbf{d}_2, \textbf{b} \rangle$.  With this in hand, let us examine the membership \eqref{eq:tilt-proof-membership}.

Notice that for each $\eta \in \mathbb{R}$, it holds 
\begin{equation} \label{eq:tilt-proof-7}
    \frac{\textbf{b}}{\| \textbf{a} \|_{\mathcal{B}}} + \eta \textbf{a} \, \notin \, \mathbb{R}_{\geq 0} \textbf{d}_2.
\end{equation}
This is by taking inner products with $\textbf{b}$:  all vectors on the right-hand side of \eqref{eq:tilt-proof-7} have a non-positive inner product with $\textbf{b}$ using $\langle \textbf{d}_2, \textbf{b} \rangle < 0$.  
Meanwhile on the left-hand side of \eqref{eq:tilt-proof-7}, we have 
$\langle \textbf{b}, \frac{\textbf{b}}{\| \textbf{a} \|_{\mathcal{B}}} + \eta \textbf{a} \rangle = \frac{\|\textbf{b}\|_2^2}{\| \textbf{a} \|_{\mathcal{B}}} > 0$ (the equality is from $\langle \textbf{a}, \textbf{b} \rangle = 0$ and the strict inequality is from the  assumption $\textbf{b} \neq 0$).

On the other hand, there do exist scalars $\eta \in \mathbb{R}$ and $\beta \in \mathbb{R}_{\geq 0}$ satisfying
\begin{align} \label{eq:tilt-proof-8}
    \frac{\textbf{b}}{\|\textbf{a}\|_\B^2} + \eta \textbf{a} =\beta \textbf{d}_1.
\end{align}
Indeed using that $\textbf{a}, \textbf{b}$ are an orthogonal basis for $\mathcal{S}$, $\| \textbf{a} \|_2^2 = 1$ and $\textbf{d}_1 \in \mathcal{S}$, \nolinebreak note
\begin{align}\label{eq:expand-d1}
    \textbf{d}_1 = \langle \textbf{d}_1, \textbf{a} \rangle \textbf{a} + \frac{\langle \textbf{d}_1, \textbf{b}  \rangle}{\| \textbf{b} \|_2^2}  \textbf{b}.
\end{align}
Then substituting Eq.~\eqref{eq:expand-d1} into \eqref{eq:tilt-proof-8} and equating  coefficients, we compute the following \textit{unique} solution to Eq.~\eqref{eq:tilt-proof-8}:
\begin{align}
   & \beta  =  \|\textbf{b}\|_2^2  \Big{/} \! \left( \|\textbf{a}\|_\B^2 \langle \textbf{d}_1, \textbf{b} \rangle \right), \nonumber \\[1pt]
   & \eta  = \| \textbf{b} \|_2^2 \langle  \textbf{d}_1, \textbf{a} \rangle  \Big{/} \!\! \left( \| \textbf{a} \|_\B^2 \langle \textbf{d}_1, \textbf{b} \rangle  \right). \label{eq:crazy-eta}
\end{align}

This completes the case when $TC_{\textbf{a}/\| \textbf{a} \|_{\mathcal{B}}}(\widetilde{\mathcal{B}})$ is conically spanned by independent vectors.  
As for the third case afforded by Example~\ref{rem:coneR2}, when the tangent cone is a half-plane, let the boundary of the half-plane be spanned by $\textbf{d}_1 \in \mathcal{S} \cong \mathbb{R}^2$.  
Again we arrive at Eq.~\eqref{eq:tilt-proof-8} but without the constraint that $\beta \geq 0$.
Solving as before, $\eta$ is uniquely determined and given by Eq.~\eqref{eq:crazy-eta}.

This completes the proof that $\eta$ exists and is unique.  In sum: if $Q_{\textbf{p}}(\widehat{\textbf{s}})=0$ then $\tilt(\widehat{\textbf{s}}) =0$, and otherwise $\tilt(\widehat{\textbf{s}})$ is given by Eq.~\eqref{eq:crazy-eta}.  \hfill \qed
\end{proof}

In the next statement, we assume the local continuous differentiability of the norm to get a more explicit expression for the tilt function.
The proof is in Appendix~\ref{app:simple-title-C1}.

\begin{Proposition}[\normalfont{Simplifications for tilt in the case of $C^1$-norm}] \label{prop:simple-title-C1}
Regard the norm $\| \cdot \|_{\mathcal{B}}$ as a function from $\mathbb{R}^D$ to $\mathbb{R}$.
\begin{enumerate}
\item Let $\widehat{\textup{\textbf{a}}}$ be a point in $\mathbb{R}^D$ with $\| \widehat{\textup{\textbf{a}}} \|_{\mathcal{B}} = 1$. 
If $\| \cdot \|_{\mathcal{B}}$ is continuously differentiable in a neighborhood of $\widehat{\textup{\textbf{a}}}$, then 
the tangent cone to the $\| \cdot \|_{\mathcal{B}}$-unit sphere at $\widehat{\textup{\textbf{a}}}$ is the hyperplane:
\begin{align}\label{eq:simpler-TC}
    TC_{\widehat{\textup{\textbf{a}}}}(\partial \mathcal{B}) = \left\{ \textup{\textbf{v}} \in \mathbb{R}^D : \left\langle \textup{\textbf{v}}, \, \operatorname{grad}\| \cdot \|_{\mathcal{B}} (\widehat{\textup{\textbf{a}}})\right\rangle = 0 \right\}.
\end{align}
\item Assume the setup of Proposition/Definition~\ref{def:tilt-const}, and further that $\| \cdot \|_{\mathcal{B}}$ is continuously differentiable in a neighborhood of the point $L_{\p}({\widehat{\s}})$.  Then, the tilt function equals
\begin{align}
    \tilt(\widehat{\s}) = \frac{-\left\langle \operatorname{grad} \| \cdot \|_{\mathcal{B}} (L_{\p}(\widehat{\s})), \, \tfrac{1}{2} Q_{\p}(\widehat{\s}) \right\rangle}{\left\langle \operatorname{grad} \| \cdot \|_{\mathcal{B}} (L_{\p}(\widehat{\s})), \, L_{\p}(\widehat{\s})  \right\rangle} \, \frac{1}{\left\| L_{\p}(\widehat{\s}) \right\|_{\mathcal{B}}^2}.
\end{align}
\end{enumerate}
\end{Proposition}

\subsection[\textbf{Laplacian-like operator and main theorem statement}]{\textbf{Laplacian-like operator and main theorem statement}} \label{subsec:thm-statement}

Here we state our main result  
 in Theorem~\ref{thm:limit}.  We first need to define  both sides of Eq.~\eqref{eq:main-limit}, in particular the differential operator $\Delta_{\mathcal{M}, \mathcal{B}}$ (Definition~\ref{def:Laplace-like}).
For simplicity, we consider only the standard \textup{Gaussian kernel} with width $\sigma$:
\begin{align*}
K_{\sigma} : \mathbb{R}_{\geq 0} \rightarrow \mathbb{R}_{> 0} \quad \textup{ defined by } \quad
K_{\sigma}(t) = \textup{exp}(-t^2/\sigma^2).
\end{align*}

\begin{Definition}[\normalfont Point-cloud  Laplacian with respect to an arbitrary norm] \label{discrete_laplacian_B}
Let $\| \cdot \|_\B$ be a norm on $\mathbb{R}^D$.
Let $\mathcal{X} = \{ \x_1, \ldots, \x_n \} \subseteq \mathbb{R}^D$ be a set of points.
Then we define the \textit{point-cloud Laplacian} computed using the norm 
$\| \cdot \|_\B$ and the point set 
$\mathcal{X} \subseteq \mathbb{R}^D$ to act on functions whose domains contain $\mathcal{X}$ as follows: for each 
$f : \mathcal{Y} \rightarrow \mathbb{R}$ where $\mathcal{X} \subseteq \mathcal{Y}$, 
define $\mathcal{L}_{n, \mathcal{B}}f : \mathcal{Y} \rightarrow \mathbb{R}$ by
\begin{equation} \label{eq:discrete-Lap-1}
    \mathcal{L}_{n, \B} f(\p) := \frac{1}{n} \sum_{i=1}^n K_{\sigma_n}(\| {\bf x}^{(i)} - {\bf p} \|_\B)(f({\bf x}^{(i)}) - f({\bf p})).
\end{equation}
\end{Definition}
Compare Eq.~\eqref{eq:discrete-Lap} with Eq.~\eqref{eq:discrete-Lap-1}. 
In what follows, $d\textup{\textbf{s}}$ is the Lebesgue measure on $T_{\textup{\textbf{p}}}\mathcal{M}$, and $d\widehat{\textup{\textbf{s}}}$ is the uniform measure on the sphere $\{\widehat{\textup{\textbf{s}}} \in T_{\textup{\textbf{p}}}\mathcal{M} : \|\widehat{\textup{\textbf{s}}}\|_2 =1 \}$.

\begin{Definition}[\normalfont Laplacian-like differential operator with respect to an arbitrary norm] \label{def:Laplace-like}
The \textit{Laplacian-like operator} on a submanifold $\mathcal{M} \subseteq \mathbb{R}^D$ with respect to a norm $\| \cdot \|_\B$ is defined to act on functions $f : \mathcal{M} \rightarrow \mathbb{R}$ according to
\begin{align} \label{eq:def-DeltaMB}
    (\Delta_{\mathcal{M}, \mathcal{B}} f)(\textbf{p})
    :=
    &\left\langle (\operatorname{hess} \widetilde{f})(0) , \, \tfrac{1}{2} \int_{\{\textup{\textbf{s}} \in T_{\textup{\textbf{p}}}\mathcal{M}: \| L_{\textup{\textbf{p}}}(\textup{\textbf{s}}) \|_\B \leq 1\}} \textup{\textbf{s}} \textup{\textbf{s}}^{\top} d \textup{\textbf{s}} \right\rangle \nonumber \\[2pt]
    & +\left\langle (\operatorname{grad} \widetilde{f})(0), \,  \int_{\{\widehat{\textup{\textbf{s}}} \in T_{\textup{\textbf{p}}}\mathcal{M}     : \| \widehat{\textup{\textbf{s}}} \|_2 =1 \}} \widehat{\textup{\textbf{s}}} \, \|L_{\textup{\textbf{p}}}(\widehat{\textup{\textbf{s}}})\|_\B^{-d}     \tilt(\widehat{\textup{\textbf{s}}}) \, d\widehat{\textup{\textbf{s}}}     \right\rangle
\end{align} 
where $\widetilde{f} = f \circ \textup{exp}_{\textup{\textbf{p}}} : T_{{\bf p}}\mathcal{M} \rightarrow \mathbb{R}$ and
$L_{\textup{\textbf{p}}} = D\textup{exp}_{\textup{\textbf{p}}}(0): T_{\textup{\textbf{p}}}\mathcal{M} \rightarrow \mathbb{R}^D$.
\end{Definition}

\begin{Remark}[\normalfont{Extrinsic interpretation of the integration domains and integrands in Eq.~\eqref{eq:def-DeltaMB}}] \label{rem:extrinsic-nice}
Both domains of integration in Definition~\ref{def:Laplace-like} are subsets of the abstract tangent space  $T_{\textbf{p}}\mathcal{M}$, since we have written the integrals in parameterized form.
Using the isometry $L_{\textbf{p}}$, we can 
identify the first domain with the $d$-dimensional intersection of the embedded tangent space and the unit ball:
\begin{align*}
    L_{\textbf{p}}(T_{\textbf{p}}\mathcal{M}) \cap \mathcal{B},
\end{align*}
and the first integral in Eq.~\eqref{eq:def-DeltaMB} with the second-moment of this convex body:
\begin{align*}
    \int_{\textbf{t} \in L_{\textbf{p}}(T_{\textbf{p}}\mathcal{M}) \cap \mathcal{B}} \textbf{t} \textbf{t}^{\top} d\textbf{t}.
\end{align*}
Meanwhile, under the mapping $s \mapsto L_{\textbf{p}}(\textbf{s})/\|L_{\textbf{p}}(\textbf{s})\|_{\mathcal{B}}$, the second domain of integration in Definition \ref{def:Laplace-like} identifies with the $(d-1)$-dimensional intersection of the embedded tangent with the unit sphere:
\begin{align*}
    L_{\textbf{p}}(T_{\textbf{p}}\mathcal{M}) \cap \partial \mathcal{B},
\end{align*}
and the second integral in Eq.~\eqref{eq:def-DeltaMB} is a weighted first-moment of this boundary. 
\end{Remark}

\begin{Theorem}[\normalfont Main result: Convergence of the point-cloud Laplacian based on an arbitrary norm] \label{thm:limit}
Let $\| \cdot \|_\B$ be a norm on $\mathbb{R}^D$ with unit ball $\mathcal{B}$.
Let 
$\mathcal{M}$ be a compact $d$-dimensional embedded Riemannian submanifold of $\mathbb{R}^D$.
Let $\x_1, \ldots, \x_n$ be i.i.d. draws from the uniform measure on $\mathcal{M}$.
Fix any constant $\alpha>0$, and set  $\sigma_n = n^{-1/(2d + 4 + \alpha)}$ and 
 $c_n := \Gamma(\frac{d+4}{2}) \sigma_n^{d+2}$.
Then given a three-times continuously differentiable function $f: \mathcal{M} \rightarrow \mathbb{R}$ and a point $\textup{\textbf{p}} \in \mathcal{M}$, we have the following almost sure convergence:
\begin{align}  \label{eq:main-limit} \frac{1}{c_n}
\mathcal{L}_{n, \mathcal{B}}f(\textup{\textbf{p}}) \xrightarrow{\,\,\,\,\, \textup{a.s.} \,\,\,\,} \frac{1}{\textup{vol}(\mathcal{M})} \Delta_{\mathcal{M}, \mathcal{B}} f(\textup{\textbf{p}}).
\end{align} 
\end{Theorem}

\noindent Theorem~\ref{thm:limit} is proved in Section~\ref{sec:main-proof}. 

\begin{Remark}[Comparing $\Delta_{\mathcal{M}, \mathcal{B}}$ and $\Delta_{\mathcal{M}}$]
Note two key features that distinguish the continuum limit for general norms from the Laplace-Beltrami operator:
\begin{itemize}\setlength\itemsep{0.5em}
    \item The operator $\Delta_{\mathcal{M}, \mathcal{B}}$ is typically \textbf{extrinsic}.  By Remark~\ref{rem:extrinsic-nice}, both terms in \eqref{eq:def-DeltaMB} vary with the orientation of the embedded tangent space $L_{\textbf{p}}(T_{\textbf{p}}\mathcal{M}) \subseteq \mathbb{R}^D$ in relation to the ball $\mathcal{B} \subseteq \mathbb{R}^D$.  (For a concrete example, see Section~\ref{subsec:circle-example}.) 
    \item The operator $\Delta_{\mathcal{M},\mathcal{B}}$ has a \textbf{first-order derivative term}.
\end{itemize}
The Euclidean norm is special on both counts. 
However despite the added complexity of general norms, there can be \textbf{practical advantages} to using $\Delta_{\mathcal{M}, \mathcal{B}}$ over $\Delta_{\mathcal{M}}$.  
At least this can hold for a well-chosen norm, when 
reducing the dimension of certain data sets.
We illustrate this numerically in Section~\ref{sec:experiments}.  
\end{Remark}

We finish this section with an easy extension of Theorem~\ref{thm:limit} to the case where the  sampling of $\mathcal{M}$ is non-uniform.  
\begin{Theorem}[\normalfont{Convergence of the point-cloud Laplacian based on an arbitrary norm with non-uniform sampling}] \label{thm:nonuniform}
 Assume the same setup as Theorem~\textup{\ref{thm:limit}} above, except  $\textup{\textbf{x}}_1, \ldots, \textup{\textbf{x}}_n$ are i.i.d. draws from a probability distribution on $\mathcal{M}$ described by a $C^3$ probability density function, $dP(\textup{\textbf{x}}) = P(\textup{\textbf{x}}) d\mu(\textup{\textbf{x}})$.   
 Then, the almost sure limit of the LHS of \eqref{eq:main-limit} exists and equals $\Delta_{\mathcal{M}, \mathcal{B}, P} f(\p)$, where 
 \begin{equation}
     \Delta_{\mathcal{M}, \mathcal{B}, P} := \operatorname{vol}(\mathcal{M})P \Delta_{\mathcal{M}, \mathcal{B}} + \delta_{\mathcal{M}, \mathcal{B}, P}.
 \end{equation}
 Here  $\delta_{\mathcal{M}, \mathcal{B}, P}$ only modifies the first-order derivative term, and is defined by
 \begin{align} \label{eq:additional-term}
        (\delta_{\mathcal{M},\mathcal{B}, P} f)(\p) & := \left\langle  ( \operatorname{grad} \widetilde{f}(0) ) (\operatorname{grad} \widetilde{P}(0))^{\top} , \int_{\{\textup{\textbf{s}} \in T_{\textup{\textbf{p}}}\mathcal{M}: \| L_{\textup{\textbf{p}}}(\textup{\textbf{s}}) \|_\B \leq 1\}} \textup{\textbf{s}} \textup{\textbf{s}}^{\top} d \textup{\textbf{s}} \right\rangle \nonumber  \\
        & = \int_{\{\textup{\textbf{s}} \in T_{\textup{\textbf{p}}}\mathcal{M}: \| L_{\textup{\textbf{p}}}(\textup{\textbf{s}}) \|_\B \leq 1\}} \langle \operatorname{grad} \widetilde{f}(0), \s \rangle \langle \operatorname{grad} \widetilde{P}(0), \s \rangle d\s,
    \end{align}
where $\widetilde{f} = f \circ \operatorname{exp}_{\p}$ and $\widetilde{P} = P \circ \operatorname{exp}_{\p}$.    
\end{Theorem}
\begin{proof} 
Using the reduction in \textup{\cite[Sec.~5]{BelkinNiyogi2008}} and then Theorem~\ref{thm:limit}, 
the LHS of \eqref{eq:main-limit} tends to $\Delta_{\mathcal{M}, \mathcal{B}} h(\p)$ where $h : \mathcal{M} \rightarrow \mathbb{R}$ is defined by $h(\textup{\textbf{x}}) := \left( f(\textup{\textbf{x}}) - f(\textup{\textbf{p}}) \right)P(\textup{\textbf{x}})$.
Let $\widetilde{h} = h \circ \operatorname{exp}_{\p}$, so  $\widetilde{h} = \widetilde{f} \, \widetilde{P} - \widetilde{f}(0) \widetilde{P}$.  
Then $\operatorname{grad} \widetilde{h} (0) = \widetilde{P}(0) \operatorname{grad} \widetilde{f}(0)$ and
$\operatorname{hess} \widetilde{h}(0) = \widetilde{P}(0) \operatorname{hess} \widetilde{f}(0) + ( \operatorname{grad} \widetilde{f}(0) ) (\operatorname{grad} \widetilde{P}(0))^{\top} + ( \operatorname{grad} \widetilde{P}(0) ) (\operatorname{grad} \widetilde{f}(0))^{\top}$.
Inserting these formulas into Definition~\ref{def:Laplace-like} and rearranging gives the result. \hfill \qed
\end{proof}

\subsection[First properties of the Laplacian-like differential operator]{\textbf{First properties of $\Delta_{\mathcal{M}, \mathcal{B}}$}} \label{sec:first_properties}
We give a few basic properties of the limit in Theorem~\ref{thm:limit}. 
Firstly, it is elliptic.

\begin{Lemma}
[\normalfont{$\Delta_{\mathcal{M}, \mathcal{B}}$ is elliptic}] \label{lem:unif-elliptic}
For all compact embedded Riemannian submanifolds $\mathcal{M} \subseteq \mathbb{R}^D$ and all  norms $\| \cdot \|_{\mathcal{B}}$ on $\mathbb{R}^D$, the Laplacian-like operator $\Delta_{\mathcal{M}, \mathcal{B}}$ is a uniformly elliptic differential operator on $\mathcal{M}$. 
\end{Lemma}
The proof of this lemma is in Appendix~\ref{app:unif-elliptic}.

Next, we investigate the regularity properties of the coefficients of $\Delta_{\mathcal{M}, \mathcal{B}}$. 
Surprisingly, the first-order coefficient function need not even be continuous everywhere. 
Below,  $T\mathcal{M} = \bigsqcup_{\textbf{p} \in \mathcal{M}} T_{\textbf{p}} \mathcal{M}$ is the \textup{tangent bundle} of $\mathcal{M}$,  and $\operatorname{Sym}^2(T\mathcal{M}) = \bigsqcup_{\textbf{p} \in \mathcal{M}} \operatorname{Sym}^2(T_{\textbf{p}} \mathcal{M})$ is its \textup{symmetric square} bundle  \cite[Ch.~10]{LeeBook2012}.

\begin{Proposition}
[\normalfont{Continuity properties of $\Delta_{\mathcal{M}, \mathcal{B}}$}] \label{prop:continuity-properties}
For all compact embedded Riemannian submanifolds $\mathcal{M} \subseteq \mathbb{R}^D$ and all norms $\| \cdot \|_{\mathcal{B}}$ on $\mathbb{R}^D$, the Laplacian-like operator $\Delta_{\mathcal{M}, \mathcal{B}}$ has the following continuity properties.  
\begin{enumerate}
\item As a section of $\operatorname{Sym}^2(T \mathcal{M})$, the coefficient of the second-order term, 
\begin{align}\label{eq:cont-pf-state-1}
     \frac{1}{2} \int_{\s \in T_{\p}\mathcal{M} : \| L_{\p}(\s)\|_{\mathcal{B}} \leq 1} \s \s^{\top} d\s,
\end{align}
is continuous at all points $\p \in \mathcal{M}$.
\item As a section of $T \mathcal{M}$, the coefficient of the first-order term,
\begin{align}\label{eq:cont-pf-state-2}
     \int_{\widehat{\s} \in T_{\p}\mathcal{M} : \|\widehat{\s}\|_2=1} \widehat{\s} \| L_{\p}(\widehat{\s}) \|^{-d}_{\mathcal{B}} \tilt(\widehat{\s}) d\widehat{\s},
\end{align}
is continuous at all points $\p \in \mathcal{M}$ such that the norm $\| \cdot \|_{\mathcal{B}}$ is continuously differentiable in a neighborhood of $L_{\p}(T_{\p}\mathcal{M}) \cap \mathbb{S}^{D-1}$.  The first-order coefficient can be discontinuous at other points $\p \in \mathcal{M}$.
\end{enumerate}
\end{Proposition}
The proof of the second item relies on the expression for the tilt function in Proposition~\ref{prop:simple-title-C1},  where  the norm is locally continuously differentiable.  Details are in Appendix~\ref{app:continuity-properties}.

\subsection[\textbf{Example: any manifold, Euclidean norm}]{\textbf{Example: any manifold, Euclidean norm}} \label{sec:example_euclidean}

Let us first check that Theorem~\ref{thm:limit} agrees with the standard Euclidean theory.  Let $\mathcal{M} \subseteq \mathbb{R}^D$ be any $d$-dimensional compact smooth embedded Riemannian manifold, $\textbf{p} \in \mathcal{M}$, and consider the \textup{Euclidean norm} $\| \cdot \|_2$ with Euclidean unit ball $\mathcal{B} = \{\x: \|\x\|_2 \le 1 \} \subseteq \mathbb{R}^D$.

We first argue that $\tilt \equiv 0$.  Let $\widehat{\textbf{s}} \in T_{\textbf{p}}\mathcal{M}$ with $\| \widehat{\textbf{s}} \|_2 =1$.  Set $\textbf{a} = L_{\textbf{p}}(\widehat{\textbf{s}})$, $\textbf{b} = \frac{1}{2} Q_{\textbf{p}}(\widehat{\textbf{s}})$ and $\mathcal{S} = \textup{Span}\{\textbf{a}, \textbf{b}\}$.  If $\textbf{b} = 0$, then $\tilt(\widehat{\textbf{s}}) = 0$.  Else, put $\widetilde{\mathcal{B}} := \mathcal{B} \cap \mathcal{S}$.  By construction, $\tilt(\widehat{\textbf{s}}) = \eta$ for $\eta \in \mathbb{R}$  uniquely determined \nolinebreak by
\begin{align*}
    \textbf{b} + \eta \textbf{a} \in TC_{\textbf{a}}(\partial \widetilde{\mathcal{B}}),
\end{align*}
using $\| \textbf{a} \|_2 = 1$ since $L_{\textbf{p}}$ is an isometry (Eq. \eqref{eq:isometric}).
However, $\widetilde{\mathcal{B}}$ is a Euclidean unit disk in $\mathcal{S} \cong \mathbb{R}^2$, and $\partial \widetilde{\mathcal{B}}$ is a Euclidean unit circle in $\mathbb{R}^2$.
So, $TC_{\textbf{a}}(\partial \widetilde{\mathcal{B}})$ is the orthogonal complement of $\mathbb{R} \textbf{a}$ inside $\mathcal{S}$. This gives $TC_{\textbf{a}}(\partial \widetilde{\mathcal{B}}) = \mathbb{R} \textbf{b}$ since $Q_{\textbf{p}}$ takes values in the normal space (Eq. \eqref{eq:perp}). Clearly then, $\eta = 0$ and $\tilt \equiv 0$. We have verified that the first-order term vanishes in $\Delta_{\M, \mathcal{B}}$.

As for the second-order term, we compute the following second moment:
\begin{align}
 \int_{\{\textup{\textbf{s}} \in T_{\textup{\textbf{p}}}\mathcal{M}: \| L_{\textup{\textbf{p}}}(\textup{\textbf{s}}) \|_2 \leq 1\}} \textup{\textbf{s}} \textup{\textbf{s}}^{\top} d \textup{\textbf{s}} 
 & = \int_{\{\textup{\textbf{s}} \in T_{\textup{\textbf{p}}}\mathcal{M}: \| \textup{\textbf{s}} \|_2 \leq 1\}} \textup{\textbf{s}} \textup{\textbf{s}}^{\top} d \textup{\textbf{s}}  \quad & \!\! [\textup{by Eq.}~\eqref{eq:isometric}]\nonumber \\
 & = \left( \int_{\{\textup{\textbf{s}} \in T_{\textup{\textbf{p}}}\mathcal{M}: \| \textup{\textbf{s}} \|_2 \leq 1\}} s_1^2 d\textbf{s} \! \right) \! I_{d}  & \!\![\textup{oddness, symmetry}] \nonumber \\
 & \propto I_d.  \label{eq:prop-constant}
\end{align}
Here the second equality used that the integration domain $\{\textup{\textbf{s}} \in T_{\textup{\textbf{p}}}\mathcal{M}: \| \textup{\textbf{s}} \|_2 \leq 1\}$ is preserved under sign flips of individual coordinates of $\textbf{s}$, hence the off-diagonal terms $s_i s_j$ ($i \neq j$) integrate to 0.

Plugging into Definition~\ref{def:Laplace-like}, we obtain
\begin{align*}
    \Delta_{\mathcal{M}, \mathcal{B}}f(\textbf{p})
   & \propto
    \left\langle \operatorname{hess} \widetilde{f}(0), I_d \right\rangle \\
    &= \textup{trace}\left({\text{hess} \widetilde{f}}(0) \right) \\
   & =\Delta_{\mathcal{M}}f(\textbf{p}).
\end{align*}
This is the usual Laplace-Beltrami operator on $\mathcal{M}$ applied to $f$ and evaluated at $\p$.
Thus, we have checked that Theorem~\ref{thm:limit} indeed recovers Theorem~\ref{thm:classical}.

\begin{Remark} 
Due to the lack of normalization in Definition~\ref{def:Laplace-like}, our Laplacian-like operator  $\Delta_{\mathcal{M},\mathcal{B}}$ in the Euclidean case differs from the usual Laplace-Beltrami operator $\Delta_{\mathcal{M}}$ by a multiplicative constant.  From Eq.~\eqref{eq:prop-constant}, the constant is 
\begin{align*}
& \frac{1}{2}\int_{\{\textbf{s} \in T_{\textbf{p}} \mathcal{M} : \| \textbf{s} \|_2 \leq 1\}} s_1^2 d\textbf{s} \,\, = \,\, \frac{1}{2d} \int_{\{\textbf{s} \in T_{\textbf{p}} \mathcal{M} : \| \textbf{s} \|_2 \leq 1\}} \| \textbf{s} \|_2^2 d\textbf{s} \\[0.5em]
& = \frac{1}{2d}\int_{0}^{1} r^{d+1}\operatorname{vol}(\mathbb{S}^{d-1}) dr \,\, = \,\, \frac{1}{2d} \int_{0}^{1} r^{d+1} \frac{2 \pi^{d/2}}{\Gamma(\frac{d}{2})} dr \,\, = \,\, \frac{\pi^{d/2}}{4\Gamma(\frac{d+4}{2})}.
\end{align*}
This is simply the ratio of prefactors in the scales $c_n$ in Theorems~\ref{thm:classical} and \ref{thm:limit}.
\end{Remark}

\subsection[\textbf{Example: circle in the plane, weighted Manhattan norm}]{\textbf{Example: circle in the plane, weighted $\ell_1$-norm}} \label{subsec:circle-example}
Next, we look at a non-Euclidean example in full detail. 
Consider the (Euclidean) unit circle in $\mathbb{R}^2$ where the ambient norm is a weighted $\ell_1$-norm.
That is, let $\mathcal{M} = S^1 = \{\textbf{x} = (x_1, x_2)^{\top} \in \mathbb{R}^2 : x_1^2 + x_2^2 = 1 \}$ and use the norm $\| \cdot \|_{\textbf{w},1}$ defined by $\| \textbf{x} \|_{\textbf{w},1} = w_1 |x_1| + w_2 |x_2|$  where $\textbf{w} = (w_1, w_2)^{\top} \in \left(\mathbb{R}_{>0}\right)^2$. The unit ball of $\| \cdot \|_{\textbf{w},1}$ is the region
\begin{align*}
    \mathcal{B} = \left\{ \textbf{x} = (x_1, x_2)^{\top} \in \mathbb{R}^2 : w_1 |x_1| + w_2 |x_2| \leq 1 \right\},
\end{align*}
while the unit sphere of $\| \cdot \|_{\textbf{w},1}$ is
\begin{align*}
    \partial \mathcal{B} = \left\{ \textbf{x} = (x_1, x_2)^{\top} \in \mathbb{R}^2 : w_1 |x_1| + w_2 |x_2| = 1 \right\},
\end{align*}
a rhombus with vertices $(\pm (1/w_1), 0), (0, \pm (1/w_2))$.  
Let $\textbf{p} = (\textup{cos} (\theta), \textup{sin} (\theta))^{\top} \in S^1$.
Parameterize $T_{\textbf{p}}S^1$ (with respect to a fixed unit basis vector) using $\psi \in \mathbb{R}$.  The exponential map is
\begin{align*}
\textup{exp}_{\textbf{p}} : T_{\textbf{p}}S^1 \rightarrow S^1, \quad \psi \mapsto \left(\textup{cos}(\theta + \psi), \textup{sin}(\theta + \psi) \right)^{\top}.
\end{align*}
The differential of this is 
\begin{align*}
    L_{\textbf{p}}(\psi) = \psi \left(-\textup{sin}(\theta), \textup{cos}(\theta) \right)^{\top}\!\!.
\end{align*}
The second fundamental form is 
\begin{align*}
    Q_{\textbf{p}}(\psi) = -\psi^2\left(\textup{cos}(\theta), \textup{sin}(\theta) \right)^{\top}\!\!.
\end{align*}

We take on the terms in the limiting operator in Definition~\ref{def:Laplace-like} in turn.  For the second-order term, we need the second moment of a line segment:
\begin{align*}
  \tfrac{1}{2}  \int_{\{\psi : \|L_{\textbf{p}}(\psi)\|_{\textbf{w},1} \leq 1\}} \psi^2 d\psi &= \tfrac{1}{2} \int_{|\psi| \, \leq \, \| (-\textup{sin}(\theta), \textup{cos}(\theta))^{\top} \|_{\textbf{w},1}^{-1}} \psi^2 d\psi \nonumber \\[3.5pt]
    &= \frac{1}{3\left( w_1 | \textup{sin}(\theta)| + w_2 |\textup{cos}(\theta)|\right)^3}.
\end{align*}
As for the first-order coefficient, this becomes a sum of over the two endpoints of the line segment.  We shall show the first-order coefficient equals
\begin{equation} \label{eq:toughy}
    \textup{sign}(\textup{cos}(\theta)\textup{sin}(\theta)) \frac{-w_1 | \textup{cos}(\theta) | + w_2 |\textup{sin}(\theta)|}{(w_1|\textup{sin}(\theta)| + w_2 |\textup{cos}(\theta)|)^4},
\end{equation}
where $\textup{sign} : \mathbb{R} \rightarrow \{-1,0,1\}$ is given by $\textup{sign}(t) := 1$ if $t>0$; $\textup{sign}(t) := -1$ if $t < 0$; and $\textup{sign}(0) := 0$. By the symmetry of the rhombus $\partial\mathcal{B}$ with respect to individual coordinate sign flips in $\mathbb{R}^2$, one easily sees  formula \eqref{eq:toughy} is correct for $\theta$ an integer multiple of $\frac{\pi}{2}$ (the first-order coefficient is zero). Otherwise, we may reduce to verifying correctness of the expression \eqref{eq:toughy} when $\theta \in (0, \frac{\pi}{2})$.  
Then in this case, it is enough to show
\begin{equation} \label{eq:toughy-2}
    \tilt(1) = \frac{1}{2} \frac{-w_1  \textup{cos}(\theta)  + w_2 \textup{sin}(\theta)}{(w_1 \textup{sin}(\theta) + w_2 \textup{cos}(\theta))^3}.
\end{equation}

To this end, let $\alpha \in (0, \frac{\pi}{2})$ be half the angle $\partial \mathcal{B}$ makes at $(\frac{1}{w_1}, 0)^{\top}$\!, so $\textup{tan}(\alpha) = w_1/w_2$.  
Let $\textbf{a} = L_{\textbf{p}}(1)$ and $\textbf{b} = \frac{1}{2} Q_{\textbf{p}}(1)$.
Let $\omega$ be the signed angle at $\textbf{a} / \|\textbf{a}\|_{\textbf{w},1} \in \partial \mathcal{B}$ from 
$\textbf{a} / \|\textbf{a}\|_{\textbf{w},1} \, + \, \mathbb{R}_{\geq 0} (\frac{-1}{w_1}, \frac{-1}{w_2})^{\top}$ to $\textbf{a} / \|\textbf{a}\|_{\textbf{w},1} \, + \, \mathbb{R}_{\geq 0} \textbf{b}$, where counterclockwise counts as positive.
By elementary angle chasing,
\begin{align*}
    \omega = \theta - \alpha.
\end{align*}
Thus (see Figure~\ref{fig:understand-tilt}),
\begin{align*}
    \tilt(1)
    &=
    \left\| \frac{\textbf{b}}{\|\textbf{a} \|_{\textbf{w},1}^2} \right\|_{2} \! \textup{tan}(\omega) = \frac{1}{2 \|\textbf{a} \|_{\textbf{w},1}^2} \textup{tan}(\theta - \alpha)
    \\
    &=
    \frac{1}{2\left( w_1 \textup{sin}(\theta) + w_2 \textup{cos}(\theta) \right)^2} \frac{\textup{tan}(\theta) - (w_1/w_2)}{1 + \textup{tan}(\theta) (w_1/w_2)},
\end{align*}
which indeed simplifies to Eq.~\eqref{eq:toughy-2}.

Summarizing: for each angle $\theta \in [0,2\pi]$, Theorem~\ref{thm:limit} implies the Laplacian-like differential operator  $\Delta_{\mathcal{M}, \mathcal{B}}$ is given by
 \begin{equation} \label{eq:limit-op-circle}
    \textup{sign}(\cos \theta \, \sin \theta) \, \frac{- w_1 |\cos \theta|  + w_2 |\sin \theta|  }{\left(w_1 |\sin \theta|  + w_2 |\cos \theta| \right)^4} \, \frac{d}{d \theta} \,\, + \,\, \frac{1}{3 \left(w_1 |\sin \theta|  + w_2 |\cos \theta| \right)^3} \, \frac{d^2}{d \theta^2}.
\end{equation}

\begin{figure} 
    \includegraphics[width=\linewidth]{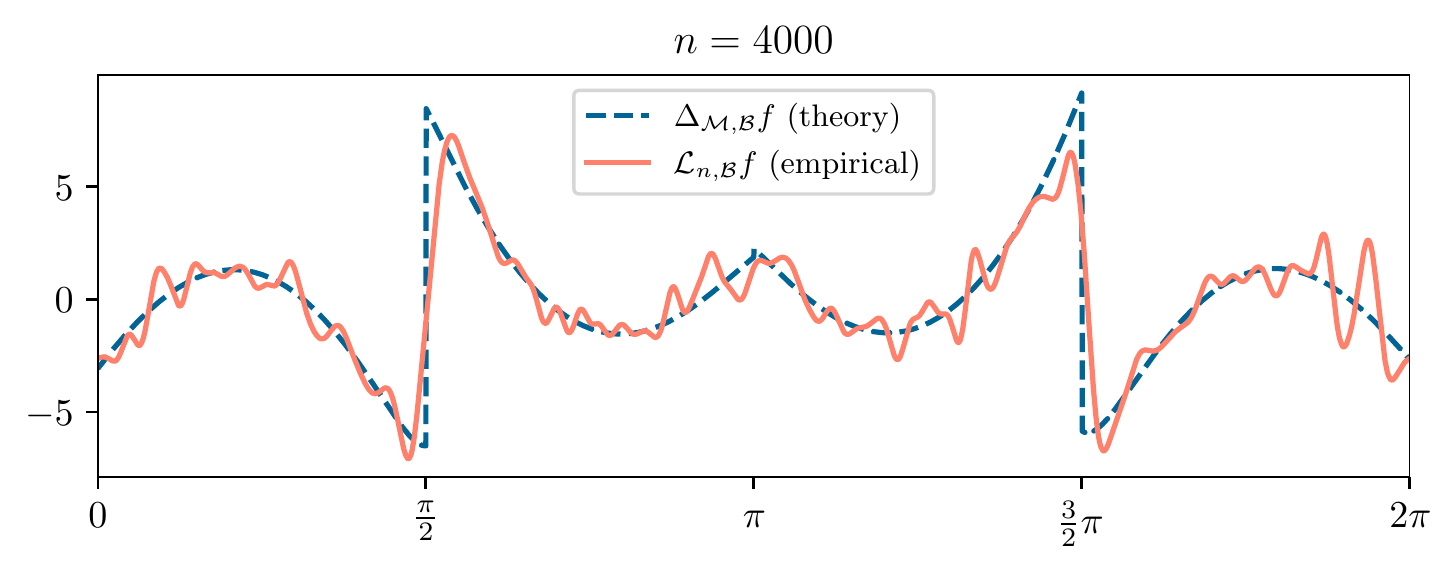}
    \includegraphics[width=\linewidth]{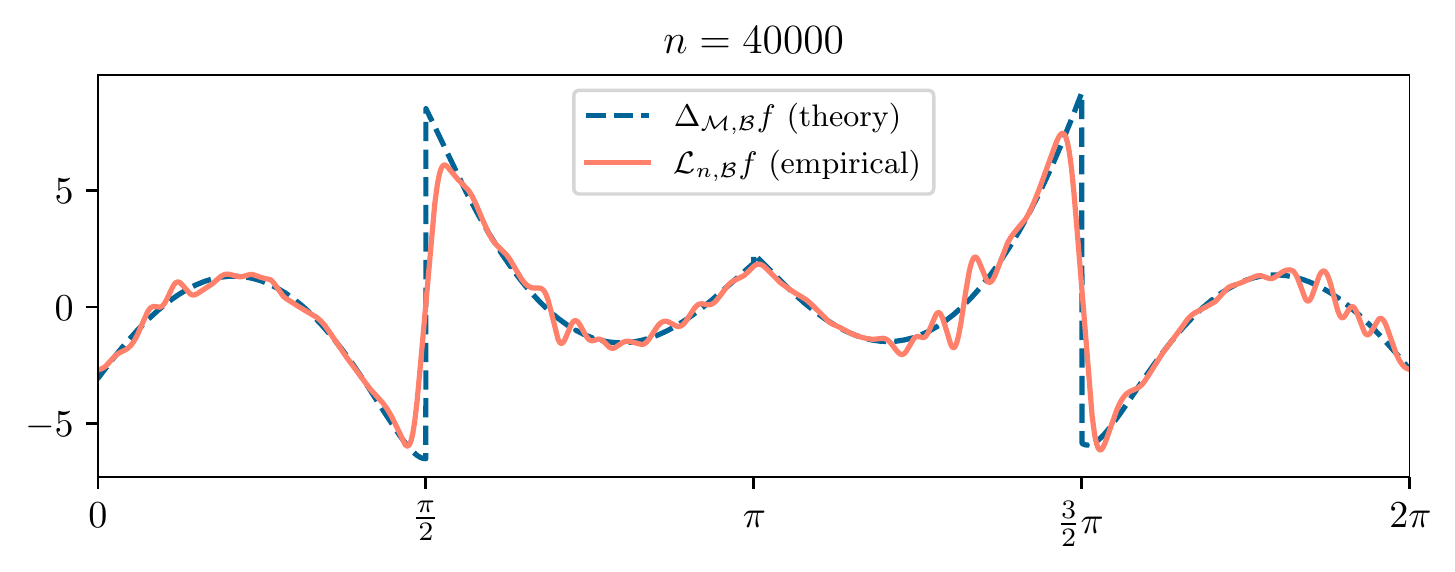}
    \caption{\textit{Empirical vs. theoretical weighted $\ell_1$ Laplacian on the circle} ($w_1 = 1, w_2 = 1.5$) applied to the function $f(\theta) = \sin(\theta) + \cos(2 \theta) + \cos(5\theta)$. For the empirical Laplacian, the samples were drawn from the uniform distribution on the unit circle. (top panel) $n = 4,000$ samples; (bottom panel) $n = 40,000$ samples.  Here $\mathcal{L}_{n, \mathcal{B}}f$ is scaled by $\operatorname{vol}(S^1)/(\Gamma(\tfrac{d+4}{2})\sigma_n^{d+2})$.} \label{fig:empirical-v-theoretical}
\end{figure}

As an independent numerical verification of this formula, we performed the following experiment.
We fixed a particular function 
$f : S^1 \rightarrow \mathbb{R}$ (namely a certain trigonometric polynomial). 
We drew $n$ points uniformly i.i.d from the circle.
We computed the empirical point-cloud Laplacian applied to $f$, using Eq.~\eqref{eq:discrete-Lap-1} and evaluating  $\mathcal{L}_{n,\mathcal{B}} f$ along a dense regular grid.
For comparison, we evaluated the Laplacian-like operator applied to $f$,  using Eq.~\eqref{eq:limit-op-circle} and evaluating $\Delta_{\mathcal{M}, \mathcal{B}}f$ along the same grid.
Figure~\ref{fig:empirical-v-theoretical} shows a convincing match: as the sample size $n$ grows, the empirical and theoretical plots match up increasingly well.
Appendix~\ref{sec:numerical_eigenfunction_computation} presents numerical results on the \textup{eigenfunctions} of
 \eqref{eq:limit-op-circle}.

\begin{Remark} \label{rem:discontinuous}
The coefficient of $\frac{d}{d \theta}$ in Eq.~\eqref{eq:limit-op-circle} is discontinuous
at $\theta = 0, \frac{\pi}{2}, \pi, \frac{3\pi}{2}$. Thus, this example confirms the second sentence of  Proposition~\ref{prop:continuity-properties}, item 2.
\end{Remark}

\section{\textbf{Proof of Theorem \ref{thm:limit}}} \label{sec:main-proof}
To improve readability, we split the proof of Theorem~\ref{thm:limit} into several steps.
First, we reduce to the population limit ($n=\infty$), replacing sums by integrals, via concentration of measure.
The integrals are then parameterized by geodesic normal coordinates on the manifold. Both of these are standard steps in the analysis of empirical Laplacians based on the $\ell_2$-norm.
We then  
replace the Gaussian kernel by the 0/1 kernel, so all considerations are local. 
The domain of integration becomes the intersection of the manifold $\mathcal{M}$ with the convex body $\sigma \mathcal{B}$ (for $\sigma \rightarrow 0$).
This being a potentially unwieldy domain, we substitute the Taylor series expansion of the exponential map 
to replace the manifold $\mathcal{M}$ by first- and second-order approximations around $\textbf{p}$.
For the term involving the second fundamental form, we switch to spherical coordinates.
Then we study the radial domain of integration.   
We consider this step (Section~\ref{subsec:boundary}) to be the proof's most technical. 
Following this analysis, the tilt function emerges  (Proposition~\ref{prop:obtain-tilt}), and dominated convergence is used to finish.

\subsection[Step 1: reduce to the population limit]{\textbf{Step~1: reduce to the population limit ($n \to \infty$)}}

This is a standard application of concentration of measure.
Let 
\begin{align*}
    S_n^{(i)}
    &:=
    \operatorname{vol}(\mathcal{M})\frac{1}{nc_n}  K_{\sigma_n}(\| \x_i - {\bf p} \|_\B)(f(\x_i) - f({\bf p})),\\
    \quad S_n
    &:= (\operatorname{vol}(\mathcal{M})/c_n) \mathcal{L}_{n,\B} f(\p)
    =
    \sum_{i=1}^n S_n^{(i)}.
\end{align*}
    \noindent For a fixed sample size $n$,  the values $S_n^{(1)}, \ldots, S_n^{(1)}$  are i.i.d. random variables.
    By the continuity of $f$ and compactness of $\mathcal{M}$, there is a constant 
    $C_0 > 0$  such that  $|f(\textbf{x})| \leq c_0$  for all  $\mathbf{x} \in \mathcal{M}$. 
    Recalling $K_{\sigma_n} \le 1$, it follows that 
    \begin{equation} \label{eq:Sin-bound}
        |S_n^{(i)}| \leq (2C_0\operatorname{vol}(\mathcal{M}))/(nc_n).
    \end{equation}
    Let $\epsilon > 0$.  By inequality~\eqref{eq:Sin-bound} together with Hoeffding's inequality,
   \begin{equation} \label{eq:hoeff}
        \mathbb{P}\left( \lvert S_n - \mathbb{E}[S_n] \rvert \geq \frac{\epsilon}{2} \right)  \! \leq \! 2  \textup{exp}\left( \frac{-\epsilon^2 n c_n^2}{32C_0^2\operatorname{vol}(\mathcal{M})^2} \right) \! = \!  2   \textup{exp}\left( \frac{-\epsilon^2 \Gamma(\frac{d+4}{2}) n^{\frac{\alpha}{2d+4+\alpha}}}{32C_0^2\operatorname{vol}(\mathcal{M})^2} \right)
    \end{equation}
where we substituted $c_n = \Gamma(\frac{d+4}{2}) \sigma_n^{d+2}$ and $\sigma_n = n^{-1/(2d+4+\alpha)}$.  Here
\begin{align*}
    \mathbb{E}[S_n] \,\, = \,\, \frac{1}{\Gamma(\frac{d+4}{2}) \sigma_n^{d+2}} \int_{\textbf{x} \in \mathcal{M}} K_{\sigma_n}(\| \textbf{x} - \textbf{p} \|_\B) \left(f(\textbf{x}) - f(\textbf{p})\right)  d\mu(\textbf{x}), 
\end{align*}
where $d\mu$ is the Riemannian volume density on $\mathcal{M}$.
Since $\sigma_n \rightarrow 0$ as $n \rightarrow \infty$, \textbf{assuming we proved}
\begin{equation} \label{eq:pop-limit-gamma}
   \lim_{\sigma \rightarrow 0} \frac{1}{\Gamma(\frac{d+4}{2}) \sigma^{d+2}} \int_{\textbf{x} \in \mathcal{M}} K_{\sigma}(\| \textbf{x} - \textbf{p} \|_\B) \left(f(\textbf{x}) - f(\textbf{p})\right)  d\mu(\textbf{x}) \, = \,  \Delta_{\mathcal{M},\mathcal{B}} f(\textbf{p}),
\end{equation}
   \textbf{\textup{then it would follow}} there exists $n_0 = n_0(\epsilon)$ such that for all $n > n_0$,
 \begin{equation} \label{eq:if-would}
     \bigl\lvert \mathbb{E}[S_n] - \Delta_{\mathcal{M}, \mathcal{B}}f(\textbf{p})  \bigr\rvert \leq \frac{\epsilon}{2}.
 \end{equation}
Combining inequalities \eqref{eq:hoeff} and  \eqref{eq:if-would} gives, for all $n > n_0$,
 \begin{align} \label{eq:almost-there}
     \mathbb{P}(\bigl\lvert S_n - \Delta_{\mathcal{M}, \mathcal{B}}f(\p) f(\textbf{p})\bigr\rvert \geq \epsilon)  & \leq  \mathbb{P}(\bigl\lvert S_n - \mathbb{E}[S_n]\bigr\rvert  \geq  \tfrac{\epsilon}{2}) \\[0.4em] & \leq 2  \textup{exp}\!\left( \frac{-\epsilon^2 \Gamma(\tfrac{d+4}{2}) n^{\frac{\alpha}{2d+4+\alpha}}}{32C_0^2\operatorname{vol}(\mathcal{M})^2} \right)\!.
 \end{align}
By $\alpha > 0$, the RHS of \eqref{eq:almost-there} converges to $0$ as $n \rightarrow \infty$.  
Since $\epsilon$ was arbitrary, this shows that $S_n$ converges to $\Delta_{\M,\B}f(\textbf{p})$ in probability.  Dividing by $\operatorname{vol}(\M)$ gives $\mathcal{L}_{n,\B}f(\textbf{p}) \xrightarrow{\, p \,} (1/\operatorname{vol}(\mathcal{M})) \Delta_{\mathcal{M}, \mathcal{B}}f(\p)$.
We can upgrade this to almost sure convergence, simply by noting that the series
\[
\sum_{n=1}^{\infty} 2 \, \textup{exp}\!\left( \frac{-\epsilon^2 \Gamma(\tfrac{d+4}{2}) n^{\frac{\alpha}{2d+4+\alpha}}}{32C_0^2 \operatorname{vol}(\mathcal{M})^2} \right)\!
\]
converges and citing the Borel-Cantelli lemma.
\textup{It remains to actually prove~\eqref{eq:pop-limit-gamma}}.

\subsection[\textbf{Step 2: reduce to the indicator function kernel}]{\textbf{Step~2: reduce to the indicator function kernel}} \label{subsec:switch-indicator}
In this step, we replace the Gaussian kernel $K_{\sigma_n}$ by the indicator function kernel $\mathbb{1}_{\sigma_n}$, defined by
\begin{align*}
   \mathbb{1}_{\sigma_n} : \mathbb{R}_{\geq 0} \rightarrow \mathbb{R}_{\geq 0} \textup{ where }  \mathbb{1}_{\sigma_n}(t) := 1 \textup{ if } t \in [0, \sigma_n] \textup{ and } \mathbb{1}_{\sigma_n}(t) := 0 \textup{ if } t > \sigma_n.
\end{align*}
Precisely, we show 
\begin{equation} \label{eq:pop-limit}
   \lim_{\sigma \rightarrow 0} \frac{1}{\sigma^{d+2}} \int_{\textbf{x} \in \mathcal{M}} \! \mathbb{1}_{\sigma}(\| \textbf{x} - \textbf{p} \|_\B) \left(f(\textbf{x}) - f(\textbf{p})\right)  d\mu(\textbf{x}) \, = \,  \Delta_{\M,\B}f(\textbf{p})
\end{equation}
implies the required formula \eqref{eq:pop-limit-gamma}, and thereby we will reduce to proving \eqref{eq:pop-limit}.

To achieve this reduction, we now assume \eqref{eq:pop-limit}.  Write the Gaussian kernel $K_{\sigma}$ as a sum of indicator functions: 
\begin{equation} \label{eq:kappa-def}
    K_{\sigma}(t) = \int_{s=0}^{\infty} \kappa_{\sigma}(s) \mathbb{1}_{s}(t) ds = \int_{s=t}^{\infty} \kappa_{\sigma}(s) ds,
\end{equation}
where $\kappa_{\sigma} : \mathbb{R}_{\geq 0} \rightarrow \mathbb{R}_{\geq 0}$ is given by $\kappa_{\sigma}(s) := (2s/\sigma^2) \textup{exp}(-s^2/\sigma^2)$.  Then,
\begin{align}
   & \int_{\textbf{x} \in \mathcal{M}} K_{\sigma}(\| \textbf{x} - \textbf{p}\|_\B) (f(\textbf{x}) - f(\textbf{p})) d\mu(\textbf{x}) & \nonumber \\
   &=  \int_{\textbf{x} \in \mathcal{M}} \left( \int_{\textbf{s} = \| \textbf{x} - \textbf{p} \|_\B}^{\infty} \kappa_{\sigma}(s) ds \right) (f(\textbf{x}) - f(\textbf{p})) d\mu(\textbf{x}) \,\, \quad &[\textup{substituting } \eqref{eq:kappa-def}]\nonumber \\
   &= \int_{s=0}^{\infty} \kappa_{\sigma}(s) \left( \int_{\textbf{x} \in \mathcal{M} : \| \textbf{x} - \textbf{p} \|_\B \leq s} f(\textbf{x}) - f(\textbf{p}) d\mu(\textbf{x}) \right)\!ds \,\, \quad &[\textup{Fubini's theorem}]. \label{eq:fubini} 
\end{align}

Define 
\begin{align*}
    e(s) :=  \left(\int_{\textbf{x} \in \mathcal{M} : \| \textbf{x} - \textbf{p} \|_\B \leq s} f(\textbf{x}) - f(\textbf{p}) d\mu(\textbf{x})\right) \, - \, s^{d+2} \Delta_{\M,\B}f(\textbf{p}).
\end{align*}
In light of \eqref{eq:pop-limit}, we have
\begin{equation} \label{eq:little-o}
    e(s) = o(s^{d+2}) \,\, \textup{ as } s \rightarrow 0.
\end{equation}

Fix $\epsilon > 0$.  By \eqref{eq:little-o}, we can fix $\delta > 0$ such that
\begin{equation} \label{eq:little-o-explicit}
0 \leq s \leq \delta \, \Rightarrow \, |e(s)| \leq \epsilon s^{d+2}.
\end{equation}
Returning to Eq.~\eqref{eq:fubini}, we may change the upper limit of integration with the following control on the approximation error:
\begin{align} \label{eq:drop-to-delta}
    & \int_{s=0}^{\infty} \kappa_{\sigma}(s) \left( \int_{\textbf{x} \in \mathcal{M} : \| \textbf{x} - \textbf{p} \|_\B \leq s} f(\textbf{x}) - f(\textbf{p}) d\mu(\textbf{x}) \right)ds \\
    & = \int_{s=0}^{\delta} \kappa_{\sigma}(s) \left( \int_{\textbf{x} \in \mathcal{M} : \| \textbf{x} - \textbf{p} \|_\B \leq s} f(\textbf{x}) - f(\textbf{p}) d\mu(\textbf{x}) \right)ds \,\, + \,\, \textup{exp}(-\delta^2 / \sigma^2)\textup{poly}(\sigma). \nonumber
\end{align}
To justify Eq.~\eqref{eq:drop-to-delta} holds, we note the parenthesized integral has absolute value bounded by $2c_0 $ for all $s \in [0, \infty]$, by the compactness of $\mathcal{M}$. Thus, a tail bound for the Gaussian kernel implies \eqref{eq:drop-to-delta} (set $k=0$ in Eq.~\eqref{eq:gaussian-even-tail} in Appendix~\ref{app:gaussian}).
Now the main term in Eq.~\eqref{eq:drop-to-delta} is
\begin{align*}
    \int_{s=0}^{\delta} \kappa_{\sigma}(s) \big{(}s^{d+2} \Delta_{\M,\B}f(\textbf{p}) \, + \, e(s) \big{)} ds.
\end{align*}
From \eqref{eq:little-o-explicit}, this is bounded above by
\begin{equation} \label{eq:plus-eps}
    \int_{s=0}^{\delta} \kappa_{\sigma}(s) s^{d+2} \left( \Delta_{\M,\B}f(\textbf{p}) + \epsilon\right)  ds, 
\end{equation}
and below by
\begin{equation} \label{eq:neg-eps}
    \int_{s=0}^{\delta} \kappa_{\sigma}(s) s^{d+2} \left( \Delta_{\M,\B}f(\textbf{p}) - \epsilon\right)  ds.
\end{equation}
By additional bounds for the Gaussian (Appendix~\ref{app:gaussian}), the upper and lower bounds \eqref{eq:plus-eps} and \eqref{eq:neg-eps} are equal to
\begin{align*}
   \int_{s=0}^{\infty} \kappa_{\sigma}(s) s^{d+2} \left( \Delta_{\M,\B}f(\textbf{p}) \pm \epsilon \right) ds \, + \, \textup{exp}(-\delta^2/\sigma^2)\textup{poly}(\sigma).
\end{align*}
But, the main term is 
\begin{align*}
    \sigma^{d+2} \Gamma(\tfrac{d+4}{2}) \left( \Delta_{\M,\B}f(\textbf{p}) \pm \epsilon  \right)\!,
\end{align*}
by the formula for half the $k$-th absolute moment of $\kappa_{\sigma}$  (Eq.~\eqref{eq:gaussian-central-moment}, Appendix~\ref{app:gaussian}).  Using $\lim_{\sigma \rightarrow 0} \textup{exp}(-\delta^2/\sigma^2) \textup{poly}(\sigma) = 0$, and the fact that $\epsilon$ is arbitrary, we achieve what we wanted:
\begin{align*} 
\int_{\textbf{x} \in \mathcal{M}} \lim_{\sigma \rightarrow 0} \frac{1}{\sigma^{d+2}} K_{\sigma}(\| \textbf{x} - \textbf{p}\|_\B) (f(\textbf{x}) - f(\textbf{p})) d\mu(\textbf{x}) \,\, = \,\, \Gamma(\tfrac{d+4}{2}) \Delta_{\M,\B}f(\textbf{p}).
\end{align*}

To sum up, Eq.~\eqref{eq:pop-limit} implies Eq.~\eqref{eq:pop-limit-gamma}. It remains to prove Eq.~\eqref{eq:pop-limit}.

\subsection[\textbf{Step 3: use geodesic normal coordinates and Taylor expand}]{\textbf{Step~3: use geodesic normal coordinates and Taylor expand}}

In this step, we express the integral in the LHS of \eqref{eq:pop-limit} in normal coordinates,
\begin{equation} \label{eq:key-int-again}
    \int_{\textbf{x} \in \mathcal{M} : \| \textbf{x} - \textbf{p} \|_\B \leq \sigma} f(\textbf{x}) - f(\textbf{p}) d\mu(\textbf{x}).
\end{equation}
We parameterize it using the exponential map (Section~\ref{subsec:prelim-riem}),
\begin{align*} 
     \textup{exp}_{{\bf p}}:  U \xrightarrow{\sim} V, 
\end{align*}
where $U \subseteq T_{\textbf{p}}\mathcal{M}$ and $V \subseteq \mathcal{M}$ are open neighborhoods of $0$ and $\textbf{p}$ respectively.
Note that there exists some constant $\sigma_0 > 0$ such that for all $\sigma \leq \sigma_0$ the domain of integration in \eqref{eq:key-int-again}  is contained in $V$,
\begin{equation} \label{eq:exp-UV}
    \{\textbf{x} \in \mathcal{M} : \| \textbf{x} - \textbf{p} \|_\B \leq \sigma \} \subseteq V.
\end{equation}
This follows from the fact that $\mathcal{M}$ is an embedded submanifold of $\mathbb{R}^D$, hence 
$V$ can be written as an open set of $\mathbb{R}^D$ intersected with $\mathcal{M}$,
and the fact $\| \cdot \|_{\mathcal{B}}$ is equivalent to the Euclidean norm on $\mathbb{R}^D$ and so induces the same open sets.
 Therefore, by a change of variables, for each $\sigma \leq \sigma_0$, the integral \eqref{eq:key-int-again} equals
\begin{equation} \label{eq:switch-geo}
    \int_{\textbf{s} \in U : \| \textup{exp}_{\textbf{p}}(\textbf{s}) - \textbf{p}\|_\B \leq \sigma} \left( \widetilde{f}(\textbf{s}) - \widetilde{f}(0) \right) \bigl\lvert \textup{det} D \textup{exp}_{\textbf{p}}(\textbf{s}) \bigr\rvert d\textbf{s},
\end{equation}
where $\textbf{s} = (s_1, \ldots, s_d)^{\top}$ denotes coordinates for $T_{\textbf{p}}\mathcal{M}$ with respect
to an orthonormal basis and $d\textbf{s}$ denotes the Lebesgue measure on $(T_{\textbf{p}}\mathcal{M}, \langle \cdot , \cdot \rangle_{\textbf{p}})$.

Our goal is to approximate the integral \eqref{eq:switch-geo} up to order $\sigma^{d+2}$. To this end, we will consider three Taylor expansions:
\begin{align}
    & \widetilde{f}(\textbf{s}) \, = \, \widetilde{f}(0)  +  \text{grad}\widetilde{f}(0)^{\top} \textbf{s}  +  \tfrac{1}{2} \textbf{s}^{\top} \text{hess} \widetilde{f}(0) \textbf{s}  +  O(\| \textbf{s} \|_2^3), \label{eq:taylor-f}\\[0.8pt]
    & \textup{det}  D\textup{exp}_{\textbf{p}}(\textbf{s}) \, = \, 1  -  \tfrac{1}{6} \textbf{s}^{\top} \textup{Ric}(\textbf{p})\textbf{s}  +  O(\| \textbf{s} \|_2^3) \, = \, 1 + O(\| \textbf{s} \|_2^2), \label{eq:taylor-Dexp}\\[0.8pt]
    & \textup{exp}_{\textbf{p}}(\textbf{s}) \,\, = \,\, \textbf{p} \, + \, L_{\textbf{p}}(\textbf{s}) \, + \, \tfrac{1}{2} Q_{\textbf{p}}(\textbf{s}) \, + \, O(\| \textbf{s} \|_2^3). \label{eq:taylor-exp}
\end{align}
Here 
$\textup{Ric}(\textbf{p}) \in \mathbb{R}^{d \times d}$ stands for the \textup{Ricci curvature} of $\mathcal{M}$ at $\textbf{p}$ (see \cite[Ch.~7]{Lee-Riem-Book}).  Also, see Section~\ref{subsec:prelim-riem} for discussion on $L_{\textbf{p}}$ and $Q_{\textbf{p}}$. 

Substituting equations \eqref{eq:taylor-f} and \eqref{eq:taylor-Dexp} into the integral~\eqref{eq:switch-geo} leads to
\begin{equation} \label{eq:taylor-int}
\int_{\textbf{s} \in U : \| \textup{exp}_{\textbf{p}}(\textbf{s}) - \textbf{p}\|_\B \leq \sigma} \text{grad} \widetilde{f}(0)^{\top} \textbf{s} \, + \, \tfrac{1}{2} \textbf{s}^{\top} \text{hess} \widetilde{f}(0) \textbf{s} \, + \, O(\| \textbf{s} \|_2^3)  \,\, d\textbf{s}.
\end{equation}

\subsection[\textbf{Step 4: approximate the domain of  integration}]{\textbf{Step~4: approximate the domain of  integration}}

In this step, we approximate the \textup{domain} of integration in \eqref{eq:taylor-int} using the Taylor expansion of the exponential map (Definition~\ref{def:domain-approx}). Then we assess the quality of our approximations   (Proposition~\ref{prop:domain-bounds}).

\begin{Definition} \label{def:domain-approx}
For each $\sigma >0$,  we define three subsets of $T_\p \M$ as follows.  
\begin{align*}
    & \exact  := \{ \s \in U : \left\| \textup{exp}_\p(\s) - \p \right\|_\B \leq  \sigma \}, \\[0.5pt]
    & \approxone := \{ \s \in T_\p \M : \left\| L_\p(\s)  \right\|_\B \leq  \sigma \}, \\[0.5pt]
    & \approxtwo := \{ \s \in T_\p \M : \left\| L_\p(\s) + \tfrac12 Q_\p(\s) \right\|_\B \leq  \sigma \}.
\end{align*}
The set $\exact \subseteq T_\p \M$ is the exact domain of integration, parameterized on the tangent space.
The sets $\approxone$ and $\approxtwo$ are approximations to $\exact$, where the manifold around $\p$ is approximated to first and second order, respectively.

\end{Definition}

\smallskip

\begin{Proposition} \label{prop:domain-bounds}

\begin{enumerate}[i.]
    \item \!There exist constants $c_1, \sigma_1 > 0$ such that for all $\sigma \leq \sigma_1$,
 \begin{equation} \label{eq:nice-exactU}
     \exact \subseteq \{\s \in T_\p \M : \left\|\s
\right\|_2 \leq c_1 \sigma\} .
 \end{equation}
    \item \!There exist constants $c_2, c_3, \sigma_2 > 0$ such that for all $\sigma \leq \sigma_2$,
\begin{equation} \label{eq:lem-inc}
    (1-c_2 \sigma) \approxone \subseteq \exact \subseteq (1+c_3 \sigma) \approxone.
\end{equation} \label{eq:domain-approx1}
    \item \!There exist constants $c_4, c_5, \sigma_3 > 0$ such that for all $\sigma \leq \sigma_3$,
\begin{equation} \label{eq:lem-inc-approx2-nice}
    (1-c_5 \sigma^2) \approxtwo \subseteq \exact \subseteq (1+c_4 \sigma^2) \approxtwo.
\end{equation}
\end{enumerate}
\end{Proposition}
\begin{proof}
    \underline{\textup{part~i.}}
    First, note that for any $\epsilon > 0$, we can shrink $U$ and $V$ in \eqref{eq:exp-UV} so as to guarantee that for all $\s \in U$ we have
    \begin{equation} \label{eq:assume-U}
        \left\|
            \textup{exp}_{\textbf{p}}(\textbf{s})
            -
            \textbf{p}
            -
            L_{\textbf{p}}(\textbf{s})
        \right\|_\B
        \leq
        \epsilon
        \| \textbf{s} \|_2.
\end{equation}

    Let $\textbf{s} \in \exact$. 
    The result follows from:
    \begin{align*}
        \sigma &\geq \left\|\textup{exp}_{\textbf{p}}(\textbf{s}) - \textbf{p} \right\|_\B
        &&
        \text{[definition of $\exact$]}
        \\
        &=
        \left\| L_{\textbf{p}}(\textbf{s}) - (L_{\textbf{p}}(\textbf{s}) -\textup{exp}_{\textbf{p}}(\textbf{s}) + \textbf{p})   \right\|_B
        \\
        &\ge \| L_{\textbf{p}}(\textbf{s})\|_\B - \| L_{\textbf{p}}(\textbf{s}) -\textup{exp}_{\textbf{p}}(\textbf{s})
+ \textbf{p}\|_\B
        && 
        \text{[reverse triangle inequality]}
        \\
        &\ge \| L_{\textbf{p}}(\textbf{s})\|_\B - \epsilon \| \textbf{s} \|_2
        &&
        \text{[using \eqref{eq:assume-U}]}
        \\
        & \ge c  \| L_{\textbf{p}}(\textbf{s})\|_2 - \epsilon \| \textbf{s} \|_2 && \text{[norm equivalence, see \eqref{eq:equiv-norms}]} \\
        &= (c-\epsilon) \| \s \|_2.
        &&
        \text{[$L_\p$ is an isometry]}
    \end{align*}
     \medskip
     
     \underline{\textup{part~ii.}}
     For the right inclusion, assume $\textbf{s} \in \exact$.  Then,
     \begin{align*}
        \sigma &\ge \left\|\textup{exp}_{\textbf{p}}(\textbf{s}) - \textbf{p} \right\|_\B && \quad \,\,\,\,\,\,\,
        \text{[definition of $\exact$]}\\
        &=
        \left\| L_\p(\s) + O(\|\s\|^2)\right\|_\B && \quad \,\,\,\,\,\,\, \text{[from \eqref{eq:expp-taylor}]}
        \\
        &=
        \left\| L_\p(\s) + O(\sigma^2)\right\|_\B && \quad \,\,\,\,\,\,\, \text{[shown in part i]}
        \\
        &= \left\| L_\p(\s) \right\|_B + O(\sigma^2). && \quad \,\,\,\,\,\,\, \text{[triangle inequality]}
     \end{align*}
Take $c_3$ to be the implicit constant inside the $O(\sigma^2)$ term, it follows that $\| L_\p(\s) \|_B \le \sigma + c_3 \sigma^2$ and therefore $\s \in (1+c_3 \sigma) \approxone$.

For the left inclusion, let $\s \in (1-c_2 \sigma) \approxone$.
It follows by definition that $\|L_\p(\s)\|_\B \le \sigma - c_2 \sigma^2$.
We have just shown that
\begin{align*}
    \left\|\textup{exp}_{\textbf{p}}(\textbf{s}) - \textbf{p} \right\|_\B
    &=
    \left\| L_\p(\s) \right\|_B + O(\sigma^2)
\end{align*}
Therefore,
\begin{align*}
    \left\|\textup{exp}_{\textbf{p}}(\textbf{s}) - \textbf{p} \right\|_\B
    &\le
    \sigma - c_2 \sigma^2 + O(\sigma^2)
\end{align*}
Picking $c_2$ to be the implicit constant inside the $O(\sigma^2)$ term guarantees that $\left\|\textup{exp}_{\textbf{p}}(\textbf{s}) - \textbf{p} \right\|_\B \le \sigma$ and hence that $\textbf{s} \in \exact$.
    
    \medskip
    
    \underline{\textup{part~iii.}}  From Eq. \eqref{eq:expp-taylor} we have that
    \begin{align} \label{eq:thing}
        \left\| \textup{exp}_{\textbf{p}}(\textbf{s}) - \textbf{p} \right\|_\B
        &=
        \left\| L_{\textup{p}}(\textbf{s}) + \tfrac{1}{2} Q_{\textbf{p}}(\textbf{s}) + O(\|\textbf{s}\|^3) \right\|_\B
        \\
        &=
        \left\| L_{\textup{p}}(\textbf{s}) + \tfrac{1}{2} Q_{\textbf{p}}(\textbf{s})\right\|_\B + O(\|\textbf{s}\|^3)
    \end{align}
    By part i, $\|\textbf{s}\|_2 = O(\sigma)$.  From \eqref{eq:thing} and the triangle inequality, 
    \begin{equation} \label{eq:c4'}
        \left\| \textup{exp}_{\textbf{p}}(\textbf{s}) - \textbf{p} \right\|_\B = \left\| L_{\textbf{p}}(\textbf{s}) + \tfrac{1}{2} Q_{\textbf{p}}(\textbf{s}) \right\|_\B \, \leq \, \sigma + c_4' \sigma^3,
    \end{equation}
    for some constant $c_4' > 0$ and all sufficiently small $\sigma$.
    We want to find a constant $c_4 > 0$ such that $ \textbf{s} / (1 + c_4 \sigma^2) \in \approxtwo$.
    To this end, compute
    \begin{align} \label{eq:approx2-bound}
       &\left\| \frac{1}{1 + c_4 \sigma^2} L_{\textbf{p}}(\textbf{s}) \, + \, \frac{1}{(1 + c_4 \sigma^2)^2 } \frac{1}{2} Q_{\textbf{p}}(\textbf{s}) \right\|_\B  \nonumber \\[0.2pt]
       &\leq \,\,\,  \frac{1}{1 + c_4 \sigma^2} \left\| L_{\textbf{p}}(\textbf{s}) + \tfrac{1}{2} Q_{\textbf{p}}(\textbf{s}) \right\|_\B \,\, + \, \left(\frac{1}{1+c_4 \sigma^2} - \frac{1}{(1+c_4 \sigma^2)^2} \right) \left\| \tfrac{1}{2} Q_{\textbf{p}}(\textbf{s}) \right\|_\B \nonumber \\[1.5pt]
       &\leq \,\,\,  \frac{1}{1+c_4\sigma^2} \left( \sigma + c_4' \sigma^3 \right) \,\, + \,\, \left(\frac{1}{1+c_4 \sigma^2} - \frac{1}{(1+c_4 \sigma^2)^2} \right) O(\sigma^2) \nonumber \\[3pt]
       &= \,\,\, \left(1 - c_4 \sigma^2 + O(\sigma^4) \right) \left(\sigma + c_4' \sigma^3 \right) \,\, + \,\, \left( c_4 \sigma^2 + O(\sigma^4) \right) O(\sigma^2) \nonumber \\[3pt]
       &= \,\,\, \sigma \,\, + \,\, (c_4' - c_4) \sigma^3 \,\, + \,\, O(\sigma^4).
    \end{align}
    Here we used the triangle inequality, 
   the bound \eqref{eq:c4'}, part i, and Taylor expansions in $\sigma$ for $(1 + c_4 \sigma^2)^{-1}$ and $(1 + c_4 \sigma^2)^{-2}$.  Thus, take $c_4 = 2c_4'$.   For small enough $\sigma$, the RHS of Eq.~\eqref{eq:approx2-bound} is at most $\sigma$, so $ \textbf{s} / (1 + c_4 \sigma^2) \in \approxtwo$.
Now consider the leftmost inclusion in part iii. Assume $\textbf{s} \in \approxtwo$.  We first prove that $\| \textbf{s} \|_2 = O(\sigma)$.
   Indeed, 
   \begin{align} \label{eq:approx2-Osigma}
       & \sigma \,\, \geq \,\, \left\| L_{\textbf{p}}(\textbf{s}) + \tfrac{1}{2}Q_{\textbf{p}}(\textbf{s}) \right\|_\B \,\, \gtrsim \,\, \left\| L_{\textbf{p}}(\textbf{s}) + \tfrac{1}{2} Q_{\textbf{p}}(\textbf{s}) \right\|_2 \nonumber \\[0.5pt]
       & = \,  \sqrt{\left\| L_{\textbf{p}}(\textbf{s}) \right\|_2^2 \, + \, \tfrac{1}{4} \left\| Q_{\textbf{p}}(\textbf{s}) \right\|_2^2} \,\,\, \geq \, \left\| L_{\textbf{p}}(\textbf{s}) \right\|_2 \, = \, \left\| \textbf{s} \right\|_2.
   \end{align}
   The first equality comes from orthogonality between the images of $Q_{\textbf{p}}$ and $L_{\textbf{p}}$ \eqref{eq:perp}.   
   Let the implicit constant in \eqref{eq:approx2-Osigma} be $c_5'$.  Similarly to above, let us set $c_5 = 2c_5'$ and compute (for sufficiently small $\sigma$):
   \begin{align*}
     &  \, \left\| \textup{exp}_{\textbf{p}}\!\left((1-c_5 \sigma^2)\textbf{s}\right) - \textbf{p} \right\|_\B  
      \leq  \left\| (1 - c_5 \sigma^2 ) L_{\textbf{p}}(\textbf{s})  +   (1 - c_5 \sigma^2)^2 \tfrac{1}{2} Q_{\textbf{p}}(\textbf{s}) \right\|_\B    +  c_5' \sigma^3 \nonumber \\[2pt]
     & \leq  (1 - c_5 \sigma^2) \left\|L_{\textbf{p}} + \tfrac{1}{2} Q_{\textbf{p}}(\textbf{s}) \right\|_\B    +  \left( (1 - c_5 \sigma^2 ) - (1 - c_5 \sigma^2)^2 \right) \left\| \tfrac{1}{2} Q_{\textbf{p}}(\textbf{s}) \right\|_\B  +  c_5' \sigma^3 \nonumber \\[2pt]
     & \leq  (1 - c_5 \sigma^2)\sigma  +  O(\sigma^4)  +  c_5' \sigma^3  =  \sigma - c_5' \sigma^3  +  O(\sigma^4)  \leq  \sigma.
   \end{align*}
   We used the Taylor expansion \eqref{eq:taylor-exp}, the triangle inequality,  the bound \eqref{eq:approx2-Osigma}, and the triangle inequality again.  Hence $(1 - c_5 \sigma^2) \textbf{s} \in \exact$.  
\hfill \qed
\end{proof}

\subsection[Step 5: drop the cubic error term and obtain the second-order term]{\textbf{Step~5: drop  $O(\|\textbf{s}\|_2^3)$ and obtain the second-order term}}  \label{subsec:drop-stuff}

In this step we prove that each of the terms in the integral \eqref{eq:taylor-int} can be approximated up to an additive error of $O(\sigma^{d+3})$ by switching from the exact domain $\exact$ to the approximate domains $\approxone$ and $\approxtwo$ defined in Definition \ref{def:domain-approx}. 

\begin{Proposition} \label{prop:int-bounds}
The following bounds hold:
\begin{enumerate}[i.] \setlength\itemsep{0.5em}
    \item  $\int_{\s \in \exact} O(\| \textup{\textbf{s}} \|_2^3) d\textup{\textbf{s}} \, = \, O(\sigma^{d+3})$, 
    \item 
   $ \int_{\s \in \exact} \textup{\textbf{s}} \textup{\textbf{s}}^{\top} d\textup{\textbf{s}} \,\, =   \int_{\textup{\textbf{s}} \in \approxone} \textup{\textbf{s}} \textup{\textbf{s}}^{\top} d\textup{\textbf{s}}  \, + \, O(\sigma^{d+3})$,
    \item  
    $\int_{\textup{\textbf{s}} \in \exact} \textup{\textbf{s}} d\textup{\textbf{s}} \, = \int_{\textup{\textbf{s}} \in \approxtwo} \textup{\textbf{s}} d\textup{\textbf{s}} \, + \, O(\sigma^{d+3})$.
\end{enumerate}
\end{Proposition}
\begin{proof} Let $\symmdiff$ denote the symmetric difference of sets.

   \noindent \underline{\textup{part~i.}}
    Let $\sigma \leq \sigma_1$.
    By Proposition~\ref{prop:domain-bounds}, part~i, we have
    \begin{align*}
      & \int_{\textbf{s} \in \exact} O(\| \textbf{s} \|_2^3) \, d\textbf{s}
       \,\, \lesssim \,\,
       \int_{\|\textbf{s}\|_2 \leq c_1 \sigma} \| \textbf{s} \|_2^3 \\[0.4em]
       & \quad \le (c_1 \sigma)^3 \text{vol}\{\s \in \mathbb{R}^d:\|s\|_2 \le c_1 \sigma\}
       = O(\sigma^{d+3}).
    \end{align*}

   \noindent \underline{\textup{part~ii.}} Let $\sigma \leq \sigma_2$.  By Proposition~\ref{prop:domain-bounds}, part~ii, we see
\begin{align*}
    \exact   \symmdiff \approxone 
    \, \subseteq \, (1+c_3 \sigma) \approxone \setminus (1 - c_2 \sigma) \approxone.
\end{align*}
Then we have
\vspace{-0.5em}
\begin{align*}
   & \left\| \int_{\textbf{s} \in \exact} \textbf{s} \textbf{s}^{\top} d\textbf{s}  -  \int_{\textbf{s} \in \approxone} \textbf{s} \textbf{s}^{\top} d\textbf{s} \right\|_F \nonumber \\[-0.5pt]
   &= \left\| \int_{\textbf{s} \in \exact \setminus \approxone} \textbf{s} \textbf{s}^{\top} d\textbf{s}\right\|_F \nonumber \\[-0.5pt]
   &\le  \int_{\textbf{s} \in \exact \setminus \approxone} \left\| \s \s^\top \right\|_F d\s \nonumber \\[-0.5pt]
   & \leq \, \int_{\textbf{s} \in \exact  \symmdiff \approxone} \left\| \textbf{s} \textbf{s}^{\top} \right\|_F  d\textbf{s} \nonumber \\[-0.5pt]
  &  \leq \, \int_{\textbf{s} \in (1 + c_3 \sigma) \approxone \setminus (1 - c_2 \sigma) \approxone} \| \textbf{s} \|_2^2 d\textbf{s} \nonumber \\[-0.5pt]
  &  = \, \int_{\textbf{s} \in (\sigma + c_3 \sigma^2) \operatorname{Approx}^{(1)}(1) \setminus (\sigma - c_2 \sigma^2) \operatorname{Approx}^{(1)}(1)} \| \textbf{s} \|_2^2
d\textbf{s} \nonumber \\[-0.5pt]
 &  = \, \left( (\sigma + c_3 \sigma^2)^{d+2} - (\sigma - c_2 \sigma^2)^{d+2}\right) \int_{\| L_{\textbf{p}}(\textbf{s}) \|_\B \leq 1} \| \textbf{s} \|_2^2 d\textbf{s}. \nonumber 
\end{align*}
\vspace{-0.2em}
This last quantity is $O(\sigma^{d+3})$, because $\B$ is bounded and  $L_\p$ is an isometry. 
    
    \medskip
    \medskip
    
  \noindent   \underline{\textup{part~iii.}} 
    Let $\sigma \leq \sigma_3$.
    By Proposition~\ref{prop:domain-bounds}, part~iii, 
    \begin{align*}
    &\exact  \, \symmdiff \,  \approxtwo \\
    &\subseteq \, (1 + c_4 \sigma^2) \approxtwo \setminus (1-c_5\sigma^2) \approxtwo.
    \end{align*}
    Then, 
    \begin{align*}
        &\left\|  \int_{\textbf{s} \in \exact} \textbf{s} d\textbf{s} \, - \, \int_{\textbf{s} \in    \approxtwo} \textbf{s} d\textbf{s} \right\|_2 
        =
        \left\|  \int_{\textbf{s} \in \exact \setminus \approxtwo} \textbf{s} d\textbf{s}\right\|_2     \\[-0.5pt]
        &\le
        \int_{\textbf{s} \in \exact \setminus \approxtwo} \left\|\s\right\|_2 d\s
        \,\, \leq \,\,
        \int_{\s \in \exact \symmdiff \approxtwo} \| \s \|_2 d\s \nonumber \\[-0.5pt ]
        & \leq \, \int_{\textbf{s} \in (1+c_4 \sigma^2)\approxtwo \setminus (1-c_5 \sigma^2) \approxtwo}    \| \textbf{s} \|_2 d\textbf{s}.
    \end{align*}
The upper bound equals 
\begin{align*}
     &  \left( (1 + c_4 \sigma^2)^{d+1} - (1 - c_5 \sigma^2)^{d+1} \right) \int_{\s \in \approxtwo} \| \s \|_2 d\s \nonumber \\[-0.5pt]
    &  = O(\sigma^2) \int_{\s' \in \operatorname{Approx}^{(2)}(1)} \| \sigma \s' \|_2 \sigma^d d \s' \nonumber 
    = O(\sigma^{d+3}) \int_{\s' \in \operatorname{Approx}^{(2)}(1)} \| \s' \|_2 d \s'
   \end{align*}
    The last  quantity is $O(\sigma^{d+3})$, because $\textbf{s} \in \approxtwo$ implies $\|\textbf{s}\|_2 = O(\sigma)$, as shown in the argument for 
    Proposition~\ref{prop:domain-bounds}, part~iii.
    \hfill \qed
\end{proof}

\noindent Now, plug Proposition~\ref{prop:int-bounds} into the integral \eqref{eq:taylor-int}:
\begin{align*}
    & \int_{\textbf{s} \in \exact} \text{grad} \widetilde{f}(0)^{\top} \textbf{s} \, + \, \tfrac{1}{2} \textbf{s}^{\top} \text{hess} \widetilde{f}(0) \textbf{s} \, + \, O(\| \textbf{s} \|_2^3)  \,\, d\textbf{s} \nonumber \\
    &=\left\langle \text{grad} \widetilde{f}(0), \int_{\textbf{s} \in \approxtwo} \textbf{s} d\textbf{s}\right\rangle \nonumber \\
    &\ \ \ \ + 
    \left\langle \text{hess} \widetilde{f}(0), \tfrac{1}{2} \int_{\textbf{s} \in \approxone} \textbf{s} \textbf{s}^{\top} d\textbf{s}\right\rangle_F + O(\sigma^{d+3}),
\end{align*}
where linearity of $L_{\textbf{p}}$ gives
\begin{align*}
\int_{\textbf{s} \in \approxone} \textbf{s} \textbf{s}^{\top} d\textbf{s} = \sigma^{d+2} \int_{\textbf{s} : \| L_{\textbf{p}}(\textbf{s}) \|_\B \leq 1} \textbf{s} \textbf{s}^{\top} d\textbf{s}.
\end{align*}
Thus, \eqref{eq:taylor-int} divided by $\sigma^{d+2}$ tends to $\Delta_{\mathcal{M}, \B}f(\textbf{p})$ as $\sigma \rightarrow 0$, as desired,  \textbf{provided we can show}
\begin{equation} \label{eq:I1-integral}
    \lim_{\sigma \rightarrow 0} \frac{1}{\sigma^{d+2}} \int_{\textbf{s} \in \approxtwo} \textbf{s} d\textbf{s} 
    \, = \int_{ \| \widehat{\textup{\textbf{s}}} \|_2 =1 } \widehat{\textup{\textbf{s}}} \, \|L_{\textup{\textbf{p}}}(\widehat{\textup{\textbf{s}}})\|_\B^{-d} \tilt(\widehat{\textup{\textbf{s}}}) \, d\widehat{\textup{\textbf{s}}}.
\end{equation}

\subsection[\textbf{Step 6: use spherical coordinates}]{\textbf{Step~6: use spherical coordinates}} \label{subsec:spherical-coords}
It remains to estimate 
\begin{equation} \label{eq:hard-int}
    \int_{\textbf{s} \in \approxtwo} \textbf{s} d\textbf{s}.
\end{equation}
First, we provide intuition for why this integral should scale like $\sigma^{d+2}$.  Let $\mathbb{S}^{d-1}\subseteq\mathbb{R}^{d} \cong T_{\textbf{p}}\mathcal{M}$ denote the $\ell_2$-unit sphere, with  density $d\widehat{\textbf{s}}$, where $\widehat{\textbf{s}} \in \mathbb{S}^{d-1}$.  Let $ r \in \mathbb{R}_{\geq 0}$ be a radial variable with density $dr$.  Consider the integral \eqref{eq:hard-int} in these spherical coordinates.  Substituting $\textbf{s} = r\widehat{\textbf{s}}$ and $d \textbf{s} = r^{d-1} dr d\widehat{\textbf{s}}$, 
\begin{equation} \label{eq:hard-int-sph}
    \int_{\textbf{s} : \| L_{\textbf{p}}(\textbf{s}) + \frac{1}{2} Q_{\textbf{p}}(\textbf{s}) \|_\B \leq \sigma} \textbf{s} d\textbf{s} =  \int_{\widehat{\textbf{s}} \in \mathbb{S}^{d-1}} \widehat{\textbf{s}}  \int_{r \in \operatorname{RadialDomain}(\widehat{\textbf{s}}, \sigma)} r^{d} dr d\widehat{\textbf{s}},
\end{equation}
where we define
\begin{equation} \label{eq:rad-dom-1}
    \operatorname{RadialDomain}(\widehat{\textbf{s}}, \sigma) := 
    \left\{ r \geq 0: \left\|r L_{\textbf{p}}(\widehat{\textbf{s}}) +  \tfrac{r^2}{2}Q_{\textbf{p}}(\widehat{\textbf{s}}) \right\|_\B \leq \sigma \right\}.
\end{equation}
This is the second-order approximation set $\approxtwo$ intersected with the ray in the direction of $\widehat{\textbf{s}}$. Note that by definition, \label{}
\begin{align} \label{eq:radialdomain_vs_approxtwo}
    \operatorname{RadialDomain}(\widehat{\textbf{s}}, \sigma) = \{ r \ge 0 : r \widehat{\s} \in \approxtwo\}.
\end{align}
Compare this domain of integration against the domain for $-\widehat{\textbf{s}}$:
\begin{align} \label{eq:rad-dom-2}
    \operatorname{RadialDomain}(-\widehat{\textbf{s}}, \sigma) & = 
    \left\{ r \geq 0: \left\|r L_{\textbf{p}}(-\widehat{\textbf{s}}) + \tfrac{r^2}{2}Q_{\textbf{p}}(-\widehat{\textbf{s}}) \right\|_\B \leq \sigma \right\} \nonumber \\[3pt]
    & \,=  \left\{ r \geq 0: \left\|r L_{\textbf{p}}(\widehat{\textbf{s}}) - \tfrac{r^2}{2}Q_{\textbf{p}}(\widehat{\textbf{s}}) \right\|_\B \leq \sigma \right\}.
\end{align}
Speaking roughly, the condition determining membership in \eqref{eq:rad-dom-1} differs from that in \eqref{eq:rad-dom-2} at the $O(r^2)$ term;  the conditions would be the same without the $Q_{\textbf{p}}$ term.  On the other hand, the integrand in \eqref{eq:hard-int-sph} is \textup{odd} (that is, it flips sign upon inversion in the origin).  Therefore, we should expect ``near cancellation" between the inner integrals:
\begin{equation} \label{eq:sum-hard-ints}
   \widehat{\textbf{s}} \int_{r \in \operatorname{RadialDomain}(\widehat{\textbf{s}}, \sigma)} r^{d} dr \,\,\, -  \,\,\, \widehat{\textbf{s}} \int_{r \in \operatorname{RadialDomain}(-\widehat{\textbf{s}}, \sigma)} r^{d} dr.
\end{equation}
Supposing $r = O(\sigma)$ for $r$ in each radial domain (justified in Lemma \ref{lem:radial-dom-bound}), then each of the two terms in \eqref{eq:sum-hard-ints} is $O(\sigma^{d+1})$.  Thus after ``near, but not complete, cancellation" the sum \eqref{eq:sum-hard-ints} is expected to be $O(\sigma^{d+2})$. Then, integrating over the unit sphere gives $O(\sigma^{d+2})$ in \eqref{eq:hard-int-sph}. This informal discussion explains why we expect the integral \eqref{eq:hard-int} to be $O(\sigma^{d+2})$, the mechanism being 
cancellation  due to an approximate equality between $\operatorname{RadialDomain}(\widehat{\textbf{s}}, \sigma)$ and $\operatorname{RadialDomain}(-\widehat{\textbf{s}}, \sigma)$.

We shall now make this claim rigorous. 
Beyond proving that the integral
\eqref{eq:hard-int-sph} is $O(\sigma^{d+2})$, we will prove that dividing \eqref{eq:hard-int-sph} by $\sigma^{d+2}$ produces a well-defined limit as $\sigma \rightarrow 0$, namely the RHS of \eqref{eq:I1-integral}.  

\begin{Remark}
Steps 7-8 below are complicated by the fact 
 that we do not assume smoothness of the norm $\| \cdot \|_\B$.  As a consequence, \textit{a priori} we cannot Taylor-expand the boundary \nolinebreak points \nolinebreak of the radial domain in the variable $\sigma$.
 \end{Remark}

\begin{Lemma} \label{lem:radial-dom-bound}
There exist constants $c_6, \sigma_4 > 0$ such that for all $\sigma \leq \sigma_4$ and all $\widehat{\s} \in \mathbb{S}^{d-1}$, we have
\begin{align*}
    \operatorname{RadialDomain}(\widehat{\s}, \sigma) \, \subseteq \, c_6[0, \sigma].
\end{align*}
\end{Lemma}
\begin{proof}
    Set $c_6 := 2c_1$ and $\sigma_4 := \min\left(\sigma_1, \sigma_3, \frac{1}{\sqrt{2c_5}} \right)$.  For $\sigma \leq \sigma_{4}$, we have
    \begin{align*}
        \approxtwo  &\subseteq  \frac{1}{1-c_5\sigma^2} \exact && \text{[by \eqref{eq:lem-inc-approx2-nice}]} \\
        &\subseteq \frac{c_1}{1-c_5\sigma^2} \{\textbf{s} \in \mathbb{R}^d: \| \textbf{s} \|_2 \leq \sigma \}. && \text{[by \eqref{eq:nice-exactU}]}
    \end{align*}
    Note that $\sigma \leq \frac{1}{\sqrt{2c_5}}$ implies $\frac{c_1}{1-c_5\sigma^2} \leq 2c_1 = c_6$. Therefore,
   \begin{align*}
    \approxtwo \subseteq c_6 \{\textbf{s} \in \mathbb{R}^d : \|\textbf{s} \|_2 \leq \sigma \}.
   \end{align*}
   By \eqref{eq:radialdomain_vs_approxtwo} it follows that $\operatorname{RadialDomain}(\widehat{\textbf{s}},\sigma) \subseteq c_6[0,\sigma]$ for all $\widehat{\textbf{s}} \in \mathbb{S}^{d-1}$. \hfill \qed
\end{proof}

\subsection[Step 7: study the boundary of the radial domain]{\textbf{Step~7: study the boundary of $\operatorname{RadialDomain}(\widehat{\textbf{s}},\sigma)$}} \label{subsec:boundary}

In this step, we show that for small enough $\sigma$, the set $\operatorname{RadialDomain}(\widehat{\textbf{s}},\sigma)$ is a single closed interval in $\mathbb{R}_{\geq 0}$.  Then, we prove its nonzero boundary point is a continuous function of $(\widehat{\textbf{s}}, \sigma)$ and we bound this to second-order in $\sigma$.

Let
$
G : \mathbb{S}^{d-1} \times \mathbb{R}_{\geq 0} \longrightarrow \mathbb{R}_{\geq 0}; \,\, (\widehat{\textbf{s}}, r) \longmapsto \left\| rL_{\textbf{p}}(\widehat{\textbf{s}}) +  \tfrac{r^2}{2}Q_{\textbf{p}}(\widehat{\textbf{s}}) \right\|_\B.
$
\begin{Lemma} \label{lem:strict-increasing}
There exists a constant $c_7 > 0$ such that for all $\widehat{\textup{\textbf{s}}}\in \mathbb{S}^{d-1}$, the function $r \mapsto G(\widehat{\textup{\textbf{s}}}, r)$ is strictly increasing in $r \in [0,c_7]$.
\end{Lemma}
\begin{proof}
We will show that we can take
\begin{align*} 
c_7 := \begin{cases}
1  \quad \quad \quad \quad \quad \quad \quad \quad \quad \quad \quad \quad \quad \quad \quad \quad \quad \textup{if } Q_{\textup{\textbf{p}}}  \equiv 0, \\
 \min_{\widehat{\textup{\textbf{s}}} \in \mathbb{S}^{d-1}}\! \| L_{\textup{\textbf{p}}}(\widehat{\textup{\textbf{s}}}) \|_\B \big{/} \! \max_{\widehat{\textup{\textbf{s}}} \in \mathbb{S}^{d-1}} \! \|Q_{\textup{\textbf{p}}}(\widehat{\textup{\textbf{s}}}) \|_\B  \quad \textup{otherwise}.
\end{cases}
\end{align*}
    Obviously  $c_7 >0$, because $\| L_{\textbf{p}}(\widehat{\textbf{s}})\|_\B \gtrsim \| L_{\textbf{p}}(\widehat{\textbf{s}}) \|_2 = \|\widehat{\textbf{s}} \|_2 = 1$ by equivalence of norms and \eqref{eq:isometric}, and $Q_{\textbf{p}}(\widehat{\textbf{s}}) = O(1)$ by continuity  and compactness. 
    
    Fix $\widehat{\textbf{s}} \in \mathbb{S}^{d-1}$\!.  Set $g: \mathbb{R}_{\geq 0} \rightarrow \mathbb{R}_{\geq 0} $ by $g(r) = G(\widehat{\textbf{s}},r)$,  $\textbf{a} := L_{\textbf{p}}(\widehat{\textbf{s}})$ and $\textbf{b} := \frac{1}{2} Q_{\textbf{p}}(\widehat{\textbf{s}})$.
    Since $g$ is continuous, we may check $g$ is strictly increasing on the half-open interval $[0,c_7)$.

    Let $\lambda > 0$ satisfy $(1+\lambda)r < c_7$. 
    By the reverse triangle inequality and the definition of $c_7$, we have
    \begin{align*} \label{eq:g-increasing-2}
    g((1+\lambda)r)  
      & = 
  \left\|(1+\lambda)(r\textbf{a}  + r^2 \textbf{b}) + (\lambda + \lambda^2)r^2\textbf{b} \right\|_\B \\[-0.2pt]
  &  \geq   (1+\lambda) \| r \textbf{a} + r^2 \textbf{b} \|_\B - (\lambda + \lambda^2)r^2 \|\textbf{b}\|_\B  \\[-0.2pt]
    &    =  \|r \textbf{a} + r^2 \textbf{b} \|_\B + \lambda r ( \| \textbf{a} + r \textbf{b} \|_\B - (1+\lambda)r \| \textbf{b} \|_\B )  \\[-0.2pt]
        & \geq   \|r \textbf{a} + r^2 \textbf{b} \|_\B + \lambda r  ( \| \textbf{a} \|_\B - r \|\textbf{b} \|_\B - (1+\lambda)r \|\textbf{b} \|_\B)  \\[-0.2pt]
        & >  \|r \textbf{a} + r^2 \textbf{b} \|_\B + \lambda r ( \| \textbf{a}\|_\B - \tfrac{1}{2} \| \textbf{a} \|_\B - \tfrac{1}{2} \| \textbf{a} \|_\B).
    \end{align*}
    The last quantity equals $g(r)$, and the lemma follows. \hfill \qed
  \end{proof}

The following quantity is well-defined as a consequence of Lemma~\ref{lem:strict-increasing}.
\begin{Cordef}\label{lem:defrstar}
There exists a constant $\sigma_5 > 0$ such that for all $\sigma \leq \sigma_5$ and all $\widehat{\textup{\textbf{s}}} \in \mathbb{S}^{d-1}$, 
$\operatorname{RadialDomain}(\widehat{\textup{\textbf{s}}}, \sigma)$ is a closed interval.  Thus, there exists a function 
\begin{equation} \label{eq:def-r*-rad}
r^* : \mathbb{S}^{d-1} \times [0,\sigma_5] \rightarrow \mathbb{R}_{\geq 0} \quad \! \textup{such that} \,\, \operatorname{RadialDomain}(\widehat{\textup{\textbf{s}}},\sigma) = [0,r^*(\widehat{\textup{\textbf{s}}}, \sigma)].
\end{equation}
\end{Cordef}

\begin{Lemma} \label{lem:continuous}
There exists a constant $\sigma_6 > 0$ such that the restriction of $r^*$  to $\mathbb{S}^{d-1} \times [0, \sigma_6]$ is a continuous function.
\end{Lemma}

\begin{proof}
    Take $\sigma_6 = \sigma_5 /2 $. We shall verify continuity of $r^*$ by bare hands. For notational convenience, within this proof, we denote the second argument of $r^*$ by $\tau$ (subscripted and/or primed) rather than by $\sigma$.  
    Fix $(\widehat{\textbf{s}}_1, \tau_1) \in \mathbb{S}^{d-1} \times [0,\sigma_6]$ and let $\epsilon >0$.  
    
    Lemma~\ref{lem:strict-increasing} says $r \mapsto G(\widehat{\textbf{s}}_1, r)$ is continuous and strictly increasing around $0$.  By elementary facts, this has a well-defined continuous inverse function around $G(0) = 0$.  The inverse function is $\tau \mapsto r^*(\widehat{\textbf{s}}_1,\tau)$ defined for $\tau \in [0,\sigma_5]$ (Lemma~\ref{lem:defrstar}).  So, we can take $\delta' \in (0, \sigma_6)$ such that for all $\tau_2' \in [0, \sigma_5]$,
    \begin{equation} \label{eq:delta-prime}
    | \tau_2' - \tau_1| < \delta' \implies  | r^*(\widehat{\textbf{s}}_1,\tau_2') - r^*(\widehat{\textbf{s}}_1,\tau_1) | < \epsilon.
    \end{equation}
    Since $L_{\textbf{p}}, Q_{\textbf{p}}$ are continuous, there exist $\delta'', \delta''' > 0$ such that for all $\widehat{\textbf{s}}_2 \in \mathbb{S}^{d-1}$,
    \begin{align}
        & \| \widehat{\textbf{s}}_2  - \widehat{\textbf{s}}_1 \|_2 < \delta'' \implies \left\| L_{\textbf{p}}(\widehat{\textbf{s}}_2) - L_{\textbf{p}}(\widehat{\textbf{s}}_1) \right\|_\B < \frac{1}{c_7} \frac{\delta'}{3}, \label{eq:def-delta''} \\[2pt]
        & \| \widehat{\textbf{s}}_2  - \widehat{\textbf{s}}_1 \|_2 < \delta''' \implies \left\| \frac{1}{2}Q_{\textbf{p}}(\widehat{\textbf{s}}_2) - \frac{1}{2}Q_{\textbf{p}}(\widehat{\textbf{s}}_1) \right\|_\B < \frac{1}{c_7^2} \frac{\delta'}{3}. \label{eq:def-delta'''}
    \end{align}

Define $\delta := \min\left(\delta' / 3, \delta'', \delta'''\right) > 0$.  Let $(\widehat{\textbf{s}}_2, \tau_2) \in \mathbb{S}^{d-1} \times [0, \sigma_6]$
satisfy $\left\|\left(\widehat{\textbf{s}}_2, \tau_2 \right) - \left(\widehat{\textbf{s}}_1, \tau_1\right)\right\|_2 < \delta$.
We shall verify $|r^{*}(\widehat{\textbf{s}}_2, \tau_2) - r^*(\widehat{\textbf{s}}_1, \tau_1)| < \epsilon$.
Put $r_1 := r^*(\widehat{\textbf{s}}_1, \tau_1), r_2 := r^*(\widehat{\textbf{s}}_2, \tau_2)$ and 
$\tau_2' := G(\widehat{\textbf{s}}_1,r_2)$.  By \eqref{eq:delta-prime}, it suffices to check  $|\tau_2' - \tau_1 | < \delta'$, as then $\tau_2' \in [0, \sigma_5]$ (because $\tau_2' \leq \tau_1 + \delta' \leq 2 \sigma_6 = \sigma_5$) and also 
$r^{*}(
\widehat{\textbf{s}}_1,\tau_2') = r^{*}(
\widehat{\textbf{s}}_1,G(\widehat{\textbf{s}}_1, r_2)) = r_2$. So, \eqref{eq:delta-prime} gives $|r_2 - r_1| < \epsilon$.

To see that $|\tau_2' - \tau_1| < \delta$ indeed holds, we write
\begin{align*}
    & \hspace{4em} \tau_2' = \left\|r_2 L_{\textbf{p}}(\widehat{\textbf{s}}_1) + r_2^2 \frac{1}{2} Q_{\textbf{p}}(\widehat{\textbf{s}}_1)  \right\|_\B  =\\
    &
    \left\|(r_2 L_{\textbf{p}}(\widehat{\textbf{s}}_2) + \frac{r_2^2}{2} Q_{\textbf{p}}(\widehat{\textbf{s}}_2) )+  
    r_2( L_{\textbf{p}}(\widehat{\textbf{s}}_1) - L_{\textbf{p}}(\widehat{\textbf{s}}_2)) + r_2^2 ( \tfrac{1}{2}Q_{\textbf{p}}(\widehat{\textbf{s}}_2) - \tfrac{1}{2}Q_{\textbf{p}}(\widehat{\textbf{s}}_1) ) \right\|_\B\!.
\end{align*}

Using the triangle inequality, \eqref{eq:def-delta''}, \eqref{eq:def-delta'''} and  $r_2 \leq c_7$ (from the proof of Lemma~\ref{lem:defrstar}), 
\begin{align*}
    |\tau_2' - \tau_1| \,\, \leq \,\, |\tau_2 -\tau_1| + |\tau_2' - \tau_2| \,\, < \,\, \frac{\delta'}{3} + c_7 \frac{1}{c_7} \frac{\delta'}{3} + c_7^2 \frac{1}{c_7^2}\frac{\delta'}{3} \,\, = \,\, \delta'.
\end{align*}
This proves  $r^{*}$ is continuous on $\mathbb{S}^{d-1} \times [0, \sigma_6]$, when $\sigma_6 = \sigma_5 /2$. \hfill \qed
\end{proof}

\begin{Lemma} \label{lem:bound-r*}
There exist constants $c_8 \geq 0$ and $\sigma_7 > 0$ such that for all $\sigma \leq \sigma_7$ and all $\widehat{\textup{\textbf{s}}} \in \mathbb{S}^{d-1}$, 
\begin{equation} \label{eq:bound-r*}
    \frac{1}{\|L_{\textup{\textbf{p}}}(\widehat{\textup{\textbf{s}}}) \|_\B} \sigma - c_8 \sigma^2 \, \leq \, r^*(\widehat{\textup{\textbf{s}}}, \sigma) \, \leq \, \frac{1}{\|L_{\textup{\textbf{p}}}(\widehat{\textup{\textbf{s}}}) \|_\B} \sigma + c_8 \sigma^2.
\end{equation}
\end{Lemma}
\begin{proof}
We shall prove that we may take 
\begin{equation} \label{eq:c8-def}
    c_8 =  \frac{\max_{\widehat{\textbf{s}} \in \mathbb{S}^{d-1}} \|Q_{\textbf{p}}(\widehat{\textbf{s}})\|_\B}{\min_{\widehat{\textbf{s}} \in \mathbb{S}^{d-1}} \|L_{\textbf{p}}(\widehat{\textbf{s}})\|_\B^3}.
\end{equation}
    
    Given $\widehat{\textbf{s}} \in \mathbb{S}^{d-1}$.  Write $\textbf{a} = L_{\textbf{p}}(\widehat{\textbf{s}})$, $\textbf{b} = \frac{1}{2}Q_{\textbf{p}}(\widehat{\textbf{s}})$, $a = \|\textbf{a}\|_\B$ and $b = \|\textbf{b}\|_\B$.  If $\textbf{b} = 0$, then $r^*(\widehat{\textbf{s}}, \sigma) = (1/a) \sigma$ for all $\sigma \geq 0$, so \eqref{eq:bound-r*} is obviously satisfied.   Assume $\textbf{b} \neq 0$.  The triangle inequality gives 
    \begin{equation} \label{eq:two-quads}
     g_{-}(r) \, \leq \, g(r) \, \leq \, g_{+}(r) \quad \textup{for all } r \in \mathbb{R}_{\geq 0},
    \end{equation}
    where $g(r) := G(\widehat{\textbf{s}},r) = \| r \textbf{a} + r^2 \textbf{b} \|_\B$, $g_{-}(r) := ar - br^2$ and $g_{+}(r) := ar + br^2$.  Note $g_{+}$ is strictly increasing over $r \in [0, \infty)$, while $g_{-}$ is strictly increasing over $r \in [0, \frac{a}{2b}]$ and $g_{-}(\frac{a}{2b}) = \frac{a^2}{4b}$.  Let 
    \begin{align*}
        \sigma'_6 := \min\left(\sigma_5, \frac{\min_{\widehat{\textbf{s}} \in \mathbb{S}^{d-1}} \|L_{\textbf{p}}(\widehat{\textbf{s}})\|_\B^2}{4 \max_{\widehat{\textbf{s}} \in \mathbb{S}^{d-1}} \| Q_{\textbf{p}}(\widehat{\textbf{s}}) \|_\B } \right) \, > \, 0.
    \end{align*}
    It follows from \eqref{eq:two-quads} and the intermediate value theorem that for all $\sigma \in [0,\sigma_7']$,
    \begin{align*}
        r_{+,+}^*(\sigma) \, \leq \, r^*(\sigma) \, \leq \, r_{-,-}^*(\sigma),
    \end{align*}
    where $r^*_{+,+}(\sigma)$ denotes the greater of the two roots in $r$ to the quadratic equation $g_{+}(r) = \sigma$ and where  $r^*_{-,-}(\sigma)$ denotes the lesser of the two roots in $r$ to $g_{-}(r) = \sigma$.  Explicitly by the quadratic formula and Taylor series for the square root function, we have
    \begin{align} \label{eqref:quadratic-bounds}
       & r^*_{+,+}(\sigma) \,\, := \,\, \frac{-a + \sqrt{a^2 + 4b\sigma}}{2b}  \,\, = \,\, \frac{1}{a}\sigma - \frac{b}{a^3}\sigma^2 + O(\sigma^3),  \nonumber \\[4pt]
      &  r^*_{-,-}(\sigma) \,\, := \,\, \frac{a - \sqrt{a^2 - 4b\sigma}}{2b} \,\, =\,\,  \frac{1}{a}\sigma + \frac{b}{a^3}\sigma^2 + O(\sigma^3).
    \end{align}

On the other hand, from the compactness of $\mathbb{S}^{d-1}$, one can check  the implicit constants suppressed by the big $O$ notation in \eqref{eqref:quadratic-bounds} may all be taken independently of $\widehat{\textbf{s}}$.   At the same time, $\frac{2b}{a^3} \leq c_8$ for each $\widehat{\textbf{s}} \in \mathbb{S}^{d-1}$, by the definition \eqref{eq:c8-def}. Taking $\sigma_7 > 0$ to be sufficiently smaller than $\sigma_7'$ yields the lemma. \hfill \qed
\end{proof}

\begin{Definition}
Set $\sigma_8 := \min(\sigma_6,\sigma_7) > 0$.  Define  $\eta^* : \mathbb{S}^{d-1} \times (0, \sigma_8] \rightarrow \mathbb{R}$ by
\begin{equation} \label{eq:eta-def}
    r^*(\widehat{\textbf{s}}, \sigma) =: \frac{1}{\|L_{\textbf{p}}(\widehat{\textbf{s}})\|_\B}\sigma + \tfrac{1}{2}\eta^*(\widehat{\textbf{s}},\sigma) \sigma^2.
\end{equation}
\end{Definition}

By Lemma~\ref{lem:continuous},  $\tfrac{1}{2}\eta^*$ is continuous. By Lemma~\ref{lem:bound-r*}, it is bounded uniformly in absolute value by $2c_8$.

\subsection[Step 8: obtain tilt and apply dominated convergence]{\textbf{Step 8: obtain $\tilt$ and apply dominated convergence}} \label{subsec:obtain-tilt}

It remains to establish Eq.~\eqref{eq:I1-integral}: that is, to obtain the first-order term in the limiting differential operator.   We do this using spherical coordinates (Section~\ref{subsec:spherical-coords}) and the results about the radial integration domain developed in Section~\ref{subsec:boundary}. By swapping the order of limit and the integration (justified by dominated convergence), the tilt function emerges at last. 

\begin{Proposition} \label{prop:obtain-tilt}
For each $\widehat{\textup{\textbf{s}}} \in \mathbb{S}^{d-1}$, 
\begin{equation} \label{eq:tilt-lim-eq}
    \lim_{\sigma \rightarrow 0} \tfrac{1}{2} \eta^*(\widehat{\textup{\textbf{s}}}, \sigma) = \tilt(\widehat{\textup{\textbf{s}}}).
\end{equation}
In particular, the limit on the LHS exists. 
\end{Proposition}
\begin{proof}
    By the bound~\eqref{eq:bound-r*} and compactness of $[-c_8, c_8]$, it suffices to show that every accumulation point of $\frac{1}{2}\eta^*(\widehat{\textbf{s}}, \sigma)$ as $\sigma \rightarrow 0$ equals $\tilt(\widehat{\textbf{s}})$.  That is,  assume $\left( \tau_k \right)_{k=1}^{\infty} \subseteq (0,\sigma_8]$ is such that $\tau_{k} \rightarrow 0$ and $\frac{1}{2}\eta^*(\widehat{\textbf{s}}, \tau_k) \rightarrow \eta \in [-c_8, c_8]$ as $k \rightarrow \infty$; we will show $\eta = \tilt(\widehat{\textbf{s}})$.  
    Substituting \eqref{eq:eta-def} into \eqref{eq:def-r*-rad}, and putting $\textbf{a} = L_{\textbf{p}}(\widehat{\textbf{s}})$, $\textbf{b} = \frac{1}{2} Q_{\textbf{p}}(\widehat{\textbf{s}})$, $\eta_k = \eta(\widehat{\textbf{s}}, \tau_k)$, gives
    \begin{align*}
        \tau_k = \left\| \left(\frac{\tau_k}{\| \textbf{a}\|_\B} + \frac{1}{2} \eta_k \tau_k^2\right)\!\textbf{a} \, \, + \, \left(\frac{\tau_k}{\| \textbf{a}\|_\B} + \frac{1}{2} \eta_k \tau_k^2\right)^{\!\!2}\!\textbf{b} \right\|_\B.
    \end{align*}
    Rearranging and dividing by $\tau_k$, this reads
    \begin{equation} \label{eq:sub-in-tan}
    1 = \left\| \frac{\textbf{a}}{\|\textbf{a}\|_\B} \, + \, \left( \frac{1}{2}\eta_k \textbf{a} + \frac{\textbf{b}}{\|\textbf{a}\|_\B^2}\right)\!\tau_k + \frac{\eta_k \textbf{b}}{\|\textbf{a}\|_\B} \tau_k^2 + \frac{\eta_k^2 \textbf{b}}{4} \tau_k^3 \right\|_\B.
    \end{equation}
    By the definition of tangent cones, \eqref{eq:sub-in-tan} witnesses that 
    \begin{equation} \label{eq:in-tang}
          \frac{\textbf{b}}{\|\textbf{a}\|_\B^2} + \eta \textbf{a} \in  TC_{\textbf{a} / \| \textbf{a} \|_\B}\!\!\left(\partial \mathcal{B}\right).
    \end{equation}
    Indeed, in the definition \eqref{eq:def-tan-cone}, take $\mathcal{Y} = \partial \mathcal{B} \subseteq \mathbb{R}^D$; $\textbf{y} = \textbf{a} / \| \textbf{a} \|_\B \in \partial \mathcal{B}$; $\textbf{y}_k = \frac{\textbf{a}}{\|\textbf{a}\|_\B} \, + \, \left( \frac{1}{2}\eta_k \textbf{a} + \frac{\textbf{b}}{\|\textbf{a}\|_\B^2}\right)\!\tau_k + \frac{\eta_k \textbf{b}}{\|\textbf{a}\|_\B} \tau_k^2 + \frac{\eta_k^2 \textbf{b}}{4} \tau_k^3 \in \partial \mathcal{B}$; the same $\tau_k$; and $\textbf{d} = \frac{\textbf{b}}{\|\textbf{a}\|_\B^2} +  \eta \textbf{a}$.
    Then, \eqref{eq:in-tang} follows because $(\textbf{y}_k - \textbf{y})/\tau_k = \left( \frac{1}{2} \eta_k \textbf{a} + \frac{\textbf{b}}{\|\textbf{a}\|_\B^2} \right) + \frac{\eta_k \textbf{b}}{\|\textbf{a}\|_\B} \tau_k + \frac{\eta_k^2 \textbf{b}}{4} \tau_k^2 \rightarrow \textbf{d}$ as $k \rightarrow \infty$,  using $\frac{1}{2} \eta_k \rightarrow \eta$, $\tau_k \rightarrow 0$ and $\eta_k = O(1)$ (Lemma \ref{lem:bound-r*}) as $k \rightarrow \infty$.
    From Proposition/Definition~\ref{def:tilt-const} and the membership~\eqref{eq:in-tang}, we obtain $\eta = \tilt(\widehat{\textbf{s}})$.  \hfill \qed
\end{proof}

\noindent \textit{Finishing the proof of Theorem~\textup{\ref{thm:limit}}}:
From the end of Section~\ref{subsec:drop-stuff}, it remains to establish \eqref{eq:I1-integral}.  Let $\sigma \leq \min\left(\sigma_5, \sigma_6, \sigma_7, \sigma_8 \right)$.  Then using spherical coordinates:
\begin{align*}
    & \frac{1}{\sigma^{d+2}} \int_{\textbf{s} \in \approxtwo} \textbf{s} d\textbf{s} \nonumber \\[8pt] 
    & =  \frac{1}{\sigma^{d+2}} \int_{\widehat{\textbf{s}} \in \mathbb{S}^{d-1}} \widehat{\textbf{s}}  \int_{r \in \operatorname{RadialDomain}(\widehat{\textbf{s}}, \sigma)} r^{d} dr d\widehat{\textbf{s}} \hspace{12.2em} \textup{[\eqref{eq:hard-int-sph}]} \nonumber \\[8pt]
      & = \frac{1}{\sigma^{d+2}} \int_{\widehat{\textbf{s}} \in \mathbb{S}^{d-1}} \widehat{\textbf{s}} \int_{r=0}^{r^*(\widehat{\textbf{s}}, \sigma)} r^d dr d\widehat{\textbf{s}} \hspace{12.7em} \textup{[\eqref{eq:def-r*-rad}, $\sigma \leq \sigma_5$]}. \nonumber 
\end{align*}
Substituting Eq.~\eqref{eq:eta-def} for $r^{*}(\widehat{\textbf{s}}, \sigma)$ and evaluating the inner integral, we obtain

\begin{align*}
   &  \frac{1}{\sigma^{d+2}} \int_{\widehat{\textbf{s}} \in \mathbb{S}^{d-1}}  \frac{\widehat{\textbf{s}}}{d+1} \, \left( \frac{1}{\|L_{\textbf{p}}(\widehat{\textbf{s}})\|_\B} \sigma + \frac{1}{2} \eta^*(\widehat{\textbf{s}},\sigma)\sigma^2\right)^{d+1} \!\!\! d\widehat{\textbf{s}} \hspace{3.1em} \textup{[\eqref{eq:eta-def}, $\sigma \leq \sigma_8$]} \nonumber \\[8pt]
   &  =  \frac{1}{(d+1)\sigma} \int_{\widehat{\textbf{s}} \in \mathbb{S}^{d-1}}  \frac{\widehat{\textbf{s}} }{\|L_{\textbf{p}}(\widehat{\textbf{s}})\|_\B^{d+1}} d\widehat{\textbf{s}} \nonumber \nonumber \\[2pt] 
     &    + \int_{\widehat{\textbf{s}} \in \mathbb{S}^{d-1}} \widehat{\textbf{s}} \left( \frac{\tfrac{1}{2} \eta^*(\widehat{\textbf{s}},\sigma)}{\|L_{\textbf{p}}(\widehat{\textbf{s}})\|_\B^d} + O(\sigma) \right) d\widehat{\textbf{s}} \quad \quad \quad \,\, \textup{[\eqref{eq:bound-r*}, $\sigma \leq \sigma_7$, $\|L_{\textbf{p}}(\widehat{\textbf{s}})\|_\B^{-1} = \omega(1)$]}\nonumber \\[8pt]
      & =  \int_{\widehat{\textbf{s}} \in \mathbb{S}^{d-1}} \widehat{\textbf{s}} \left( \frac{\tfrac{1}{2} \eta^*(\widehat{\textbf{s}},\sigma)}{\|L_{\textbf{p}}(\widehat{\textbf{s}})\|_\B^d} + O(\sigma) \right) d\widehat{\textbf{s}}. \hspace{14em} \textup{[oddness]}
\end{align*}
By Eq.~\eqref{eq:tilt-lim-eq} and dominated convergence, as $\sigma \to 0$ this integral converges to
\begin{align*}
    \int_{\widehat{\textbf{s}} \in \mathbb{S}^{d-1}}
       \widehat{\textbf{s}} \,\| L_{\textbf{p}}(\widehat{\textbf{s}}) \|_\B^{-d}
   \, \lim_{\sigma \rightarrow 0} \frac{1}{2} \eta^*(\widehat{\textbf{s}},\sigma) \, d\widehat{\textbf{s}}
    =
    \int_{\widehat{s} \in \mathbb{S}^{d-1}} \widehat{\textbf{s}}
        \| L_{\textbf{p}}(\widehat{\textbf{s}}) \|_\B^{-d} \tilt(\widehat{\textbf{s}})\, d\widehat{\textbf{s}}. \nonumber 
\end{align*}
The use of dominated convergence is justified because $\widehat{\textbf{s}} \mapsto \frac{1}{2} \eta^{*}(\widehat{\textbf{s}},\sigma)$ is continuous in $\widehat{\textbf{s}}$ for each $\sigma \leq \sigma_6$ (Lemma~\ref{lem:continuous}), it is uniformly bounded in absolute value by $c_8$ for each $\sigma \leq \sigma_7$ (Lemma~\ref{lem:bound-r*}), and $\lim_{\sigma \rightarrow 0} \tfrac{1}{2} \eta^*(\widehat{\textbf{s}}, \sigma)$ exists and is $\tilt$ (Proposition~\ref{prop:obtain-tilt}).  This completes the proof of Theorem~\ref{thm:limit}.  \hfill \qed

\section{Application: mapping volumetric shape spaces} \label{sec:experiments}

In this section, we demonstrate the use of non-Euclidean norms for embedding a set of 3D densities with continuous variability.
The specific motivation comes from the field of single-particle cryo-electron microscopy (cryo-EM), an  imaging technique for reconstructing the 3D structure of proteins, ribosomes, and other large molecules, using a large set of electron-microscope images.
We give a description of cryo-EM and the continuous heterogeneity problem, which naturally lends itself to manifold learning. We then describe our method for mapping general volumetric shape spaces using non-Euclidean diffusion maps, and apply it to a simulated data set that satisfies the assumptions of Theorem~\ref{thm:nonuniform}..
For a broader general introduction to cryo-EM see Chapter 1 of \cite{Glaeser2021} or the more mathematically-oriented reviews
\cite{SingerSigworth2020,BendoryBartesaghiSinger2020}.  
Code for reproducing the numerical results is available at:\\
\indent \url{http://github.com/mosco/manifold-learning-arbitrary-norms}

\subsection{\textbf{Single-particle cryo-EM}}

\begin{figure}
    \includegraphics[width=0.5\linewidth]{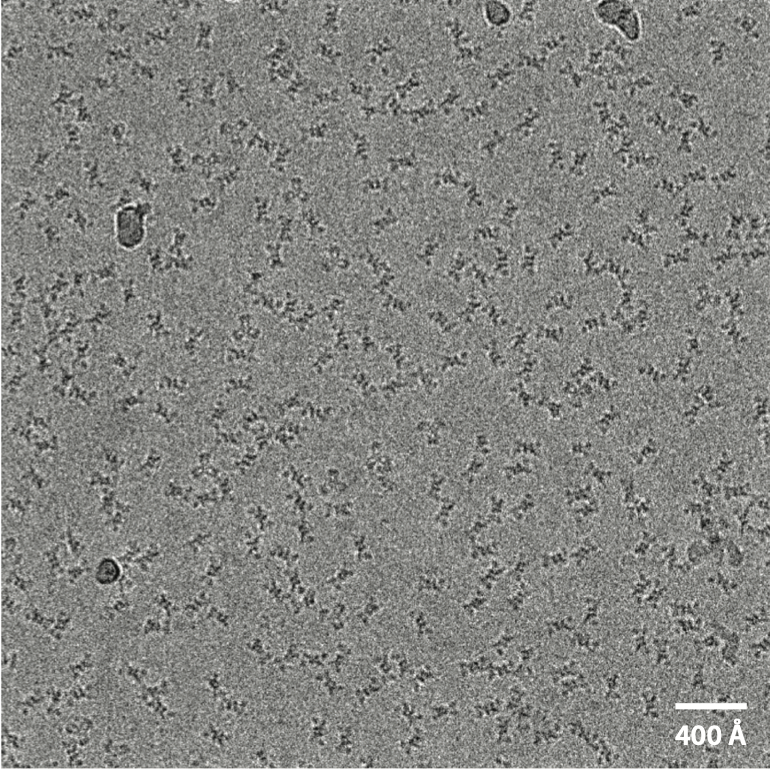}
    \includegraphics[width=0.55\linewidth]{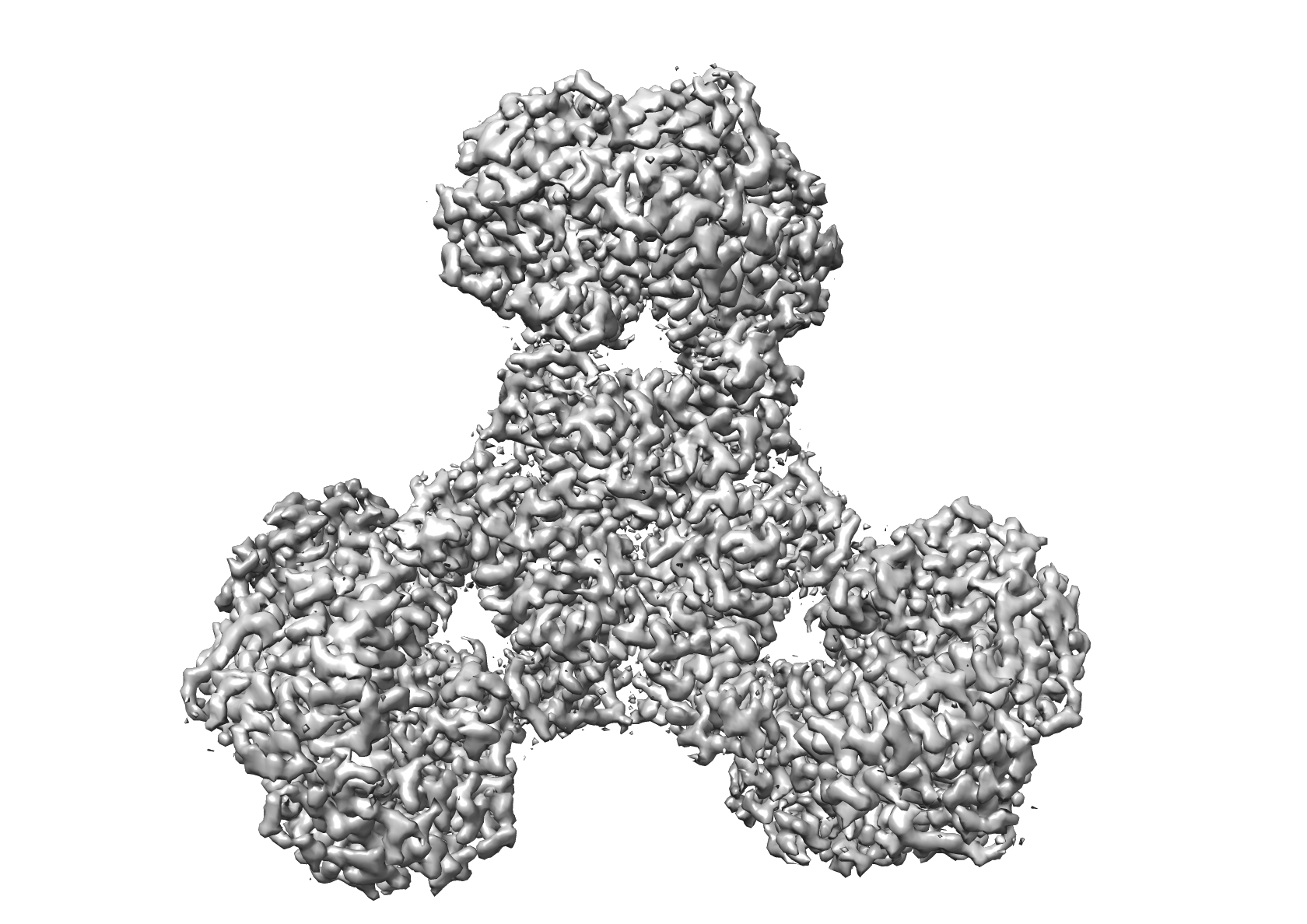}
    \caption{(left) Cryo-EM image showing $\sim \!\!220$ noisy tomographic projections of the PaaZ molecule and some contaminants  (from~\cite{SingerSigworth2020}); (right) Surface plot of the reconstructed electrostatic density of the PaaZ molecule, based on 118,203 tomographic projections (from~\cite{SathyanarayananEtal2019}). } \label{fig:PaaZ_from_micrograph_to_reconstruction}
\end{figure}

The goal of single-particle cryo-EM is to obtain the 3D structure of a molecule of interest.
This is done by obtaining a sample of the molecule and freezing it so that it forms a thin sheet of ice.
This sheet typically contains hundreds of thousands of copies of the molecule, each suspended at a different orientation.
The frozen sample is then imaged using a transmission electron microscope. This results in images that contain many noisy tomographic projections of the same molecule, viewed from different (unknown) directions (See Figure~\ref{fig:PaaZ_from_micrograph_to_reconstruction}).
The challenge is to compute a 3D reconstruction of the electrostatic density map.
The field of cryo-EM has made such progress over the last decade that it is now common to see reconstructions of large rigid molecules, composed of tens of thousands of individual atoms, with resolutions finer than 3~\r{a}ngstr\"{o}ms, which allow for the accurate fitting of atomic models using specialized software.
See the right panel of  Figure~\ref{fig:PaaZ_from_micrograph_to_reconstruction} for an example experimental reconstruction.

The basic assumption behind most single-particle cryo-EM methods is that the molecule of interest is \textit{rigid}.
Hence, the different electron microscope images are tomographic projections of the same exact 3D volume from different angles (or the at least, there is only a finite set 3D volumes).
However, this assumption does not always hold: some molecules have flexible components that can move independently.
This fact, known as the \emph{continuous heterogeneity problem} in cryo-EM, poses a difficulty for existing  reconstruction methods. 
One of the key ongoing challenges in the field is the development of new methods that can map the entire space of molecular conformations \cite{Frank2018,JinEtal2014,TagareEtal2015,FrankOurmazd2016,NakaneEtal2018,DashtiEtal2020,LedermanAndenSinger2020,ZhongEtal2021,PunjaniFleet2021}.
See \cite{SorzanoEtal2019} for a survey.
Several works have applied diffusion maps to this problem domain \cite{DashtiEtal2014,SchwanderFungOurmazd2014,DashtiEtal2020,MoscovichHaleviAndenSinger2020}.
In our conference paper \cite{ZeleskoMoscovichKileelSinger2020}, we applied diffusion maps with a particular non-Euclidean norm to a given set of 3D densities.   Specifically, we used a fast wavelet-based approximation to the Earthmover's distance (WEMD).
Those numerical results were the original motivation for the present paper, and the rest of Section~\ref{sec:experiments} extends them.

\subsection{\textbf{WEMD-based diffusion maps}}
Given a set of volumetric arrays $\x_1, \ldots, \x_n \in \mathcal M \subseteq \mathbb{R}^{N_x \times N_y \times N_z}$,
we compute an approximate Earthmover's distance between all pairs of arrays \cite{ShirdhonkarJacobs2008}.
This is done by first computing the discrete wavelet transform of each input and then using a weighted $\ell_1$-norm on the pairwise differences of wavelet coefficients,
\begin{equation}\label{eq:wemd}
    \|\x_i - \x_j\|_{\textup{WEMD}} := \sum_{\lambda} 2^{-5s/2} \, \lvert \mathcal{W}\x_i(\lambda) - \mathcal{W}\x_j(\lambda) \rvert.
\end{equation}
Here, $\mathcal{W}\x$ denotes a 3D wavelet transform of $\x$ \cite{Mallat2009}.
The index $\lambda$ contains the wavelet shifts $(m_1,m_2, m_3) \in \mathbb{Z}^3$ and scale parameter $s \in \mathbb{Z}_{\ge 0}$.
We then compute pairwise Gaussian affinities,
\begin{align} \label{eq:gaussian_kernel}
    W_{ij} = \exp\left(-\|\x_i - \x_j\|_{\textup{WEMD}}^2 \big{/}  \sigma^2\right),
\end{align}
proceed to construct a graph Laplacian, and perform the eigenvector-based embedding as described in Section~\ref{subsec:graphLaplacian}.
Since the construction uses a (fixed) norm, the theory described in Section~\ref{subsec:thm-statement} applies in this case. 
Hence, in the noiseless case, the graph Laplacian converges to an elliptic second-order differential operator on the relevant manifold $\M$ of ATP synthase conformations.  Here, $\M$ is embedded in the Euclidean space of arrays of size $N_x \times N_y \times N_z$.

\subsection{\textbf{Simulation results}}

\begin{figure}
    \centering
    \begin{tikzpicture}
        \node (ATPpartial) at (0,0) {\includegraphics[height=2in]{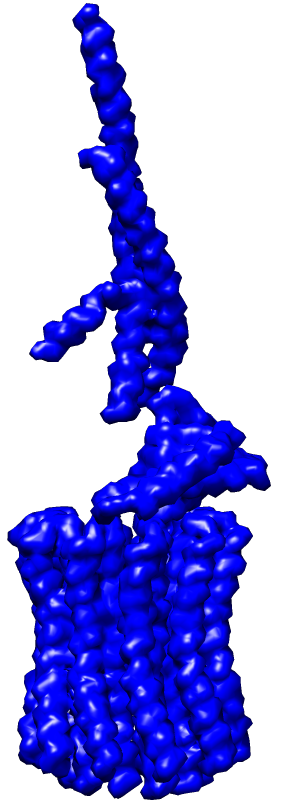}};  
        \node (ATPall) at (3.5,0.44) {\includegraphics[height=2.35in]{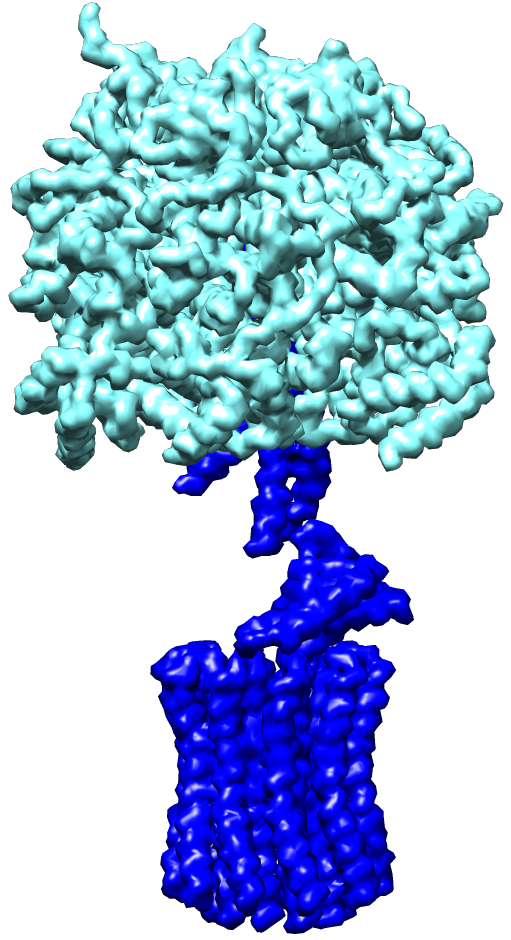}};
        \node (NoisySlice) at (7.5,0.0) {\includegraphics[height=2in]{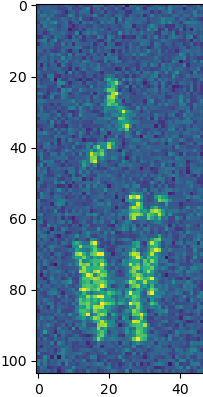}};
        \draw[x=.25cm,y=0.60cm,line width=.2ex,stealth-,rotate=+90] (11,0.1)  arc (-150:150:1 and 1);
    \end{tikzpicture}
    \caption{\textit{ATP synthase.} (left) F$_0$ and axle subunits. These jointly rotate in the presence of hydrogen ions,
together forming a molecular electric motor; (middle) the  F$_1$ subunit (in cyan) envelops the axle. As the axle rotates, the F$_1$ subunit
assembles ATP; (right) representative $2$D slice of the rotated F$_0$ and axle subunits with additive noise shown. }
    \label{fig:atp_synthase}
\end{figure}

\begin{figure}
    \center
    \includegraphics[width=0.95\linewidth]{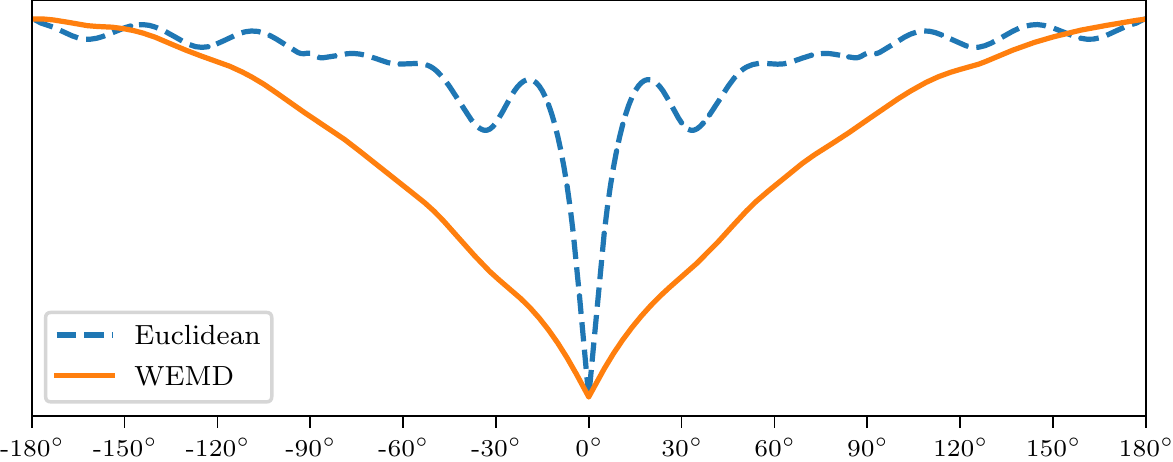}
    
    \vspace{8pt}
    
    \includegraphics[width=0.95\linewidth]{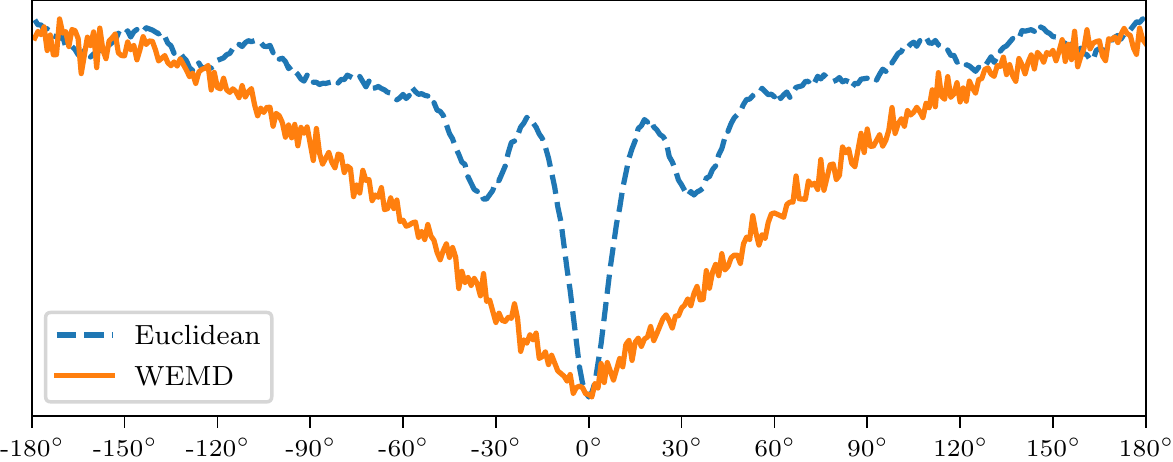}
    \caption{\textit{Euclidean distance vs. wavelet-based approximate Earthmover's distance} as functions of the angle between rotations of the ATP synthase
rotor. (top) distances for rotated volumes without noise;  (bottom) distances for the noisy data set. 
 (Euclidean distances were scaled to be comparable to WEMDs.)}
    \label{fig:compare_angle}
\end{figure}

We tested our method on a synthetic volumetric data set that mimics the motion space of ATP synthase \cite{YoshidaMuneyukiHisabori2001}, see Figure~\ref{fig:atp_synthase}.
This enzyme is a  stepper motor with a central asymmetric axle that rotates in 120$\degree$ steps relative to
the F$_1$ subunit, with short transient motions in-between the three dominant states.
Our synthetic data was generated as follows:
we produced 3D density maps  of entry 1QO1 \cite{Stock1999} from the Protein Data Bank \cite{Roseetal.2017} using the \texttt{molmap}
command in UCSF Chimera \cite{Chimera2004}.
These density maps have array dimensions $47 \times 47 \times 107$ and a resolution of 6\AA \ per voxel.
We then took random rotations of the F$_0$ and axle subunits, where the angles were drawn i.i.d. according to the following mixture distribution,
\begin{align*}
    \tfrac{2}{5} U[0,360] + \tfrac{1}{5} \mathcal{N}(0,1) + \tfrac{1}{5} \mathcal{N}(120,1) + \tfrac{1}{5} \mathcal{N}(240,1).
\end{align*}
The resulting density maps formed the clean dataset.
The noisy dataset was generated in the same manner but also included additive i.i.d. Gaussian noise with mean zero and a standard deviation of $1/10$ of the maximum voxel value.

The discrete wavelet transform of all the volumes in the dataset was computed using  PyWavelets \cite{LeeEtal2019} with the \texttt{sym3} wavelet
(symmetric Daubechies wavelets of order 3), though other wavelet choices also worked well \cite[Sec 4.2]{ShirdhonkarJacobs2008}.
The maximum scale level chosen was $s=6$ to minimize the truncation in Eq.~\eqref{eq:wemd}.
The number of resulting wavelet coefficients was 40\% larger than the number of voxels.

Figure~\ref{fig:compare_angle} compares the Euclidean norm to the WEMD norm for a range of angular differences for the noiseless and noisy datasets. 
 Note that  for the clean dataset, WEMD is monotonic in the absolute value of the X axis (equal to the angular difference between the ATP synthase rotors).
This behavior also holds for the Euclidean norm, but only for small angular differences up to $\approx \pm19\degree$.
This suggests that an affinity graph built from this dataset using the Euclidean norm can capture the right geometry only when the dataset contains a dense sampling of the angles and when the kernel width is properly calibrated to nearly cut off connections at angles $>19\degree$.

\begin{figure}
    \hspace{-12pt}
    \begin{tabular}{cccccc}
        n & $\ell_2$ (noiseless) & $\ell_2$ (noisy) & WEMD (noiseless) & WEMD (noisy) \\\\
        \begin{tabular}{c}25\\[44pt] \end{tabular}&
        \includegraphics[height=56pt]{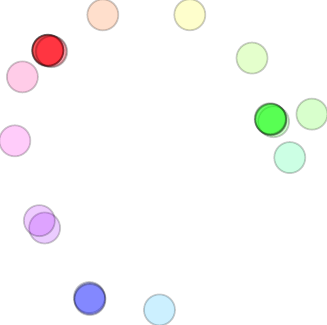}&
        \includegraphics[height=56pt]{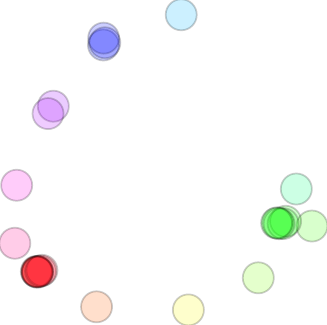}&
        \includegraphics[height=56pt]{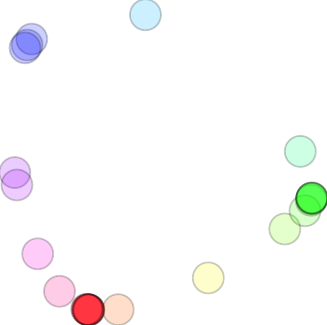}&
        \includegraphics[height=56pt]{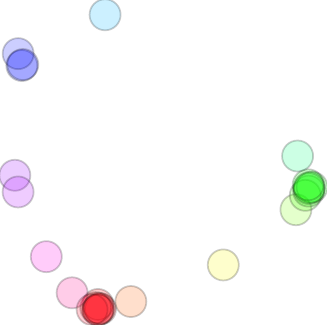}
        \\
        \begin{tabular}{c}50\\[44pt] \end{tabular}&
        \includegraphics[height=56pt]{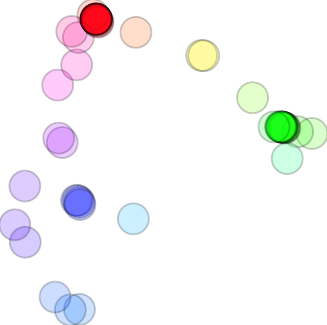}&
        \includegraphics[height=56pt]{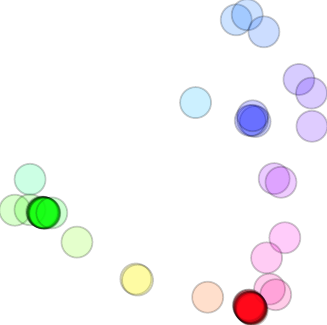}&
        \includegraphics[height=56pt]{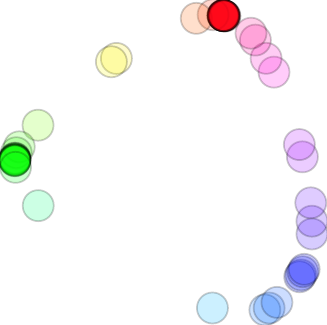}&
        \includegraphics[height=56pt]{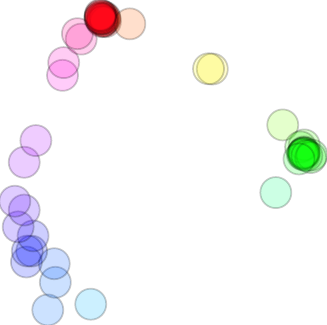}
        \\
        \begin{tabular}{c}100\\[44pt] \end{tabular}&
        \includegraphics[height=56pt]{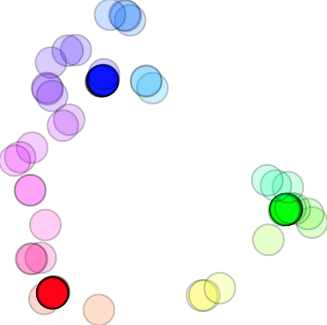}&
        \includegraphics[height=56pt]{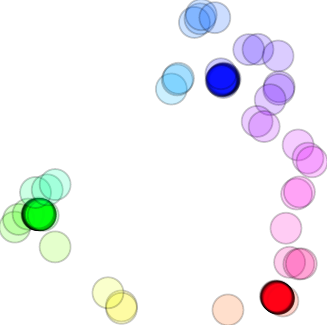}&
        \includegraphics[height=56pt]{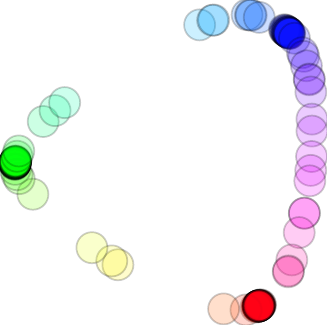}&
        \includegraphics[height=56pt]{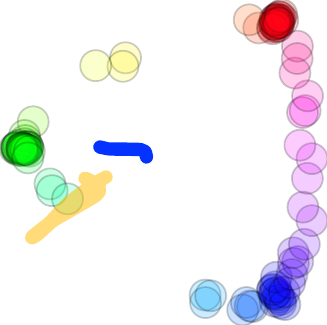}
        \\
        \begin{tabular}{c}200\\[44pt] \end{tabular}&
        \includegraphics[height=56pt]{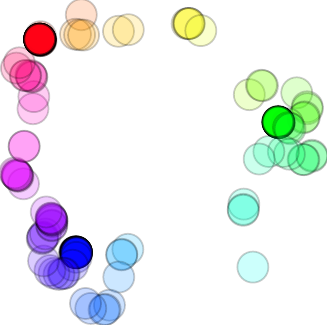}&
        \includegraphics[height=56pt]{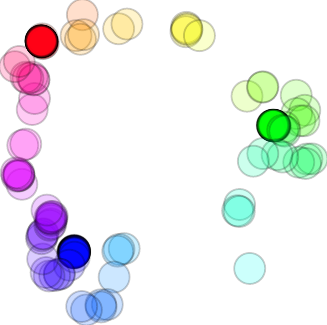}&
        \includegraphics[height=56pt]{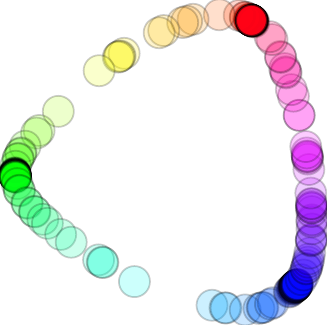}&
        \includegraphics[height=56pt]{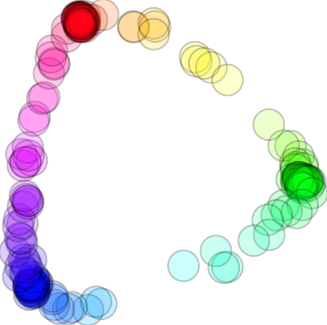}
        \\
        \begin{tabular}{c}400\\[44pt] \end{tabular}&
        \includegraphics[height=56pt]{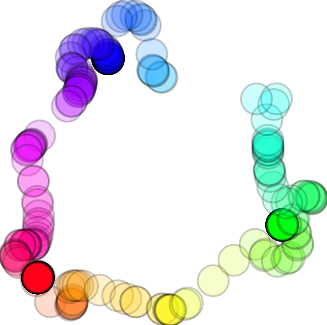}&
        \includegraphics[height=56pt]{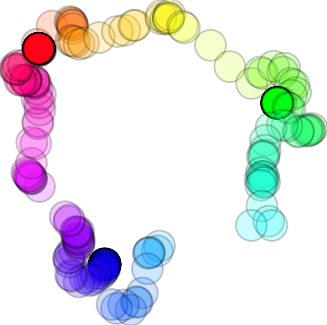}&
        \includegraphics[height=56pt]{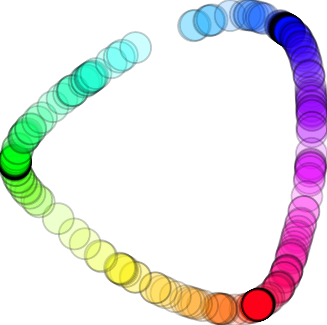}&
        \includegraphics[height=56pt]{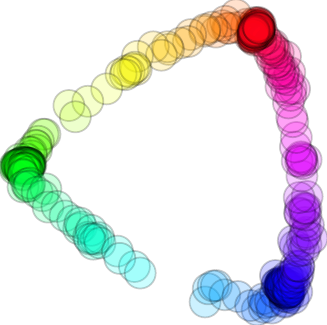}
        \\
        \begin{tabular}{c}800\\[44pt] \end{tabular}&
        \includegraphics[height=56pt]{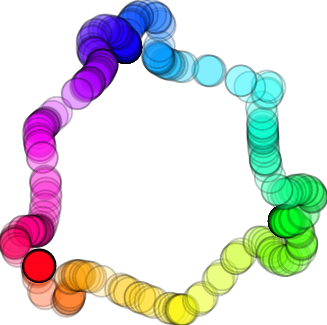}&
        \includegraphics[height=56pt]{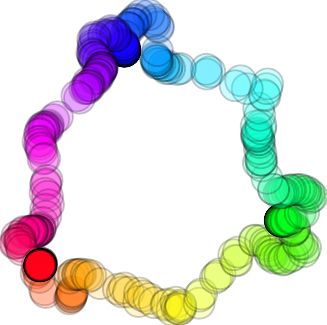}&
        \includegraphics[height=56pt]{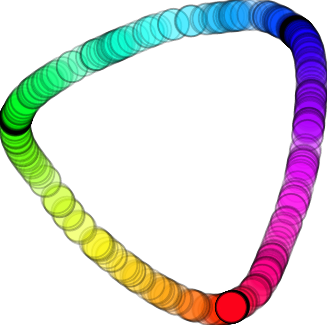}&
        \includegraphics[height=56pt]{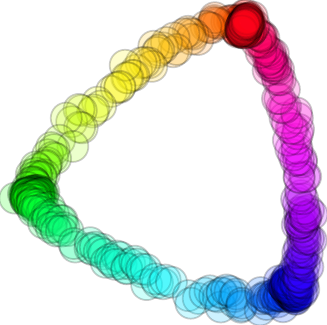}
    \end{tabular}
    \caption{\textit{Simulation results}.  Euclidean vs. WEMD-based Laplacian eigenmaps into $\mathbb{R}^2$ using the clean
and noisy ATP synthase data sets. Sample sizes of $n=25, 50, 100, 200, 400, 800$. The points are translucent to indicate
density and the color is the groundtruth angle.}

    \label{fig:embeddings}
\end{figure}

Figure~\ref{fig:embeddings} is the result of a two-dimensional Laplacian eigenmaps embedding, once with the Euclidean norm and once with WEMD norm \eqref{eq:wemd}.
These embeddings use the unweighted graph Laplacian with a Gaussian kernel, which corresponds to
the setting of Theorem~\ref{thm:limit}. For similar results that use the density normalized diffusion maps of \cite{CoifmanLafon2006},
see \cite[Fig.~5]{ZeleskoMoscovichKileelSinger2020}.
We chose $\sigma=30$ as the Gaussian kernel width in Eq.~\eqref{eq:gaussian_kernel} for the WEMD embeddings,
however the WEMD results were not very sensitive to the particular choice of $\sigma$. In contrast, the Euclidean embeddings
required fine-tuning of $\sigma$ to obtain the best results for each sample size.
This makes sense given the results of Figure~\ref{fig:compare_angle}.

The key takeaway from Figure~\ref{fig:embeddings} is that for the standard Laplacian eigenmaps embedding based on the Euclidean norm, one needs $>400$ samples to conclude that the intrinsic geometry is a circle.
In contrast, for the embeddings based on WEMD, even  small sample sizes give the right geometry.   

\subsection{\textbf{Runtime}}

The running time of the WEMD-based diffusion maps is similar to that of the standard Euclidean diffusion maps.
This follows from the fact that both algorithms need to compute $\binom{n}{2}$ pairwise $\ell_p$-distances ($p \in \{1,2\}$) for vectors of similar length. The cost of the discrete wavelet transform is negligible, since it is linear with respect to the input size. For our sample sizes, the time to form the Gaussian affinity matrix and compute its eigenvectors is also negligible. Table~\ref{fig:runtime} lists single-core running times on an Intel Core i7-8569U CPU.

\begin{table}
\caption{\textit{Running times [sec]} for computing the discrete wavelet transform (DWT), all pairwise wavelet-based Earthmover
approximations (WEMD) not including the DWT, and all pairwise Euclidean ($\ell_2$) distances.}
\label{fig:runtime}
\centering
\begin{tabular}{llll}
    \toprule
    $n$ & DWT& WEMD & $\ell_2$ \\
    \midrule
    25  & 0.3  & 0.13 & 0.09\\
    50  & 0.61 & 0.49 & 0.38\\
    100 & 1.2  & 1.93 & 1.5\\
    200 & 2.4  & 7.6  & 5.5 \\
    400 & 4.9  & 31   & 22\\
    800 & 11   & 126  & 86\\
    \bottomrule
\end{tabular}
\end{table}

\subsection[\textbf{Using wavelet sparsity}]{\textbf{Using wavelet sparsity}} \label{sec:sparsity}

\begin{table}
\caption{\textit{Running times [sec]} for computing all pairs of sparsified wavelet-based Earthmover's distances (sparse-WEMD),
as compared to the dense computation.}
\label{table:runtime_sparse}
\centering
\begin{tabular}{llll}
    \toprule
    $n$ & Sparse runtime & Sparse runtime & Dense runtime \\
        & (noiseless data)          & (noisy data)   &       \\
    \midrule
    25  & 0.01  & 0.037 & 0.13 \\
    50  & 0.013 & 0.1   & 0.49 \\
    100 & 0.026 & 0.39  & 1.93   \\ 
    200 & 0.046 & 1.5   & 7.6 \\
    400 & 0.16  & 6.2   & 31 \\
    800 & 0.6   & 25    & 126 \\
    \bottomrule
\end{tabular}
\end{table}

To compute the approximate Earthmover's distance between all pairs of volumes, we first compute a weighted discrete wavelet transform of each volume in the data set.
For smooth signals, this results in sparse vectors of wavelet coefficients \cite{Mallat2009}.
We can use this property by thresholding the vectors of weighted wavelet coefficients, and then storing them in a sparse matrix.
This is beneficial because computing the $\ell_1$-distance between two sparse vectors has a runtime that is linear in the number of their non-zero elements.
Since the computation of all pairwise $\ell_1$ differences is the slowest part of our procedure, this approach can reduce the running time significantly.
To test this, we used the ATP synthase data described in the previous section.
First, we subtracted the mean volume from all volumes in the data set.
This mean-centering does not change the pairwise WEMD distances but makes the resulting vectors more sparse.

\begin{figure}
    \centering
    \begin{tabular}{cccc}
        $n$ & WEMD (noiseless) & WEMD (noisy) \\\\
        \begin{tabular}{c}25\\[44pt] \end{tabular}&
        \includegraphics[height=56pt]{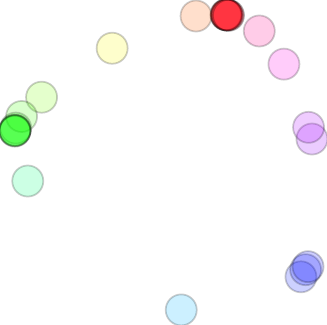}&
        \includegraphics[height=56pt]{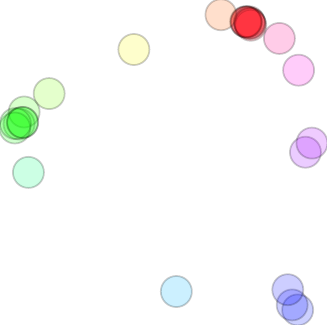}
        \\
        \begin{tabular}{c}50\\[44pt] \end{tabular}&
        \includegraphics[height=56pt]{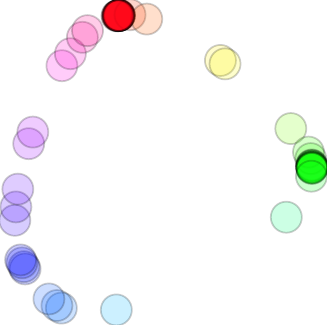}&
        \includegraphics[height=56pt]{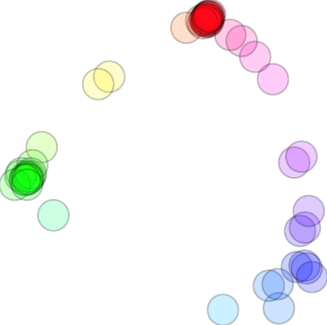}
        \\
        \begin{tabular}{c}100\\[44pt] \end{tabular}&
        \includegraphics[height=56pt]{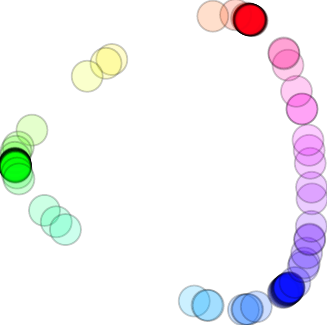}&
        \includegraphics[height=56pt]{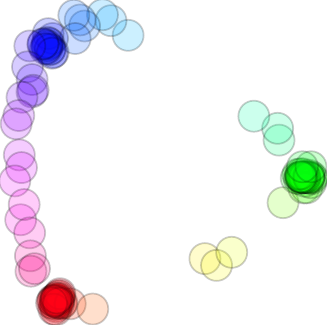}
        \\
        \begin{tabular}{c}200\\[44pt] \end{tabular}&
        \includegraphics[height=56pt]{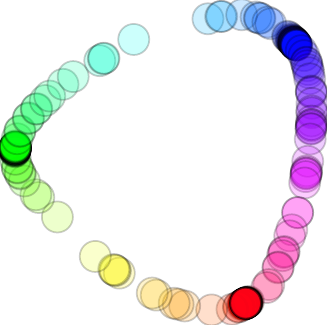}&
        \includegraphics[height=56pt]{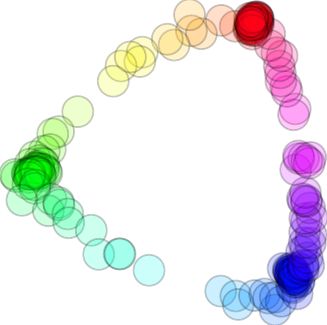}
        \\
        \begin{tabular}{c}400\\[44pt] \end{tabular}&
        \includegraphics[height=56pt]{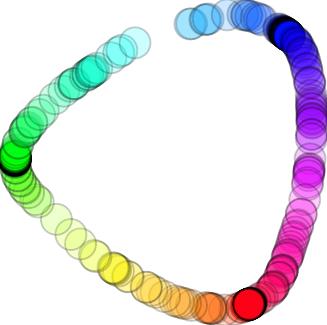}&
        \includegraphics[height=56pt]{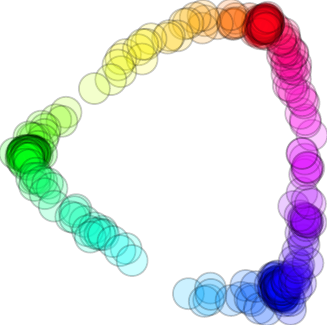}
        \\
        \begin{tabular}{c}800\\[44pt] \end{tabular}&
        \includegraphics[height=56pt]{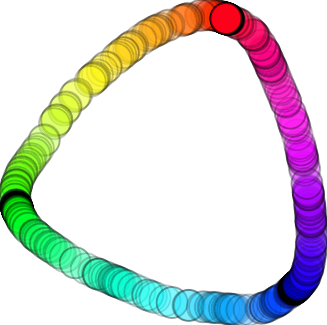}&
        \includegraphics[height=56pt]{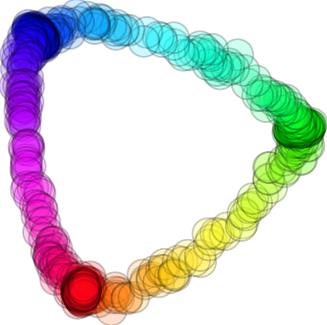}
    \end{tabular}
    \caption{\textit{Sparsified results.} Wavelet-EMD-based Laplacian eigenmaps of the clean and noisy ATP synthase
data sets, after applying hard-thresholding to obtain sparse coefficient vectors.}
    \label{fig:embeddings_sparsified}
\end{figure}

We used the hard-thresholding function $h_t$ defined as follows:
\begin{align*}
    h_t(x)
    :=
    \begin{cases}
        0& \text{for } |x| \leq t,\\
        x& \text{for } |x| > t.
    \end{cases}
\end{align*}
We found a threshold $t$ for the wavelet coefficients such that the $\ell_1$-norm of the post-thresholding weighted wavelet coefficients  are $>90\%$ of the $\ell_1$-norm of the dataset prior to  thresholding.
This threshold was computed on the smallest simulation of size $n=25$ and then applied to the rest of the runs.
Figure~\ref{fig:embeddings_sparsified} shows the results of the WEMD embedding following this sparsification step.
Table \ref{table:runtime_sparse} shows the running times for the sparsified WEMD.
Note that the running times are different for the noiseless and noisy data, since the noisy data is less sparse.
However, in both cases, there are significant gains to the running times, with few visually-noticeable changes to the embedding results. 

\newpage

\section{Conclusion}
In this paper, we placed Laplacian-based manifold learning methods that use non-Euclidean norms on a firmer theoretical footing.  
We proved the pointwise convergence of graph Laplacians computed using general norms to elliptic second-order differential operators.
In particular, our proof involved a novel second-order interaction between the manifold $\mathcal{M}$ and the unit ball $\mathcal{B}$, encoded by the function $\tilt$.
We showed that some properties of the usual Laplace-Beltrami operator are lost in the general case.
The limiting operator $\Delta_{\mathcal{M}, \mathcal{B}}$ changes with the embedding of $\mathcal{M}$.  Further, the limit $\Delta_{\mathcal{M}, \mathcal{B}}$ carries a first-order term that can exhibit discontinuities at certain points of $\mathcal{M}$.

In addition, this paper demonstrated practical advantages for using non-Euclidean norms 
in manifold learning.  
We considered the task of
learning molecular shape spaces.
Here data points are conformations represented by 3D arrays, and we  want  to  capture  the  range of motion. 
A simulation found that using Laplacian eigenmaps with the wavelet Earthmover's distance (a weighted-$\ell_1$ norm in wavelet coefficients) resulted in a qualitative improvement of sample complexity compared to the Euclidean norm.
Thresholding the wavelet coefficients before computing norms 
reduced the computational \nolinebreak cost.\\\\
\noindent This work suggests several directions worthy of future study:

\begin{itemize} \setlength\itemsep{0.7em}
\item \textbf{Convergence rates.} With what rate does the convergence in Theorem~\ref{thm:limit} occur?  How does this depend on the choice of norm $\| \cdot \|_\B$?

\item \textbf{Eigenfunctions.}   What can be said about the eigenfunctions of $\Delta_{\mathcal{M}, \mathcal{B}}$?  How do the discontinuities of the first-order term in $\Delta_{\mathcal{M}, \mathcal{B}}$ impact them?  Due to space limitations, we only gave numerical examples in Appendix~\ref{sec:numerical_eigenfunction_computation}.

\item \textbf{Spectral convergence.} For general norms, do the eigenvectors of the graph Laplacian converge to the eigenfunctions of the operator $\Delta_{\mathcal{M}, \mathcal{B}}$?

\item \textbf{Concentration.} The operator $\Delta_{\mathcal{M}, \B}$ depends on $d$- and $2$-dimensional linear sections of the convex body $\mathcal{B} \subseteq \mathbb{R}^D$.  When $D \gg d$, is there a sense in which these slices look ``increasingly Euclidean"? Does $\Delta_{\mathcal{M}, \B}$ concentrate?

\item \textbf{Data-dependent norms.} If the norm chosen is some  fixed function of the data set, does a well-defined  limit of the graph Laplacians still exist?

\item \textbf{Applications.} 
Are some applied domains better-suited for non-Euclidean norms than others? 
How should a practitioner decide which norm to use? 

\end{itemize}

\vspace{0.5em}

\begin{acknowledgements}
We thank Charles Fefferman, William Leeb, Eitan Levin and John Walker for enlightening discussions.
 Most of this work was performed while AM was affiliated  with PACM at Princeton \nolinebreak University.
This research was supported by AFOSR FA9550-17-1-0291, ARO W911NF-17-1-0512, NSF BIGDATA
 IIS-1837992, the Simons Investigator Award, the Moore Foundation Data-Driven Discovery Investigator Award, the Simons Collaboration on Algorithms
and Geometry, and start-up grants from the College of Natural Sciences and Oden Institute for Computational Engineering and Sciences at UT Austin.
\end{acknowledgements}

\appendix

\section*{Appendices}

\section[Appendix: Proof of Lemma~\ref{prop:tan-con-boundary}]{Proof of Lemma~\ref{prop:tan-con-boundary}} \label{app:tan-con-boundary}

 \underline{\textbf{Step~1: LHS $\subseteq$ RHS}}.
 By the identity~\eqref{eq:nice-tan-cone} for tangent cones of convex sets, we have 
 \begin{align*}
     TC_{\y}(\B) = \overline{\mathbb{R}_{>0}(\B - \y)}.
 \end{align*}
 
    By definition of $\partial$ and of tangent cones \eqref{eq:nice-tan-cone}, the LHS of Eq.~\eqref{eq:annoying} is  
    \begin{equation} \label{eq:partial-TC}
       \partial \left( TC_{\textbf{y}}(\mathcal{B}) \right) = \overline{\mathbb{R}_{>0} (\mathcal{B} - \textbf{y})} \setminus \left( \mathbb{R}_{>0} (\mathcal{B} - \textbf{y}) \right)^{\circ}.
    \end{equation}
    
    Let $\textbf{d} \in \partial \left( TC_{\textbf{y}}(\mathcal{B}) \right)$.  By Eq.~\eqref{eq:partial-TC},
      $\textbf{d} = \lim_{k \rightarrow \infty} \beta_k (\widetilde{\textbf{y}}_k - \textbf{y})$ for some $\beta_k \in \mathbb{R}_{>0}$ and $\widetilde{\textbf{y}}_k \in \mathcal{B}$.  Without loss of generality, we assume $\widetilde{\textbf{y}}_k \in \partial \mathcal{B}$ for each $k$. 
We break into cases.
      
      \begin{itemize}\setlength\itemsep{2em}
      \item  \textit{Case~A: $\widetilde{\textbf{y}} = \textbf{y}$.}   
      
      \vspace{0.1em}
      
      Either $\textbf{d} = 0 \in T_{\textbf{y}}(\partial \mathcal{B})$, or $\tau_k := 1 / \beta_k \rightarrow \infty$ as $k \rightarrow \infty$.  If the latter,  the sequences $(\widetilde{\textbf{y}}_k)_{k=1}^{\infty} \subseteq \partial \mathcal{B}$ and $(\tau_k)_{k=1}^{\infty}\subseteq \mathbb{R}_{>0}$ witness $\textbf{d} \in TC_{\textbf{y}}(\partial \mathcal{B})$.

    \item \textit{Case~B:  $\widetilde{\textbf{y}} \neq \textbf{y}$.}  
    
    \vspace{0.1em}
    
    Here,  $\lim_{k \rightarrow \infty} \beta_k =: \beta \in \mathbb{R}_{\geq 0}$ exists, and 
     $\textbf{d} = \beta (\widetilde{\textbf{y}} - \textbf{y})$.  If $\beta =0$, then $\textbf{d} = 0 \in T_{\textbf{y}}(\partial \mathcal{B})$.  
     Suppose $\beta \neq 0$.
Let the line segment joining $\widetilde{\textbf{y}}$ and $\textbf{y}$ be
    \begin{align*}
    \texttt{conv}\{\widetilde{\textbf{y}}, \textbf{y}\} := \{\alpha \widetilde{\textbf{y}} + (1 - \alpha)\textbf{y} \in \mathbb{R}^D : \alpha \in [0,1]\}.
    \end{align*}
     So, $\texttt{conv}\{\widetilde{\textbf{y}}, \textbf{y}\} \subseteq \mathcal{B}$.  We claim  $\texttt{conv}\{\widetilde{\textbf{y}}, \textbf{y}\} \subseteq  \partial \mathcal{B}$.
    Assume not.  That is,
     \begin{align*}
         \exists \, \alpha \in (0,1) \, \textup{ such that } \, \textbf{z} := \alpha \widetilde{\textbf{y}} + (1 - \alpha) \textbf{y} \in \mathcal{B}^{\circ}.
     \end{align*}
     But then,
     \begin{align*}
         \textbf{d} \, = \, \beta (\widetilde{\textbf{y}} - \textbf{y}) \, = \,  (\beta / \alpha) (\textbf{z} - \textbf{y}) \, \in \, \mathbb{R}_{>0}\left(\mathcal{B}^{\circ} - \textbf{y}\right) \, \subseteq \, \left( \mathbb{R}_{>0} (\mathcal{B} - \textbf{y}) \right)^{\circ}.
     \end{align*}
     This contradicts $\textbf{d} \in \partial\left( TC_{\textbf{y}}(\mathcal{B}) \right)$ (see Eq.~\eqref{eq:partial-TC}).  So, indeed $\texttt{conv}\{\widetilde{\textbf{y}}, \textbf{y}\} \subseteq  \partial \mathcal{B}$.  Now, define 
     \begin{align*}
     \widehat{\textbf{y}}_k := \frac{1}{k} \widetilde{\textbf{y}}  +  (1 - \frac{1}{k}) \textbf{y} \in \partial \mathcal{B}  \,\,\,\,\,\,\,\, \textup{and} \,\,\,\,\,\,\,\, \tau_k := \frac{1}{k} \in \mathbb{R}_{>0}. 
     \end{align*}
     Then, $\frac{\widehat{\textbf{y}}_k - \textbf{y}}{\tau_k}  = \textbf{d}$ for each $k$, and  $(\widehat{\textbf{y}}_k)_{k=1}^{\infty}$ and $(\tau_k)_{k=1}^{\infty}$ witness $\textbf{d} \in TC_{\textbf{y}}(\partial \mathcal{B})$.
     \end{itemize}
     In all cases, we have verified $\textbf{d} \in TC_{\textbf{y}}(\partial \mathcal{B})$.  This gives LHS $\subseteq$ RHS in \eqref{eq:annoying}.
     
     \medskip
     \medskip
     
     \noindent \underline{\textbf{Step 2: LHS $\supseteq$ RHS.}}   Let $\textbf{d} \in TC_{\textbf{y}}(\partial \mathcal{B})$.  By the definition of tangent cones \eqref{eq:def-tan-cone}, $\textbf{d} = \lim_{k \rightarrow \infty} \tau_k^{-1} \left(\widetilde{\textbf{y}}_k - \textbf{y}\right)$ for some $\tau_{k} \in \mathbb{R}_{>0}$ and $\widetilde{\textbf{y}}_k \in \partial \mathcal{B}$
     with $\tau_k \rightarrow 0$ and $\widetilde{\textbf{y}}_k \rightarrow \textbf{y}$ as $k \rightarrow \infty$.  By \eqref{eq:partial-TC}, we need to show $\textbf{d} \notin \left(\mathbb{R}_{>0} (\mathcal{B} - \textbf{y}) \right)^{\circ}$.
     
     First, we will prove 
     $\texttt{conv}\{\textbf{d}+\textbf{y}, \textbf{y}\} \cap \mathcal{B}^{\circ} = \emptyset$.
      Assume not, i.e.,
     \begin{align*}
         \exists \, \alpha \in (0,1) \, \textup{ such that } \, \textbf{z} := \alpha (\textbf{d}+\textbf{y}) + (1 - \alpha) \textbf{y} = \alpha \textbf{d} + \textbf{y} \in \mathcal{B}^{\circ}.
     \end{align*}
     Let $\widehat{\tau}_k = \tau_k / \alpha \in \mathbb{R}_{>0}$, so that
     \begin{equation} \label{eq:alpha-d}
        \alpha \textbf{d} = \lim_{k \rightarrow \infty} \widehat{\tau}_k^{-1}\left(\widetilde{\textbf{y}}_k - \textbf{y}\right).
     \end{equation}
     Since $\mathcal{B}^{\circ}$ is open, there exists $\delta > 0$ with 
     \begin{align*}
        \mathcal{N} := \{ \textbf{w} \in \mathbb{R}^D : \| \textbf{w} - \textbf{z} \|_2 \leq \delta \} \subseteq \mathcal{B}^{\circ}.
     \end{align*}
     By Eq.~\eqref{eq:alpha-d}, there exists $K$ such that for all $k \geq K$, 
     \begin{align*}
         \widehat{\tau}_k^{-1} (\widetilde{\textbf{y}}_k - \textbf{y}) + \textbf{y} \in \mathcal{N}.
     \end{align*}
     
  On the other hand, it is easy to see for each $\textbf{w} \in \mathcal{B}^{\circ}$, 
  \begin{equation} \label{eq:weird1}
      \left( \textbf{y} + \mathbb{R}_{\geq 0} (\textbf{w} - \textbf{y}) \right) \cap \mathcal{B} \, = \, \texttt{conv}\{\textbf{y}, \textbf{w}'\} 
  \end{equation}
  for some $\textbf{w}' \in \partial \mathcal{B}$, using convexity and compactness of $\mathcal{B}$.  In addition,
  \begin{equation} \label{eq:weird2}
       \left( \textbf{y} + \mathbb{R}_{\geq 0} (\textbf{w} - \textbf{y}) \right) \cap \partial\mathcal{B}  = \{\textbf{y}, \textbf{w}'\},
  \end{equation}
   using $\textbf{w} \in \texttt{conv}\{\textbf{y}, \textbf{w}'\}$, $\| \textbf{w} \|_\B < 1$, and the triangle inequality for $\| \cdot \|_\B$.
   Clearly, 
   \begin{equation} \label{eq:weird3}
   \| \textbf{w}' - \textbf{y} \|_2 > \|\textbf{w} - \textbf{y} \|_2.
   \end{equation}
    
    Now, let $\epsilon := \min_{\textbf{w} \in \mathcal{N}} \| \textbf{w} - \textbf{y} \|_2$.  Note $\epsilon > 0$.  For each $k \leq K$, we apply \eqref{eq:weird1}, \eqref{eq:weird2} to $\textbf{w} = \widehat{\tau}_k^{-1}(\widetilde{\textbf{y}}_k - \textbf{y}) + \textbf{y} \in \mathcal{N}$.  Then, $\textbf{w}' = \widetilde{\textbf{y}}_k$.  By \eqref{eq:weird3},
    \begin{equation} \label{eq:contra}
        \| \widetilde{\textbf{y}}_k - \textbf{y} \|_2 > \| \textbf{w} - \textbf{y} \|_2 \geq \epsilon \,\,\,\,\, \textup{for all }  k \geq K.
    \end{equation}
     But \eqref{eq:contra} contradicts $\widetilde{\textbf{y}}_k \rightarrow \textbf{y}$ as $k \rightarrow \infty$.  
     Therefore, $\texttt{conv}\{\textbf{d}+\textbf{y},\textbf{y}\} \cap \mathcal{B}^{\circ} = \emptyset$.   
     
     Translating by $-\textbf{y}$,  $\texttt{conv}\{\textbf{d}, 0\} \cap (\mathcal{B} - \textbf{y})^{\circ} = \emptyset$.
     By this and convexity, it follows there exists a \textup{properly separating hyperplane}:
     \begin{multline}
     \exists \, \textbf{v} \in \mathbb{R}^D \setminus \{0\},  \exists \, \gamma \in \mathbb{R} \, \textup{ such that }  \, \forall \, \textbf{u}_1 \in \texttt{conv}\{\textbf{d}, 0\}, \forall \, \textbf{u}_2 \in \mathcal{B} - \textbf{y}  \\
     \langle \textbf{v}, \textbf{u}_1 \rangle \geq \gamma, \langle \textbf{v}, \textbf{u}_2 \rangle \leq \gamma  \textup{ and  } \, \exists \, \widetilde{\textbf{u}}_2 \in \mathcal{B} - \textbf{y} \, \textup{ such that } \langle \textbf{v}, \widetilde{\textbf{u}}_2 \rangle < \gamma. \nonumber
     \end{multline}
     In particular, 
     \begin{align*}
     \mathbb{R}_{>0} (\mathcal{B} - \textbf{y}) \subseteq \{\textbf{u} \in \mathbb{R}^D : \langle \textbf{v}, \textbf{u} \rangle \leq \gamma\}.
     \end{align*}
     Also, for any open neighborhood $\mathcal{D} \subseteq \mathbb{R}^D$ with $\textbf{d} \in \mathcal{D}$,
     \begin{align*}
         \exists \, \widetilde{\textbf{d}} \in \mathcal{D} \, \textup{ such that } \langle \textbf{v}, \widetilde{\textbf{d}} \rangle >  \langle \textbf{v}, \textbf{d} \rangle  \geq \gamma.
     \end{align*}
     We conclude $\textbf{d} \notin \left(\mathbb{R}_{>0} (\mathcal{B} - \textbf{y}) \right)^{\circ}$, as desired. This gives $\textbf{d} \in \partial \left( TC_{\textbf{y}}(\mathcal{B})\right)$, and LHS $\supseteq$ RHS in Eq.~\eqref{eq:annoying}, completing the proof of the lemma. \hfill \qed

\section[Appendix: Proof of Proposition~\ref{prop:simple-title-C1}]{Proof of Proposition~\ref{prop:simple-title-C1}} \label{app:simple-title-C1}
For item 1, we first note that $\operatorname{grad}\| \cdot \|_{\mathcal{B}}(\widehat{\textbf{a}})$ is nonzero, since the directional derivative of the norm function at $\widehat{\textbf{a}}$ in the direction of $\widehat{\textbf{a}}$ is nonzero.  Indeed the function $\mathbb{R} \rightarrow \mathbb{R}; \lambda \mapsto \| \widehat{\textbf{a}} + \lambda \widehat{\textbf{a}} \|_{\mathcal{B}}$ has derivative $\| \widehat{\textbf{a}} \|_{\mathcal{B}} = 1$ at $\lambda = 0$, using homogeneity of $\| \cdot \|_{\mathcal{B}}$ under positive scaling.  
Item 1 now follows immediately from \cite[Thm.~3.15]{nonlinear-optim-book} and the preceding paragraph in that reference that metric regularity is implied by the linear independence of the gradients.   

For item 2, we note that due to homogeneity of the norm, since $\| \cdot \|_{\mathcal{B}}$ is $C^1$ around $L_{\textbf{p}}(\widehat{\s})$, it is also $C^1$ around $L_{\textbf{p}}(\widehat{\s}) / \| L_{\textbf{p}}(\widehat{\s}) \|_{\mathcal{B}}$ and it holds
\begin{align*}
\operatorname{grad} \| \cdot \|_{\mathcal{B}}(L_{\textbf{p}}(\widehat{\textbf{s}})/\|L_{\textbf{p}}(\widehat{\textbf{s}})\|_{\mathcal{B}}) = (1/\| L_{\textbf{p}}(\widehat{\s})\|_{\mathcal{B}}) \operatorname{grad} \| \cdot \|_{\mathcal{B}}(L_{\textbf{p}}(\widehat{\textbf{s}})).
\end{align*}
Thus, item 1 applies and implies the tangent cone in right-hand side of Eq.~\eqref{eq:tilt-def} is the hyperplane normal to $L_{\textbf{p}}(\widehat{\s})$.
Now we finish by equating the inner product of $\operatorname{grad}\| \cdot \|_{\mathcal{B}}(L_{\p}(\widehat{\s}))$ and the LHS of Eq.~\eqref{eq:tilt-def} with $0$, and solving for $\eta$. \hfill \qed

\section[Appendix: Proof of Lemma~\ref{lem:unif-elliptic}]{Proof of Lemma~\ref{lem:unif-elliptic}} \label{app:unif-elliptic}
Given $\mathcal{M}$ and $\mathcal{B}$, we need to show that there exists a positive constant $C$ (independent of  $\textbf{p}, \xi$) such that for all $\textbf{p} \in \mathcal{M}$ and all vectors $\xi \in T_{\textbf{p}}\mathcal{M}$ \nolinebreak we \nolinebreak have
    \begin{align}\label{eq:need-elliptic}
        \left\langle \xi\xi^{\top} , \, \tfrac{1}{2} \int_{\{\textup{\textbf{s}} \in T_{\textup{\textbf{p}}}\mathcal{M}: \| L_{\textup{\textbf{p}}}(\textup{\textbf{s}}) \|_\B \leq 1\}} \textup{\textbf{s}} \textup{\textbf{s}}^{\top} d \textup{\textbf{s}} \right\rangle \,\,\, \geq \,\,\, C  \left\| \xi \right\|_2^2.
    \end{align}
    To this end, use linearity of the integral to rewrite the left-hand side of \eqref{eq:need-elliptic} \nolinebreak as
    \begin{equation}\label{eq:elliptic-2}
       \frac{1}{2} \int_{\{\textup{\textbf{s}} \in T_{\textup{\textbf{p}}}\mathcal{M}: \| L_{\textup{\textbf{p}}}(\textup{\textbf{s}}) \|_\B \leq 1\}} \langle \xi \xi^{\top}, \textup{\textbf{s}} \textup{\textbf{s}}^{\top} \rangle \, d \textup{\textbf{s}} \,\, = \,\,  \frac{1}{2} \int_{\{\textup{\textbf{s}} \in T_{\textup{\textbf{p}}}\mathcal{M}: \| L_{\textup{\textbf{p}}}(\textup{\textbf{s}}) \|_\B \leq 1\}} \langle \xi, \textup{\textbf{s}} \rangle^2 \, d \textup{\textbf{s}}. 
    \end{equation}
    By the equivalence of norms on $\mathbb{R}^D$, there exists a positive constant $c$ such that for all $\textbf{v} \in \mathbb{R}^D$ we have that $\| \textbf{v} \|_{2} \leq c$ implies $\| \textbf{v} \|_{\mathcal{B}} \leq 1$.  In particular, the domain of integration in Eq.~\eqref{eq:elliptic-2} is inner-approximated by
    \begin{align*}
        \{\textbf{s} \in T_{\textbf{p}}\mathcal{M} : \| L_{\textbf{p}}(\textbf{s})\|_2 \leq c \} \, \subseteq \, \{\textbf{s} \in T_{\textbf{p}}\mathcal{M} : \| L_{\textbf{p}}(\textbf{s})\|_{\mathcal{B}} \leq 1\}.
    \end{align*}
   Since the integrand in \eqref{eq:elliptic-2} is non-negative, it follows
    \begin{equation*}
        \frac{1}{2} \int_{\{\textup{\textbf{s}} \in T_{\textup{\textbf{p}}}\mathcal{M}: \| L_{\textup{\textbf{p}}}(\textup{\textbf{s}}) \|_\B \leq 1\}} \langle \xi, \textup{\textbf{s}} \rangle^2 \, d \textup{\textbf{s}} \,\, \geq \,\, \frac{1}{2} \int_{\{\textup{\textbf{s}} \in T_{\textup{\textbf{p}}}\mathcal{M}: \| L_{\textup{\textbf{p}}}(\textup{\textbf{s}}) \|_{2} \leq c\}} \langle \xi, \textup{\textbf{s}} \rangle^2 \, d \textup{\textbf{s}}.
    \end{equation*}
    Since $L_{\textbf{p}}$ is an isometry, the right-hand side is
    \begin{equation*}
         \frac{1}{2} \int_{\{\textup{\textbf{s}} \in T_{\textup{\textbf{p}}}\mathcal{M}: \| \textup{\textbf{s}} \|_{2} \leq c\}} \langle \xi, \textup{\textbf{s}} \rangle^2 \, d \textup{\textbf{s}}.
    \end{equation*}
Using rotational symmetry of Euclidean balls, this equals
\begin{equation}\label{eq:elliptic-3}
\left( \frac{1}{2} \int_{\{\textup{\textbf{s}} \in T_{\textup{\textbf{p}}}\mathcal{M}: \| \textup{\textbf{s}} \|_{2} \leq c\}} s_1^2 \, d \textup{\textbf{s}} \right) \|\xi\|_2^2,
\end{equation}
where $s_1$ denotes the first coordinate of $\textbf{s}$ with respect to the fixed orthonormal basis on $T_{\textbf{p}}\mathcal{M}$ (Section~\ref{subsec:prelim-riem}). Now note the parenthesized quantity in \eqref{eq:elliptic-3} is a positive constant $C$ depending only on $c$ and the manifold dimension $d$.
By what we have said, it satisfies the bound \eqref{eq:need-elliptic} as desired. \hfill \qed

\section[Appendix: Proof of Proposition~\ref{prop:continuity-properties}]{Proof of Proposition~\ref{prop:continuity-properties}}\label{app:continuity-properties}
\begin{enumerate}
     \item Denote the function \eqref{eq:cont-pf-state-1} by $F : \mathcal{M} \rightarrow \operatorname{Sym}^2(T\mathcal{M})$. 
       Let $(\textbf{p}_k)_{k=1}^{\infty} \subseteq \mathcal{M}$ be a sequence converging to $\textbf{p} \in \mathcal{M}$. 
    To move to one fixed space we identify tangent spaces using the Levi-Civita connection on $\mathcal{M}$. After choosing a smooth path $\gamma:[0,1] \rightarrow \mathcal{M}$ such that $\gamma(\tfrac{1}{k}) = \textbf{p}_k$ for each $k \geq 1$ and $\gamma(0)=1$, the Levi-Civita connection gives isometries $\tau_{k}: T_{\p}\mathcal{M} \rightarrow  T_{\p_k}\mathcal{M}$.  Furthermore, $\tau_k$ converges to the identity map on $T_{\p}\mathcal{M}$ as elements of $(T\mathcal{M})^* \otimes T\mathcal{M}$ as $k \rightarrow \infty$.

     We want to show $F(\p_k) \rightarrow F(\p)$ in $\operatorname{Sym}^2(T\mathcal{M})$. 
     It suffices to show $(\tau_k^{-1} \otimes \tau_k^{-1})(F(\p_k)) \rightarrow F(\p)$ in $\operatorname{Sym}^2(T_{\p}\mathcal{M})$ (last sentence of the previous paragraph). 
    Changing variables $\s \leftarrow \tau_k^{-1}(\s)$ and using that $\tau_k$ is an isometry, we have
     \begin{align*}
         (\tau_k^{-1} \otimes \tau_k^{-1})(F(\p_k)) = 
         \frac{1}{2} \int_{\s \in T_{\p}\mathcal{M} : \| L_{\p_k}(\tau_k(\s))\|_{\mathcal{B}} \leq 1} \s \s^{\top} d\s.
         \end{align*}
        Write this as 
\begin{align*}
    \int_{\s \in T_{\p}\mathcal{M}} \mathbb{1}( \| L_{\p_k}(\tau_k(\s))\|_{\mathcal{B}} < 1) \, \textbf{s} \textbf{s}^{\top} d\textbf{s} \, \in \, \operatorname{Sym}^2(T_{\p}\mathcal{M}).
\end{align*}
Compare this to 
\begin{align*}
    F(\p) = \int_{\s \in T_{\p}\mathcal{M}} \mathbb{1}(\| L_{\p}(\s) \|_{\mathcal{B}} < 1) \textbf{s} \, \in \, \operatorname{Sym}^2(T_{\p}\mathcal{M}). \textbf{s}^{\top} d\s
\end{align*}
Since $L_{\p_k} \rightarrow L_{\p}$, $\tau_k \rightarrow \operatorname{Id}_{T_{\p}\mathcal{M}}$  and $\| \cdot \|_{\mathcal{B}}$ is continuous on $\mathbb{R}^D$, for each $\s \in T_{\p}\mathcal{M}$ there is the pointwise convergence:
\begin{align*}
    \mathbb{1}( \| L_{\p_k}(\tau_k(\s))\|_{\mathcal{B}} < 1) \, \textbf{s} \textbf{s}^{\top}  \longrightarrow \mathbb{1}(\|L_{\p}(\s)\|_{\mathcal{B}} < 1) \textbf{s} \textbf{s}^{\top}
\end{align*}
Also, letting $c \in \mathbb{R}_{>0}$ be a constant such that $\| \textbf{u} \|_{2} \leq c \| \textbf{u} \|_{\mathcal{B}}$ for all $\textbf{u} \in \mathbb{R}^D$, we have the uniform bound:
\begin{align*}
     \| \mathbb{1}( \| L_{\p_k}(\tau_k(\s))\|_{\mathcal{B}} < 1) \, \textbf{s} \textbf{s}^{\top}\|_F \leq c^2 \quad \textup{for all } \s \in T_{\p}\mathcal{M} \textup{ and } k \geq 1,
\end{align*}
since $L_{\textbf{p}}$ and $\tau_k$ are both isometries.  
Hence, the bounded convergence theorem is applicable, and implies $(\tau_k \otimes \tau_k)(F(\p_k)) \rightarrow F(\p)$.

\medskip

\item 
Denote the function \eqref{eq:cont-pf-state-2} by $G : \mathcal{M} \rightarrow \operatorname{Sym}(T\mathcal{M})$.  
Let $\p$ be a point satisfying the stated assumption.  
There exists an open neighborhood $\mathcal{U}$ of $\p$ in $\mathcal{M}$ such that for each $\p_* \in \mathcal{U}$ the norm $\| \cdot \|_{\mathcal{B}}$ is $C^1$ in some neighborhood of $L_{\p_*}(T_{\p_*}\mathcal{M}) \cap \mathbb{S}^{D-1}$.  Let $(\p_k)_{k=1}^{\infty} \subseteq \mathcal{U}$ be a sequence converging to $\p$. 
Identifying tangent spaces as above, it suffices to show $\tau_k^{-1}(F(\p_k)) \rightarrow F(\p)$.  

By the local $C^1$ assumption, Proposition \ref{prop:simple-title-C1}, item 2 applies and gives 
\begin{align*}
   F(\p) = \int_{\widehat{\s} \in T_{\p}\mathcal{M} : \|\widehat{\s}\|_2=1} \!\!\!\! -\widehat{\s} \| L_{\p}(\widehat{\s}) \|^{-d-2}_{\mathcal{B}} \frac{\left\langle \operatorname{grad} \| \cdot \|_{\mathcal{B}} (L_{\p}(\widehat{\s})), \, \tfrac{1}{2} Q_{\p}(\widehat{\s}) \right\rangle}{\left\langle \operatorname{grad} \| \cdot \|_{\mathcal{B}} (L_{\p}(\widehat{\s})), \, L_{\p}(\widehat{\s})  \right\rangle} \, d\widehat{\s}.
\end{align*}
Likewise, by a change of variables using that $\tau_k^{-1}$ preserves unit spheres:
\begin{align*}
    & \hspace{0.8em} \tau_k^{-1}(F(\p_k))= \\[0.3em] &  \int_{\widehat{\s} \in T_{\p}\mathcal{M} : \|\widehat{\s}\|_2=1} \!\!\!\!\!\!\!\!\! -\widehat{\s} \| L_{\p_k}(\tau_k(\widehat{\s})) \|^{-d-2}_{\mathcal{B}} \frac{\left\langle \operatorname{grad} \| \cdot \|_{\mathcal{B}} (L_{\p_k}(\tau_k(\widehat{\s}))), \, \tfrac{1}{2} Q_{\p_k}(\tau_k(\widehat{\s})) \right\rangle}{\left\langle \operatorname{grad} \| \cdot \|_{\mathcal{B}} (L_{\p_k}(\tau_k(\widehat{\s}))), \, L_{\p_k}(\tau_k(\widehat{\s}))  \right\rangle}  d\widehat{\s}.
\end{align*}
Boundedness and pointwise convergence hold since $\operatorname{grad} \| \cdot \|_{\mathcal{B}}$ is locally continuous.  So the bounded convergence theorem implies the first sentence in the statement.  The second sentence follows from the example in Section~\ref{subsec:circle-example}. \hfill \qed
\end{enumerate}

\section[Appendix: Tail bounds and absolute moments of the Gaussian]{Tail bounds and absolute moments of the Gaussian} \label{app:gaussian}

We recall some basic properties of the Gaussian.  As in Section \ref{subsec:switch-indicator}, $\kappa_{\sigma}(s) := 
\frac{2s}{\sigma^2}e^{-s^2/\sigma^2}$\!\!.

\begin{itemize}
    \item For each even $k \geq 0$ and $\delta \geq 0$, by substitution and then integration by parts $k/2$ times, 
    \begin{align}
        &\int_{s=\delta}^{\infty} s^k \kappa_{\sigma}(s) ds \nonumber \\ &= \sigma^k e^{-\delta^2/\sigma^2} \! \left( \! \left(\frac{\delta^2}{\sigma^2}\right)^{\frac{k}{2}} \!\! + \, \frac{k}{2} \left(\frac{\delta^2}{\sigma^2}\right)^{\frac{k}{2}-1} \!\!\!\! + \, \frac{k}{2}\left(\frac{k}{2} - 1\right) \!\! \left(\frac{\delta^2}{\sigma^2}\right)^{\frac{k}{2}-2} \!\!\! + \ldots + \left(\frac{k}{2}\right)! \! \right) \nonumber \\
        &= e^{-\delta^2/\sigma^2}\textup{poly}(\sigma, \delta). \label{eq:gaussian-even-tail}
    \end{align}
    \item For each odd $k \geq 0$ and $\delta > 0$, using $s/\delta \geq 1$ for $s \in [\delta, \infty]$ and Eq.~\eqref{eq:gaussian-even-tail},
    \begin{equation} \label{eq:gaussian-odd-tail}
        \int_{s=\delta}^{\infty} s^k \kappa_{\sigma}(s) ds \leq (1/\delta) \int_{s=\delta}^{\infty} s^{k+1} \kappa_{\sigma}(s) ds = e^{-\delta^2/\sigma^2} (1/\delta) \, \textup{poly}(\sigma, \delta).
    \end{equation}
    \item For each $k \geq 0$, from \cite[Equation~18]{Winkelbauer2012},
    \begin{equation} \label{eq:gaussian-central-moment}
        \int_{s=0}^{\infty} s^k \kappa_{\sigma}(s) ds = \sigma^k \Gamma\left(\tfrac{k+2}{2}\right).
    \end{equation}
\end{itemize}

\section[Appendix: Numerical estimation of one-dimensional eigenfunctions]{Numerical estimation of one-dimensional eigenfunctions} \label{sec:numerical_eigenfunction_computation}

\begin{figure}
    \includegraphics[width=0.97\linewidth]{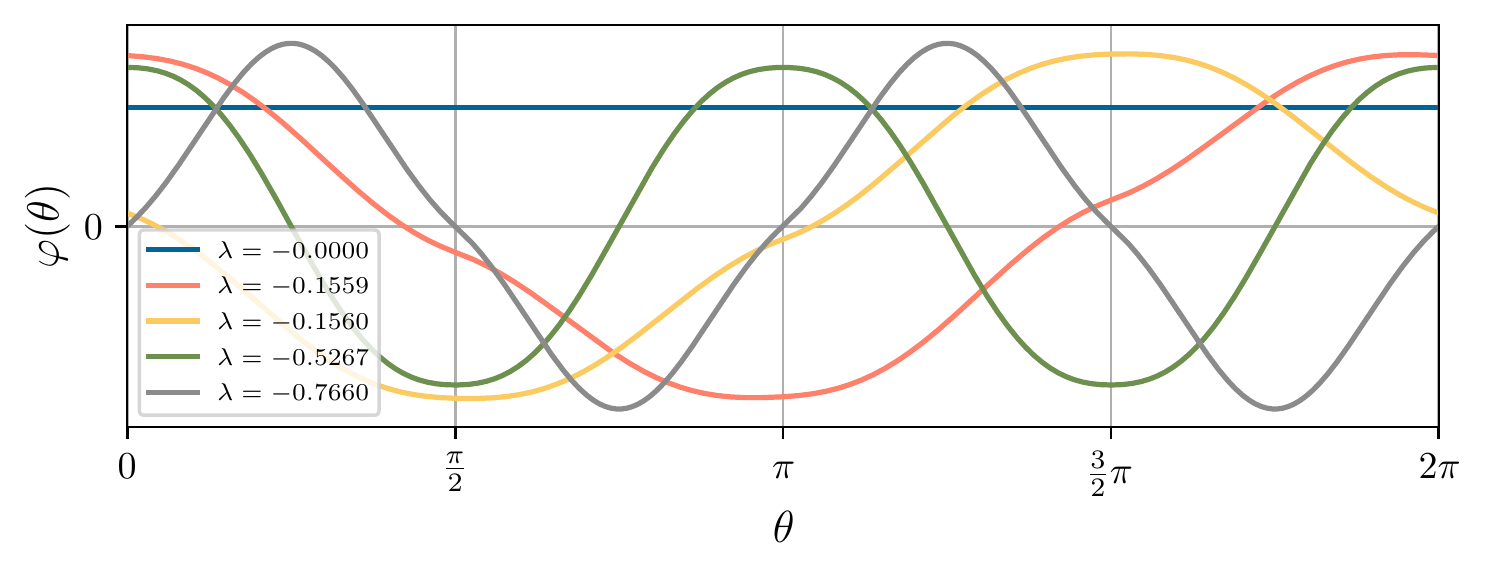}
    \includegraphics[width=0.97\linewidth]{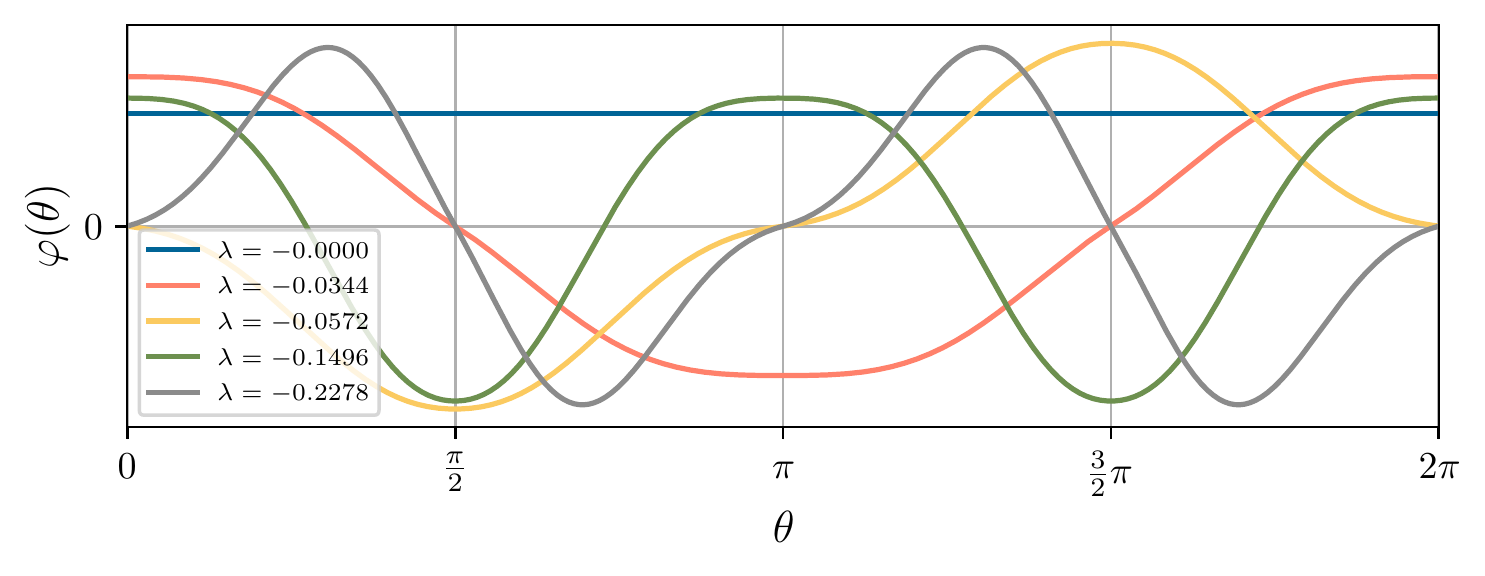}
    \includegraphics[width=0.97\linewidth]{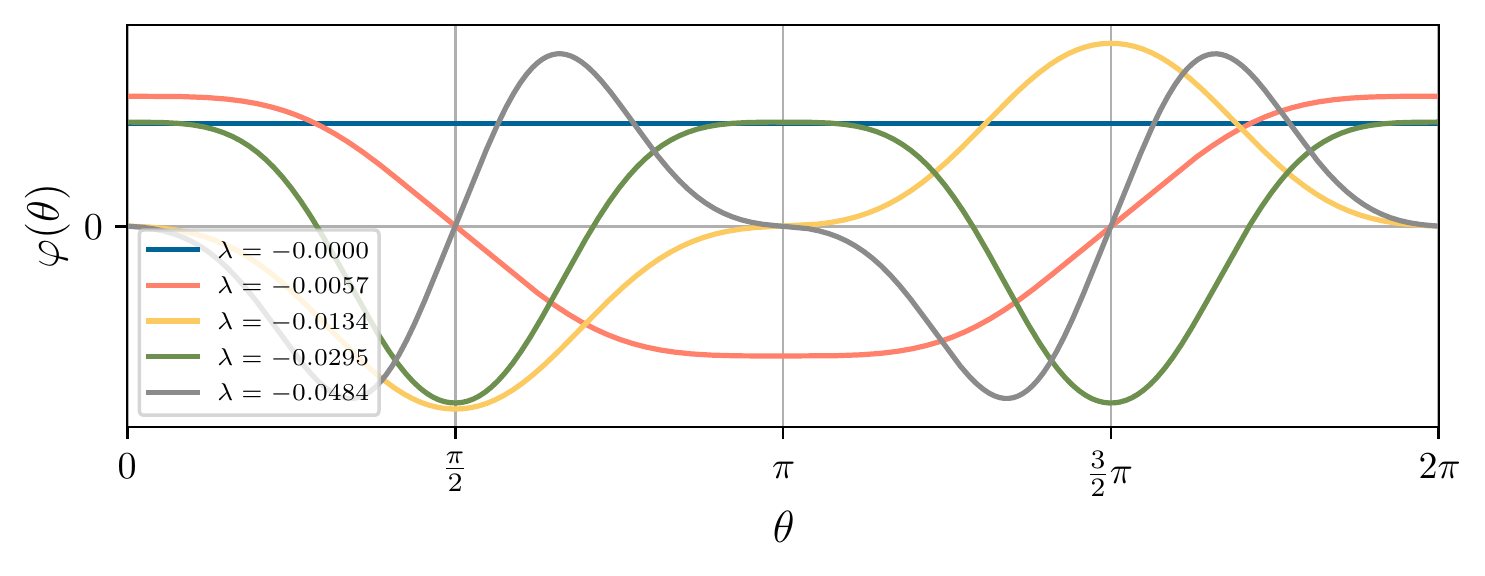}
    \includegraphics[width=0.97\linewidth]{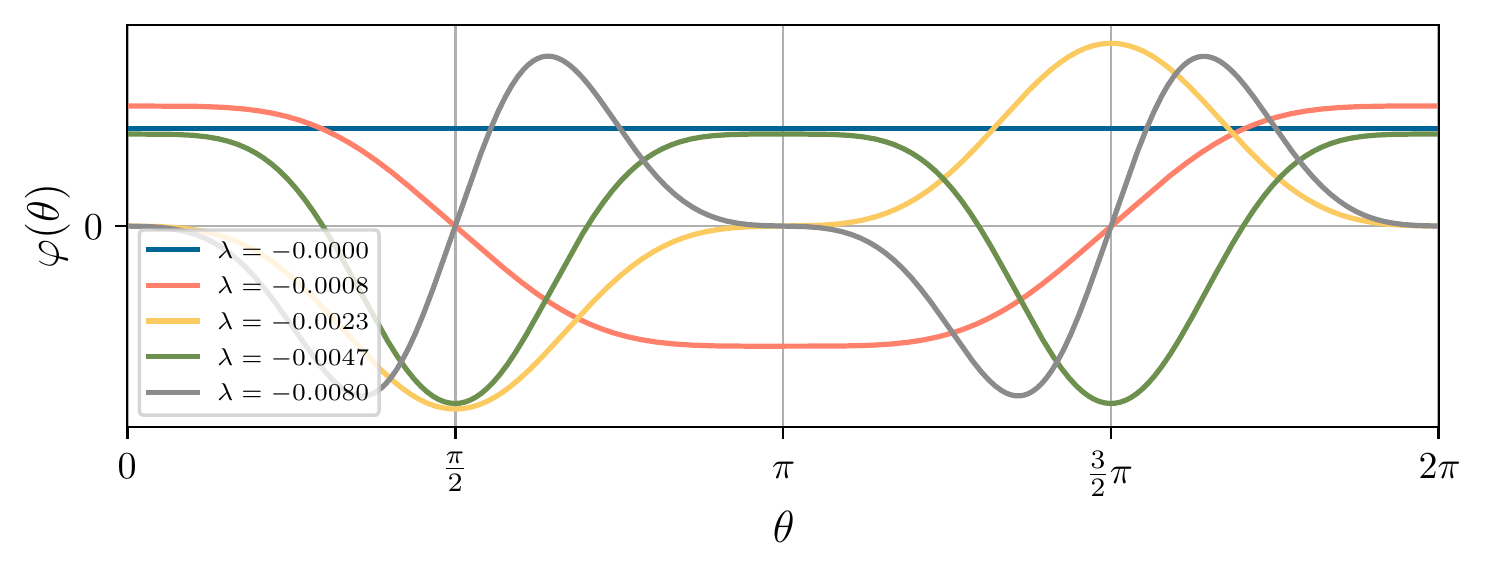}
    \caption{\textit{Eigenfunctions}.  The five eigenfunctions with eigenvalues smallest in magnitude for the weighted $\ell_1$ Laplacian on the unit circle \eqref{eq:limit-op-circle}. These were computed numerically.
                In these plots, $w_2 = 1$. All the choices $w_1 \in \{1, 2, 4, 8\}$ are displayed from top to bottom.}
    \label{fig:weighted_L1_eigenfunctions}
\end{figure}

The eigenfunctions of the limiting operator are of key interest for manifold learning methods in general.
For the case of the circle example (Section~\ref{subsec:circle-example}), these are the functions $\varphi:[0,2\pi] \to \mathbb{R}$ that solve the following generalized Helmholtz boundary value problem:
\begin{align} \label{eq:L1_laplacian_BVP}
    \Delta_{\mathcal{M}, \mathcal{B}} \, \varphi + \lambda \varphi = 0,
\end{align}
where $\Delta_{\mathcal{M}, \mathcal{B}}$ is the limiting Laplacian-like differential operator in Eq.~\eqref{eq:limit-op-circle}, subject to
 the periodic boundary conditions:
\begin{align*}
    \varphi(\theta + 2 \pi) &= \varphi(\theta),\\
    \varphi'(\theta + 2 \pi) &= \varphi'(\theta).
\end{align*}
Figure~\ref{fig:weighted_L1_eigenfunctions} shows numerically computed solutions of Eq.~\eqref{eq:L1_laplacian_BVP} for different choices of $w_1, w_2$. 
Notice the eigenfunctions are oscillatory, as dictated by Sturm-Liouville theory \cite{Algwaiz2008}.

We describe the numerical computation of these limiting eigenfunctions. 
We used a standard finite-difference scheme where the first derivative was replaced by a symmetrized difference
\begin{align} \label{eq:first_derivative_finite_diff}
    \frac{d f}{d \theta} \to \frac{f(\theta + \Delta \theta) - f(\theta - \Delta \theta)}{2 \Delta \theta},
\end{align}
and the second derivative by
\begin{align} \label{eq:second_derivative_finite_diff}
    \frac{d^2 f}{d \theta^2} \to\frac{f(\theta + \Delta\theta) - 2 f(\theta) + f(\theta - \Delta \theta)
}{(\Delta \theta)^2}.
\end{align}
In this equation, $f$ is taken to be a cyclic function defined over the discrete range 
\begin{align*}
    \left\{0, \frac{2\pi}{n}, \ldots, \frac{2\pi (n-1)}{n}\right\}.
\end{align*}
To compute the solutions we formed a sparse $n \times n$ matrix $L$ that corresponds to the finite-difference operator formed by substituting \eqref{eq:first_derivative_finite_diff} and \eqref{eq:second_derivative_finite_diff} into the first and second derivative terms in Eq.~\eqref{eq:L1_laplacian_BVP}.
The eigenvalues and eigenvectors of $L$ were found using the function \texttt{scipy.sparse.linalg.eigs()} from the \texttt{SciPy}
package \cite{SciPy2020}. It is a wrapper of the ARPACK  library for large-scale eigenvalue problems \cite{ARPACK1998}.
Recall that in our problem, all eigenvalues are non-positive.
To obtain the smallest (in magnitude)\ eigenvalues and their corresponding eigenvectors, we used the \texttt{eigs()} function
in shift-invert mode with $\sigma=1$.
The particular choice of $\sigma$ did not seem to matter much when $\sigma > 0$, however choosing $\sigma=0$ resulted in
instabilities and convergence errors.
This is due to the fact that shift-invert mode attempts to find solutions to $(L-\sigma I)^{-1}\x = \lambda \x$, and since
zero is an eigenvalue of $L$, the choice $\sigma=0$ results in the inversion of an ill-conditioned matrix.
The use of sparse matrices allows one to take large values of $n$, since applying the finite-difference
operator defined above costs only $O(n)$.

\bibliographystyle{spmpsci-modified} 

\addcontentsline{toc}{section}{References}
\bibliography{earthmover}
\end{document}